\documentclass[10pt]{article} % For LaTeX2e
\usepackage[accepted]{tmlr}
\usepackage{amsmath,amsfonts,bm}

\def\eqref#1{equation~\ref{#1}}
\def\1{\bm{1}}

\DeclareMathAlphabet{\mathsfit}{\encodingdefault}{\sfdefault}{m}{sl}
\SetMathAlphabet{\mathsfit}{bold}{\encodingdefault}{\sfdefault}{bx}{n}

\usepackage{hyperref}
\usepackage{url}
\usepackage{algorithm}    % \begin{algorithm}
\usepackage{algpseudocode}  % \begin{algorithmic}
\usepackage{amssymb}      % \varnothing
\usepackage{amsthm}       % \begin{proposition}
\usepackage{booktabs}     % \midrule
\usepackage{caption}      % \captionof
\usepackage{cleveref}     % \cref
\usepackage{float}        % float option H
\usepackage{graphicx}     % \includegraphics
\usepackage{placeins}     % \FloatBarrier
\usepackage{subfig}       % \subfloat
\usepackage{wrapfig}
\usepackage{cancel}

\theoremstyle{plain}
\newtheorem{proposition}{Proposition}[section]

\crefname{equation}{Eq.}{Eqs.}
\crefname{section}{Sec.}{Secs.}
\crefname{table}{Tab.}{Tabs.}
\crefname{figure}{Fig.}{Figs.}
\crefname{algorithm}{Alg.}{Algs.}
\crefname{proposition}{Prop.}{Props.}

\title{Stabilizing Consistency Training:\\A Flow Map Analysis and Self-Distillation}

\author{\name Youngjoong Kim \email noah.kim@snu.ac.kr \\
      \addr Department of Computer Science and Engineering\\
      Seoul National University
      \AND
      \name Duhoe Kim \email dhkimm@snu.ac.kr \\
      \addr Interdisciplinary Program in Artificial Intelligence\\
      Seoul National University
      \AND
      \name Woosung Kim \email k020419@snu.ac.kr \\
      \addr Interdisciplinary Program in Artificial Intelligence\\
      Seoul National University
      \AND
      \name Jaesik Park \email jaesik.park@snu.ac.kr \\
      \addr Department of Computer Science and Engineering\\
      Seoul National University}

\def\month{10}  % Insert correct month for camera-ready version
\def\year{2026} % Insert correct year for camera-ready version
\def\openreview{\url{https://openreview.net/forum?id=Qx2FrHDVzp}} % Insert correct link to OpenReview for camera-ready version

\begin{document}

\maketitle

\begin{abstract}
Consistency models have been proposed for fast generative modeling, achieving results competitive with diffusion and flow models. However, these methods exhibit instability and limited reproducibility when training from scratch, motivating subsequent work to explain and stabilize these issues. While these efforts have provided valuable insights, the explanations remain fragmented, and the theoretical relationships remain unclear. In this work, we provide a theoretical examination of consistency models by analyzing them from a flow map-based perspective. This joint analysis clarifies how training and convergence behavior can give rise to degenerate solutions. Building on these insights, we revisit self-distillation as a practical remedy for certain forms of suboptimal convergence. To avoid excessive gradient norms during its optimization, we reformulate its objective empirically, yielding \textit{improved Self-Distillation (iSD)}. We further extend iSD with training-time classifier-free guidance, yielding iSD-T, which improves performance and reproducibility in the studied settings. We demonstrate that iSD extends beyond image generation to diffusion-based policy learning, without reliance on pretrained diffusion models for initialization. Code is available at \url{https://github.com/SNU-VGILab/improvedSelfDistillation}.
\end{abstract}

\section{Introduction}
\label{sec:intro}

Diffusion~\citep{song2020generativemodelingestimatinggradients, ncsnv2, NEURIPS2020_DDPM, song2021scorebased} and flow matching models~\citep{rf, flowmatching} have achieved remarkable performance across a wide range of applications. This progress stems from flow-based modeling and multistep inference, but is limited by multiple network evaluations during generation.

Consistency Models~\citep{consistencymodel, shortcutmodel, meanflow, facm} have been proposed as few-step generation methods for both training-from-scratch and distillation settings. Earlier consistency training methods have often exhibited training instability and limited reproducibility in scratch settings, motivating subsequent work to stabilize training~\citep{ict, ect, sct}. While prior efforts have provided meaningful insights, the explanations remain fragmented, and the theoretical relationships among existing approaches remain unclear.

Prior work has explored distillation perspectives on consistency models from complementary viewpoints. \citet{flowmapmatching} provides a flow map framework that organizes few-step distillation methods into Eulerian, Lagrangian, and semigroup-type formulations. This framework offers a unified perspective for distillation objectives, while also identifying objective-level discrepancies between training-from-scratch and distillation.

Since these discrepancies may steer training toward undesired solutions, self-distillation~\citep{boffi2025} recasts training-from-scratch methods through a distillation view. This reduces the objective-level gap by aligning the guiding velocity with the learned flow. However, Eulerian self-distillation is reported to produce large gradient norms in high-dimensional settings, and its direct use can still pose optimization challenges.

\begin{figure}[t]
\centering
\includegraphics[width=1.0\linewidth]{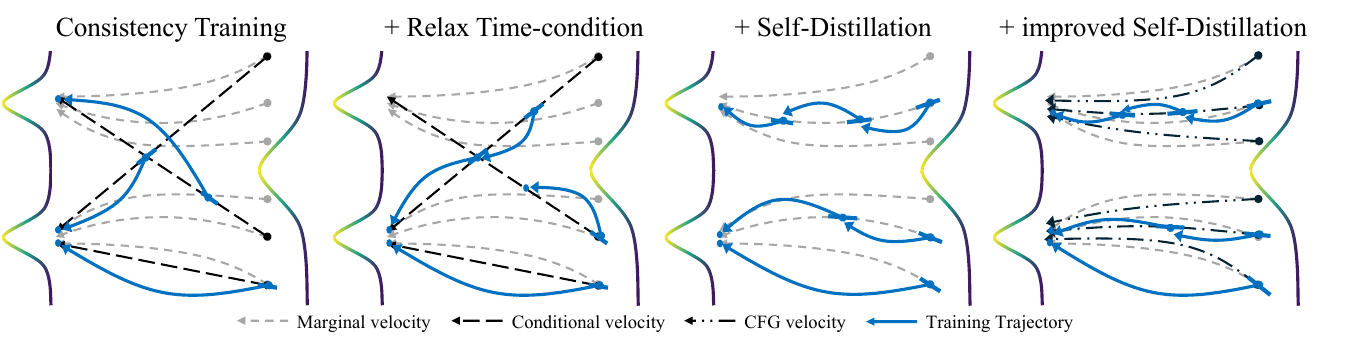}
\caption{From consistency training to improved Self-Distillation. Consistency training learns a mapping over conditional velocity, often suffering from training instability and reproducibility issues. Relaxing the time condition mitigates this instability, and self-distillation aligns a target with the marginal velocity field. However, directly applying self-distillation leads to unstable training. We therefore reformulate the objective and incorporate classifier-free guidance (iSD-T), further stabilizing training and improving reproducibility.}
\label{fig:intro}
\end{figure}

In this work, we systematically examine training instability and reproducibility in consistency models. We jointly analyze existing approaches through two training objectives, Eulerian distillation and consistency training. Along these objectives, we also explore key design factors: weight initialization and time condition.

This perspective clarifies how distinct objectives lead to different convergence behaviors. First, while prior work has identified mismatches between training objectives, it has not been formally established whether such differences necessarily imply different optima. We address this question by characterizing the stationary conditions of Eulerian distillation. We show that training with conditional velocity can admit degenerate solutions, whereas training with marginal velocity can yield the desired optimum.

Second, while consistency training admits a fixed-point solution corresponding to the desired optimum, we show that this holds only at the expectation level. Under finite-batch optimization, the residual gap can perturb the fixed-point training dynamics, providing a possible explanation for the effect of small batch sizes.

Third, we examine complementary factors that affect which solution is reached in practice. The fixed-point structure contributes to sensitivity to initialization, undermining reproducibility, while the time condition affects the loss variance and spikes of the flow map objective.

Motivated by the analysis, we revisit self-distillation as a practical mechanism for addressing certain forms of instability. Our analysis indicates that marginal velocity guidance can yield desired optima under these settings, while the relaxed time condition can improve training stability. Self-distillation naturally provides this form of marginal guidance, but existing formulations are not directly compatible with consistency training due to the gradient norm problem in high-dimensional settings~\citep{boffi2025}.

Building on the formulation introduced by \citet{alignyourflow}, we reformulate the self-distillation objective empirically to constrain gradient magnitudes while preserving its intended guidance. This smooths its loss landscape in practice, and we call this approach \textit{improved Self-Distillation} (iSD). We show that this reformulated objective can be seamlessly combined with classifier-free guidance~\citep{cfg}, yielding iSD-T (\textbf{T}raining-time CFG). We provide an overview of our flow map perspective in \cref{fig:intro}.

We evaluate iSD-T on ImageNet-1K $256\!\times\!256$ in few-step generation. It exhibits stable behavior with improved reproducibility and achieves performance comparable to that of recent methods without relying on pretrained diffusion models for initialization. We extend our evaluation to diffusion-based policy learning, demonstrating the applicability of iSD beyond the image generation domain.

\section{Related Work}

\textbf{Diffusion and Flow Matching Models.} Diffusion models~\citep{NEURIPS2020_DDPM, song2020generativemodelingestimatinggradients} and flow matching models~\citep{albergo2023building, flowmatching, rf} are generative models that gradually transform a tractable noise distribution into the data distribution. These models have achieved remarkable progress in high-fidelity generation~\citep{ldm, sdxl, sd3}. However, their reliance on multistep sampling requires substantial computational resources.

\textbf{Few-step Generation.} Several works have explored improving sampling efficiency of diffusion models~\citep{ProgressiveDistillation, DiffusionGan, ldm}. These approaches aim to distill pretrained diffusion models into fewer-step generators, adopt GANs, or leverage VAEs to reduce input size. In parallel, score distillation~\citep{dmd, dmd2, sid} has been proposed to construct one-step generators, while both approaches rely on additional distillation stages or auxiliary networks, increasing training cost.

\textbf{Consistency Models.} Consistency Models~\citep{consistencymodel} are designed to predict a sample directly from any point along a flow trajectory. Both distillation and training methods have been proposed, whereas training from scratch can exhibit large loss variance, motivating earlier work on stabilization~\citep{ict, ect, sct}. Other studies~\citep{flowmapmatching, gc, vct} have suggested that a discrepancy between distillation and training objectives can lead to high loss variance when training from scratch. Network-induced couplings~\citep{gc, vct} are introduced to reduce loss variance, and self-distillation~\citep{boffi2025} distills shortcut mappings from its jointly learned flow to reduce this discrepancy. However, Eulerian self-distillation has been reported to encounter large gradient norms in high-dimensional settings.

\textbf{Unified Framework.} Recently, several studies have aimed to relate flow matching and consistency models. UCGM~\citep{ucgm} integrates both paradigms, but does not account for the relaxed mapping constraint of arbitrary time points. \citet{flowmapmatching, ctm} present mathematical frameworks for consistency models, defining a model that learns \textit{flow maps} as mappings between any two points on the same trajectory. While these works offer valuable theoretical perspectives, instability mechanisms arising from optimization noise, initialization, or time conditioning are not the primary focus of these analyses.

\section{Preliminary}

\textbf{Flow Matching.} Given a dataset $X$ with underlying distribution $p_X$, flow matching models are trained to match the velocity fields of continuous flows, starting from a tractable distribution $p_Z$. Prior work constructs such flows via an interpolation, $x_t = \alpha_tx + \sigma_t z$, where $x\sim p_X$ and $z\sim p_Z$.

Let $\alpha_t$ and $\sigma_t$ be continuous and monotone, with $\alpha_0 = \sigma_T = 1$ and $\alpha_T = \sigma_0 = 0$ for $t\in [0, T]$. A marginal distribution $\rho_t$ induced by the flow satisfies $\rho_0=p_X$ and $\rho_T= p_Z$. For a well-defined flow map, $\nu_t = \alpha_t\sigma'_t - \sigma_t\alpha'_t \ne 0$, and we further assume $\nu_t = \nu \ne 0$ for all $t\in [0, T]$, where $\nu$ is a constant. Notably, linear and trigonometric interpolations satisfy this with $\nu=1$, which are our primary focus (see \cref{proof:interpolation}).

With the constructed flow, the flow matching models optimize the squared error between the conditional velocity $v_t(x_t|x) = \alpha'_tx + \frac{\sigma'_t}{\sigma_t}(x_t - \alpha_tx)$ and a network $F_\theta(x_t; t)$:
\begin{align}\label{eq:cfm}
    \mathcal L_\text{CFM} = \mathbb E_{x,z,t}\left[\|F_\theta(x_t; t) - v_t(x_t|x)\|^2_2\right],\qquad F_\theta(x_t; t)\approx v^*_t(x_t) = \mathbb E_{x|x_t}[v_t(x_t|x)].
\end{align}
Conditional flow matching $\mathcal L_\text{CFM}$ converges to the flow induced by the marginal velocity $v_t^*(x_t)$. If $v_t^*(x)$ is Lipschitz continuous in both $t$ and $x$, the ODE $dx_t = v^*_t(x_t)dt$ has a unique solution~\citep{flowmatching}. However, flow models are known to suffer from mean collapse, in which one-step samples collapse to the mean of the data distribution (see \cref{proof:mean-collapse} and \citealt{edm}).

\textbf{Flow Map.} From flows defined by interpolation, our goal is to draw samples from the target distribution in a few sampling steps. To this end, we adopt a \textit{flow map} $f(x_t; t, s)$, which is a mapping between two points $x_t$ and $x_s\ (s<t)$ on the same trajectory~\citep{ctm, flowmapmatching}:
\begin{align}\label{eq:flowmap-def}
    f(x_t; t, s) = x_t + \int^s_tv_\tau^*(x_\tau)d\tau = x_s.
\end{align}
If the marginal velocity is Lipschitz continuous, the flow map is well-defined and injective (see \cref{proof:injectivity}).

\textbf{Training Flow Map.} Since the flow map is defined as an integral, direct supervision from scratch is challenging. Recent studies adopt consistency training~\citep{consistencymodel} to avoid data generation, which can be derived from the Eulerian equation (see \cref{proof:eulerian-eqn} and \citealt{flowmapmatching}):
\begin{align}\label{eq:eulerian}
    \partial_t f(x_t; t, s) + v^*_t(x_t)\cdot\nabla_xf(x_t; t, s) = 0.
\end{align}
Note that if $f$ is continuous in $t$ and $s$, Lipschitz continuous in $x$, and satisfies the boundary condition $f(x_t; t, t) = x_t$, the flow map $f$ is the unique solution to the Eulerian equation.

To train a flow map network $f_\theta(x_t; t, s)$ using this equation, Eulerian distillation~\citep{flowmapmatching} formulates the objective as a squared minimization problem:
\begin{align}\label{eq:ed}
    \mathcal L_\mathrm{ED} = \mathbb E_{x,z,t,s}\left[\|\partial_tf_\theta(x_t; t, s) + v^*_t(x_t)\cdot\nabla_xf_\theta(x_t; t, s)\|^2_2\right].
\end{align}
This is the basic objective for training flow maps. It reduces to consistency distillation when a fixed-point-style update is adopted, and $s$ is fixed to zero (\citealt{consistencymodel}, see \cref{proof:interpretation}):
\begin{align}\label{eq:cd}
    \mathcal L_\mathrm{CD} = \mathbb E\left[\left\|f_\theta(x_t; t, s) - \operatorname{sg}\left[f_\theta(x_t; t, s) - \left[\partial_tf_\theta(x_t; t, s) + v^*_t(x_t)\cdot\nabla_xf_\theta(x_t; t, s)\right]\right]\right\|^2_2\right],
\end{align}
where $\operatorname{sg}[\cdot]$ is a stop-gradient operator. For consistency training from scratch, using $v_t(x_t|x)$ instead of $v^*_t(x_t)$ is a more common choice:
\begin{align}\label{eq:ct}
    \mathcal L_\mathrm{CT} = \mathbb E\left[\left\|f_\theta(x_t; t, s) - \operatorname{sg}\left[f_\theta(x_t; t, s) - \left[\partial_tf_\theta(x_t; t, s) + v_t(x_t|x)\cdot\nabla_xf_\theta(x_t; t, s)\right]\right]\right\|^2_2\right].
\end{align}
Since $\mathcal L_\mathrm{ED}$ in \cref{eq:ed} involves a Jacobian-vector product (JVP), its optimization requires second-order differentiation, introducing computational overhead. In contrast, consistency distillation $\mathcal L_\mathrm{CD}$ in \cref{eq:cd} and consistency training $\mathcal L_\mathrm{CT}$ in \cref{eq:ct} use a stop-gradient operation, avoiding this issue. However, they alter gradient dynamics, potentially affecting training stability.

\section{Analysis of Instability}

In this section, we analyze instability and reproducibility in consistency models from a flow map perspective, focusing on how different design choices affect convergence properties and training dynamics. Based on this analysis, we explain why training from scratch may deviate from the desired flow map in practical settings.

\subsection{Flow Map Representation for Analysis}
\label{section:generalization}

To facilitate a joint analysis of recent consistency models, we adopt a flow map representation to express different training objectives in a common form. This representation serves as an analytical tool rather than a new assumption, and will be used throughout the paper to compare convergence behaviors and training dynamics across methods.

\textbf{Flow Map Representation.} To enable a systematic analysis of training instability in consistency models, we propose a flow map representation for commonly used interpolations, expressed as a one-step Euler solution with a network $F_\theta$ (see \cref{proof:interpolation}):
\begin{align}\label{eq:gen-flowmap}
    f_\theta(x_t; t, s) = \nu^{-1}(A_{t,s}'x_t - A_{t,s}F_\theta(x_t; t, s)),\qquad A_{t,s}=\sigma_t\alpha_s - \sigma_s\alpha_t.
\end{align}
Under this representation, UCGM~\citep{ucgm} can be viewed as a special case with fixed $s=0$. TiM~\citep{Wang_2026_CVPR} also considers arbitrary $(t, s)$, and it shares the functional structure under the $\nu$-condition, while differing in parameterization.

With this representation, we propose a joint interpretation of recent consistency models (proof in \cref{proof:interpretation}).

\begin{proposition}\label{prop:instantiation} (Interpretation of Recent Methods) The methods listed in \cref{tab:generalization} can be interpreted within the flow map representation in \cref{eq:gen-flowmap}, satisfying the following transport equation:
\begin{align}
\partial_tf_\theta(x_t; t, s) + \tau_t(x_t, x)\cdot\nabla_xf_\theta(x_t; t, s) = 0,
\end{align}
where $x_t$ is given by the Interpolant, $\tau_t$ by the Trajectory, and $t, s$ by the Timestep, as summarized in \cref{tab:generalization}.
\end{proposition}

\begin{table}[t]
\centering
\caption{Summary of recent consistency models interpreted under the flow map representation. Interpolants in parentheses denote settings realized in the cited work. $\Phi_t(x_t)$ is the teacher network and $F_\theta(x_t; t, s)$ is the training network. Our method is aligned with the learned marginal velocity $F_\theta(x_t; t, t)$ and the relaxed time condition $s<t$, while reformulating Eulerian self-distillation (ESD) to reduce the gradient norm problem.}
\label{tab:generalization}
\resizebox{0.90\linewidth}{!}{%
\begin{tabular}{llllll}
\toprule
Model & Interpolant $x_t$ & Trajectory $\tau_t$ & Timestep & JVP & Objective \\
\midrule
\multicolumn{6}{l}{Distillation-based Methods} \\
\midrule
FMM-EMD~\citep{flowmapmatching} & Any (Linear) & $\Phi_t(x_t)$ & $s < t$ & Exact & $\mathcal L_\mathrm{ED}$ \\
AYF-EMD~\citep{alignyourflow} & Any (Linear) & $\Phi_t(x_t)$ & $s < t$ & Exact & $\mathcal L_\mathrm{CD}$ \\
sCD~\citep{sct} & Trigonometric & $\Phi_t(x_t)$ & $s=0$ & Exact & $\mathcal L_\mathrm{CD}$ \\
\midrule
\multicolumn{6}{l}{Consistency Training Methods} \\
\midrule
MeanFlow~\citep{meanflow} & Linear & $v_t(x_t|x)$ & $s < t$ & Exact & $\mathcal L_\mathrm{CT}$ \\
ConsistencyFM~\citep{consistencyflowmatching} & Linear & $v_t(x_t|x)$ & $s=0$ & Approx. & $\mathcal L_\mathrm{CT}$ \\
sCT~\citep{sct} & Trigonometric & $v_t(x_t|x)$ & $s=0$ & Exact & $\mathcal L_\mathrm{CT}$ \\
UCGM~\citep{ucgm} & Any (Linear, Trig.) & $v_t(x_t|x)$ & $s=0$ & Approx. & $\mathcal L_\mathrm{CT}$  \\
\midrule
\multicolumn{6}{l}{Self-Distillation Methods} \\
\midrule
Shortcut Model~\citep{shortcutmodel} & Linear & $F_\theta(x_t; t, t)$ & $s < t$ & Approx. & $\mathcal L_\text{SD}$ \\
ESD~\citep{boffi2025} & Any (Linear) & $F_\theta(x_t; t, t)$ & $s < t$ & Exact & $\mathcal L_\mathrm{SD}$ \\
\textbf{iSD (Ours)} & Any (Linear, Trig.) & $F_\theta(x_t; t, t)$ & $s < t$ & Approx. & $\mathcal L_\mathrm{SD\text{-}R}$ \\
\bottomrule
\end{tabular}
}
\end{table}

Based on this interpretation, the objectives $\mathcal L_\mathrm{ED}$, $\mathcal L_\mathrm{CD}$, and $\mathcal L_\mathrm{CT}$ provide a common basis for analyzing the methods in \cref{tab:generalization}. In the next subsection, we analyze these objectives and show how they can lead to certain training failures.

\subsection{Suboptimality and Instability}
\label{section:suboptimality}

\textbf{Guidance Velocity.} Most consistency training approaches are guided by a \textit{conditional velocity}. However, this setup can contribute to training instability. To examine its effect, we consider Eulerian distillation in \cref{eq:ed} with a conditional velocity field. Following \citet{flowmapmatching}, we refer to this as \textit{direct training}.
\begin{align}\label{eq:dt}
    \mathcal L_\mathrm{DT} = \mathbb E\left[\|\partial_tf_{\theta}(x_t; t, s) + v_t(x_t|x)\cdot\nabla_xf_{\theta}(x_t; t, s)\|^2_2\right].
\end{align}
Since the global optimum of Eulerian distillation corresponds to the flow map, we show that direct training does not guarantee convergence to this solution by comparing their stationary conditions (proof in \cref{proof:suboptimality}).

\begin{proposition}\label{prop:suboptimality} (Stationary Condition of Direct Training)
A stationary point of direct training satisfies
\begin{align}
    \mathcal S_\mathrm{DT} = \mathcal S_\mathrm{ED} + C = 0,\quad C = \nabla\cdot(\rho_t\Sigma_{\Delta v|x_t}\nabla_x f_\theta),
\end{align}
where $\mathcal S_\mathrm{ED}$ is the stationary condition of Eulerian distillation, $\Delta v = v_t(x_t|x) - v^*_t(x_t)$, and $\Sigma_{\Delta v|x_t}$ is the conditional covariance $\mathrm{Cov}_{x|x_t}[\Delta v]$. This follows from applying the Euler-Lagrange equation to $\mathcal L_\mathrm{DT}$. This suggests that training can admit a degenerate flow map ($\mathcal S_\mathrm{ED} = -C\ne 0$), which is a suboptimal solution.
\end{proposition}

Thus, the discrepancy is not only an objective-level mismatch: through the term $C$, \textit{it can actually shift the stationary point and admit degenerate flow maps}. This term arises from the gap $\Delta v$ between the marginal velocity $v^*_t(x_t)$ and the per-sample conditional velocity $v_t(x_t|x)$. We later show how this gap also affects stop-gradient objectives such as consistency training $\mathcal L_\mathrm{CT}$ (\cref{prop:instability}).

\begin{wrapfigure}{r}{0.48\linewidth}
\centering
\vspace{-2em}
\subfloat[$\mathcal L_\mathrm{DT}$]{%
\includegraphics[width=0.31\linewidth]{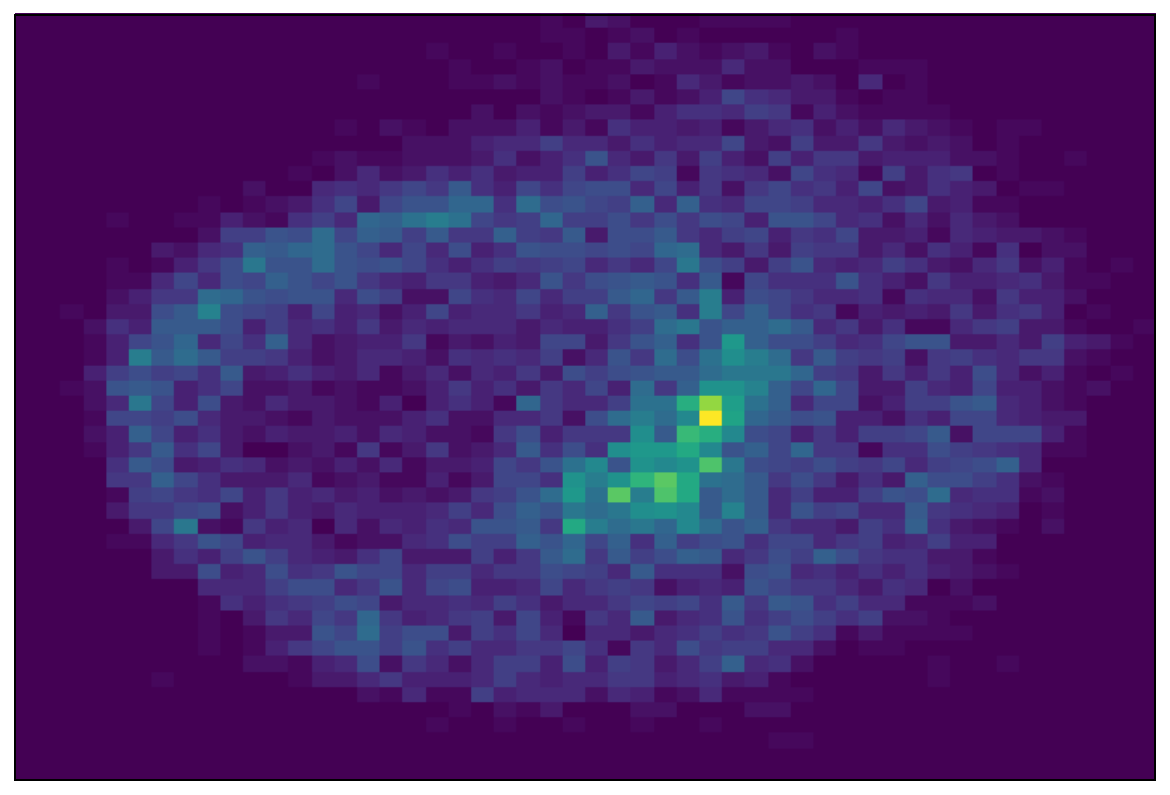}
}
\subfloat[$\mathcal L_\mathrm{ED}$]{%
\includegraphics[width=0.31\linewidth]{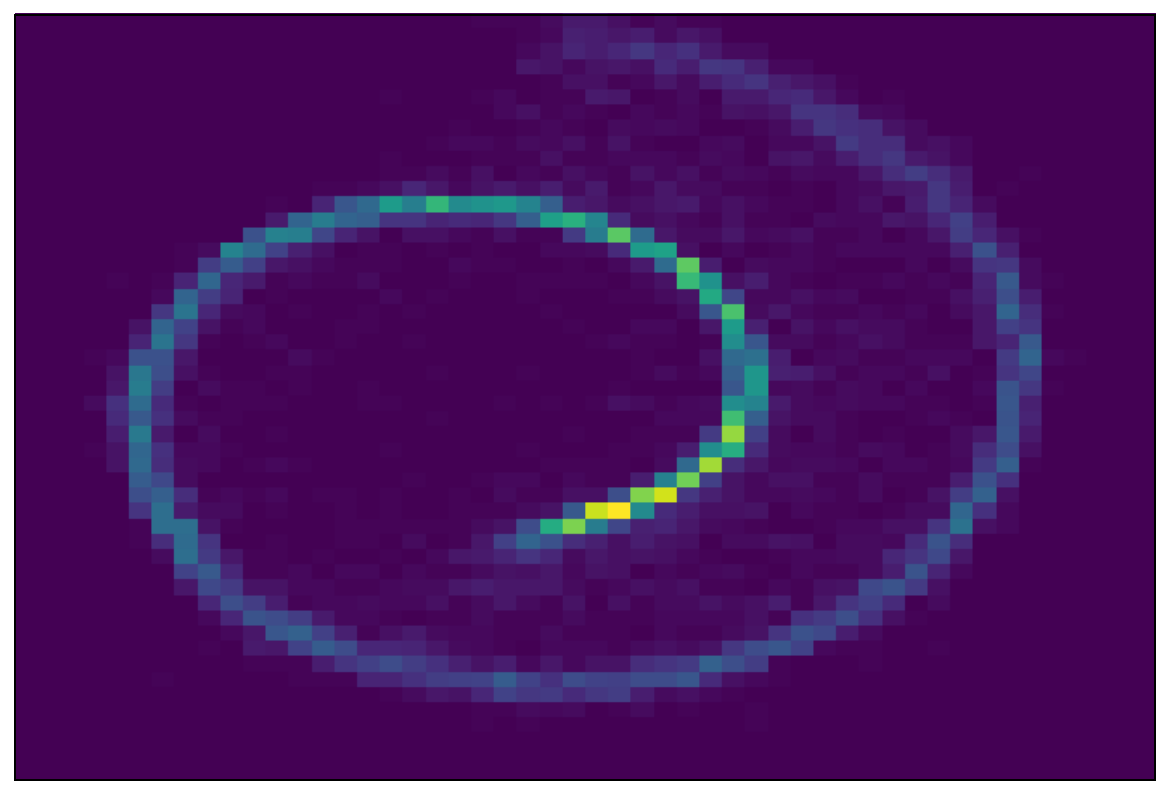}
}
\subfloat[Ground truth]{%
\includegraphics[width=0.31\linewidth]{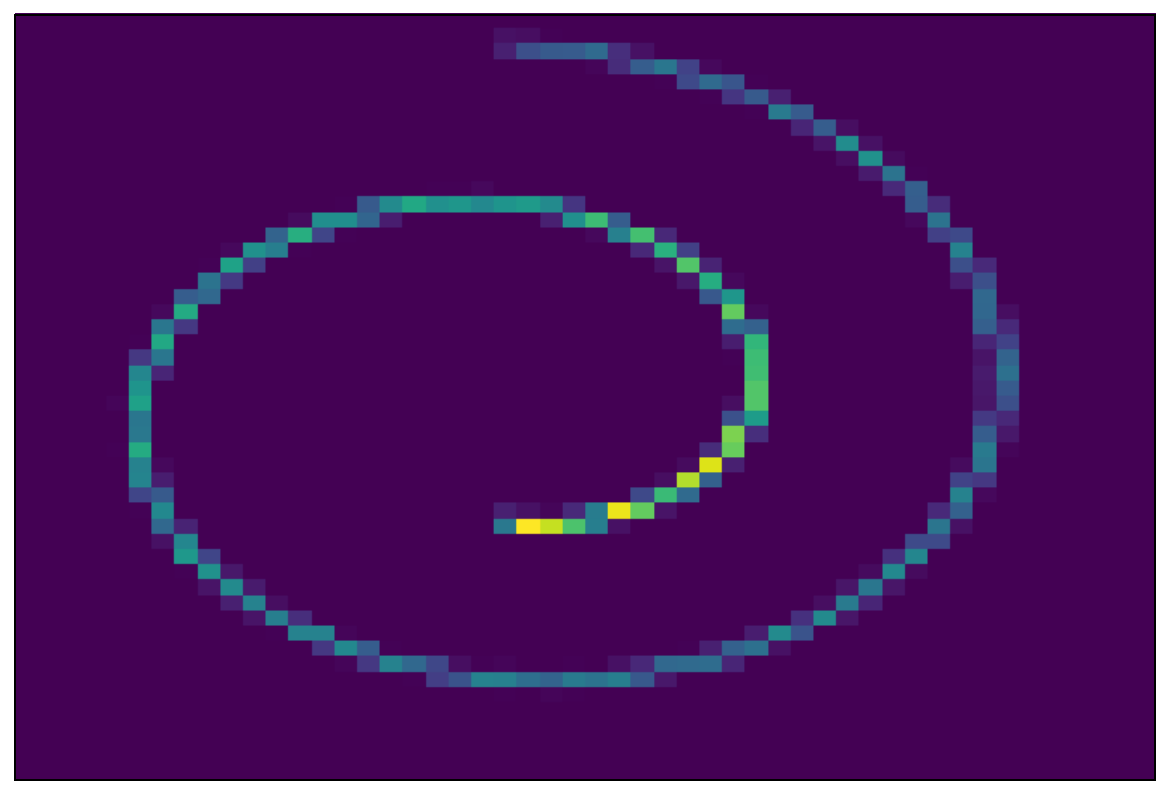}
}
\caption{Toy experiments with a 5-layer MLP (batch size of 2048). $\mathcal L_\mathrm{DT}$ drives flow map training toward a suboptimal solution, while $\mathcal L_\mathrm{ED}$ leads to a solution close to the ground truth.}
\label{fig:toy}
\vspace{-1em}
\end{wrapfigure}

To demonstrate this suboptimality of direct training, we conduct experiments on a toy dataset, as shown in \cref{fig:toy}. Compared to Eulerian distillation, which successfully learns the ground-truth distribution, direct training converges to a degenerate distribution, consistent with the proposition (see also \cref{appendix:add-expr}).

When a stop-gradient operation is applied to $\mathcal L_\mathrm{DT}$, it coincides with the consistency training objective $\mathcal L_\mathrm{CT}$ in \cref{eq:ct}. This objective can recover solutions satisfying the Eulerian equation, even when guided by conditional velocity~\citep{consistencymodel}. However, this holds only at the expectation level, and a small batch size can induce a residual gap (see \cref{proof:instability}).

\begin{proposition}\label{prop:instability} (Expectation-level Realization of Consistency Training) Let $L_*f_\theta=\partial_t f_\theta+v_t^*\cdot\nabla_x f_\theta$ denote the Eulerian operator in \cref{eq:eulerian} applied to $f_\theta$. The stationary condition of consistency training is
\begin{align}
    \mathcal S_\mathrm{CT} = L_*f_\theta + \mathbb E_{x|x_t}[\Delta v\cdot\nabla_x f_\theta] = 0,
\end{align}
when $\rho_t > 0$. This admits the desired flow map as a fixed-point solution, $L_*f_\theta = 0$, since the second term vanishes by $\mathbb E_{x|x_t}[\Delta v] = 0$ in expectation. However, under finite-batch estimation, the empirical residual induced by $\Delta v$ may remain, which can perturb the fixed-point training dynamics.
\end{proposition}

\begin{figure}[th]
\centering
\begin{minipage}{0.48\linewidth}
\begin{minipage}{0.05\linewidth}
    \centering
    \rotatebox{90}{\small{$\mathcal L_\mathrm{CT}$}}
\end{minipage}
\begin{minipage}{0.93\linewidth}
\subfloat{%
\includegraphics[width=0.3\linewidth]{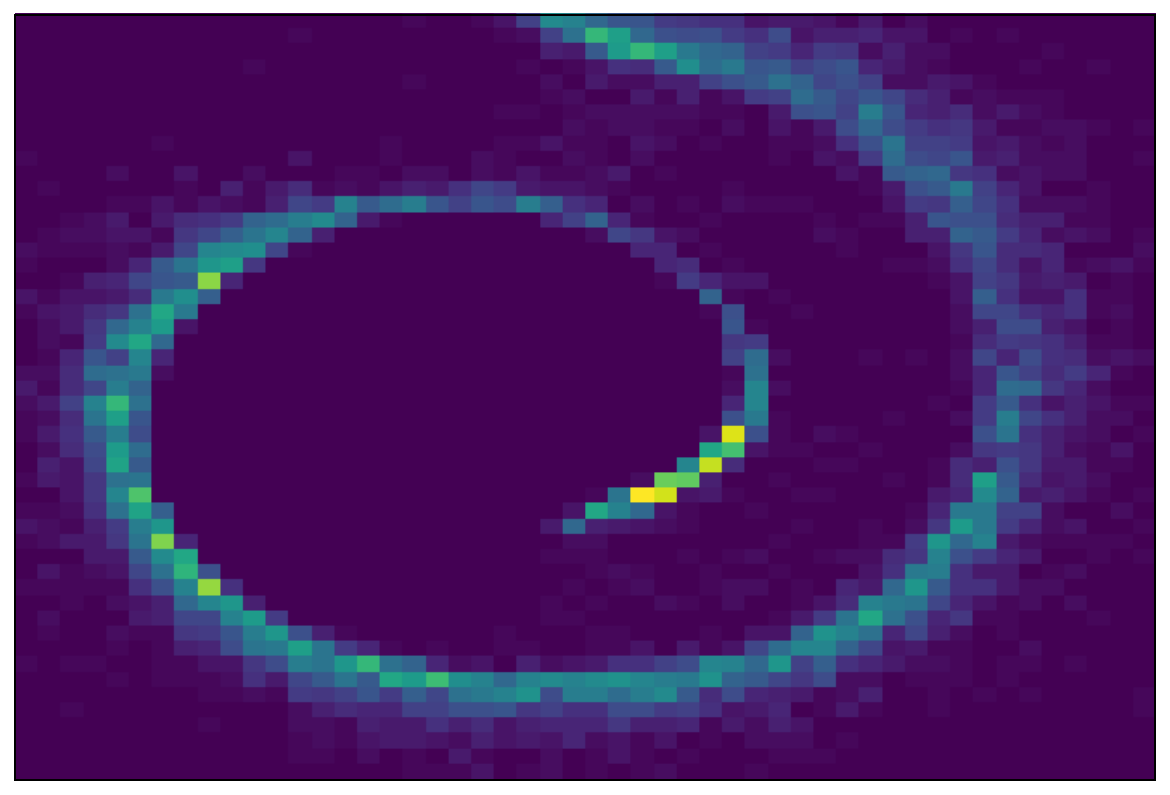}
}
\subfloat{%
\includegraphics[width=0.3\linewidth]{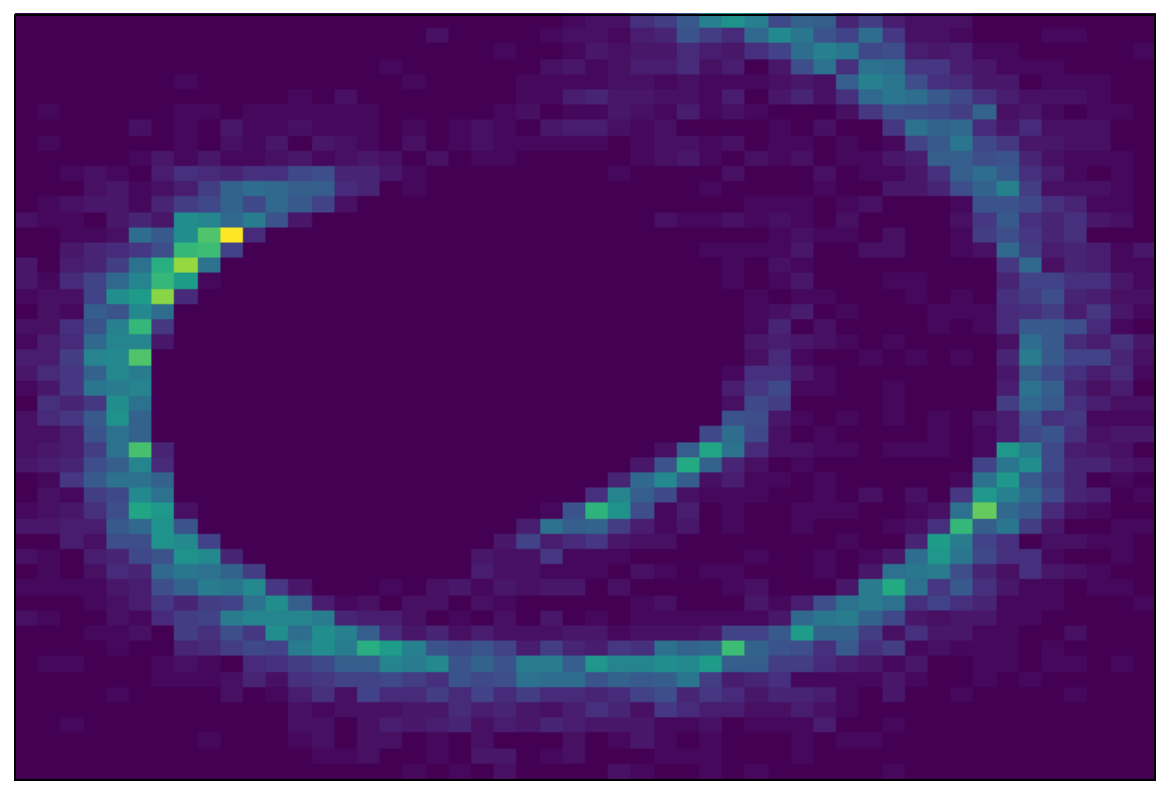}
}
\subfloat{%
\includegraphics[width=0.3\linewidth]{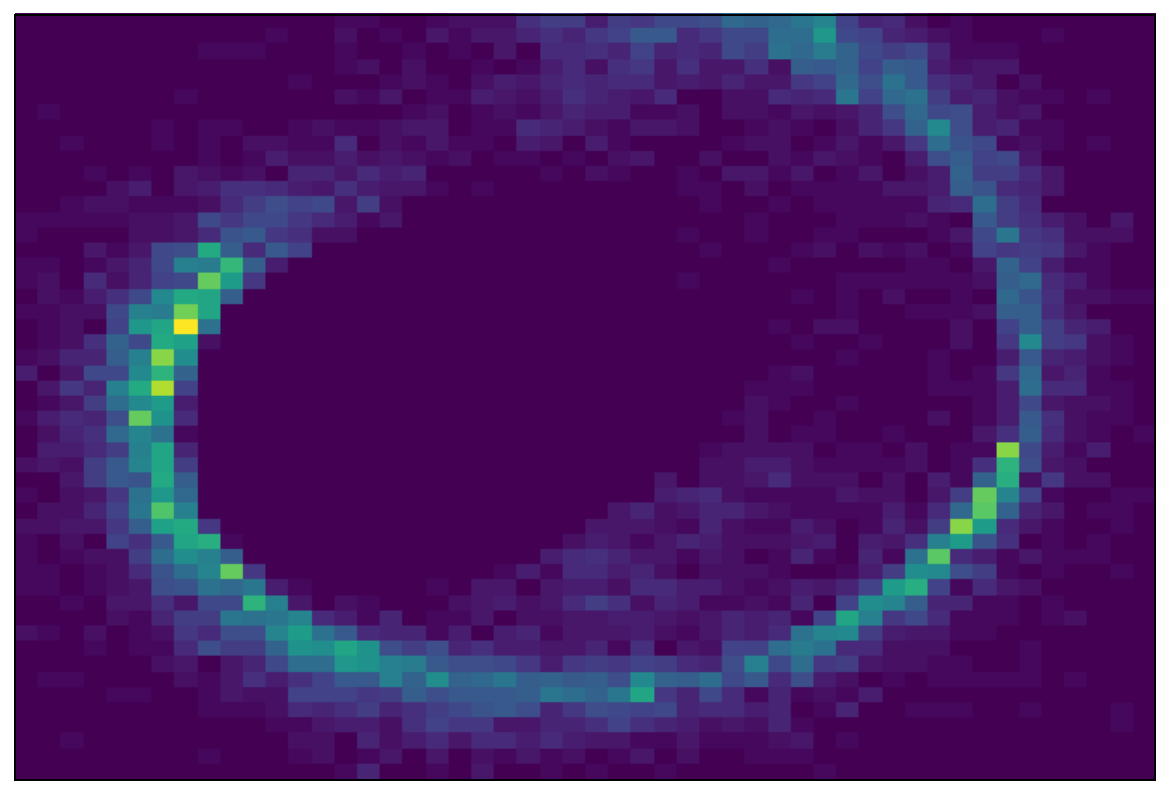}
}
\end{minipage}
\begin{minipage}{0.05\linewidth}
    \centering
    \rotatebox{90}{\small{$\mathcal L_\mathrm{ED}$}}
\end{minipage}
\begin{minipage}{0.93\linewidth}
\setcounter{subfigure}{0}
\subfloat[$B=2048$]{%
\includegraphics[width=0.3\linewidth]{figures/C_ED_2048_1000.pdf}
}
\subfloat[$B=512$]{%
\includegraphics[width=0.3\linewidth]{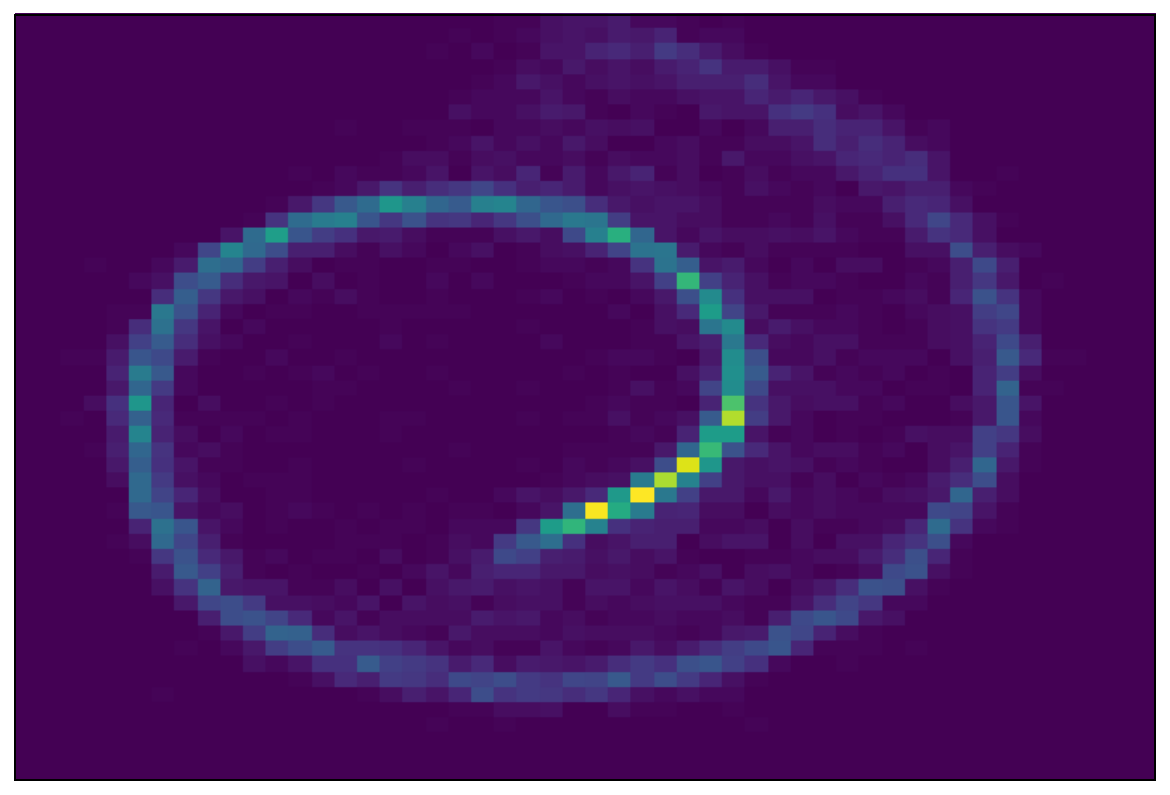}
}
\subfloat[$B=128$]{%
\includegraphics[width=0.3\linewidth]{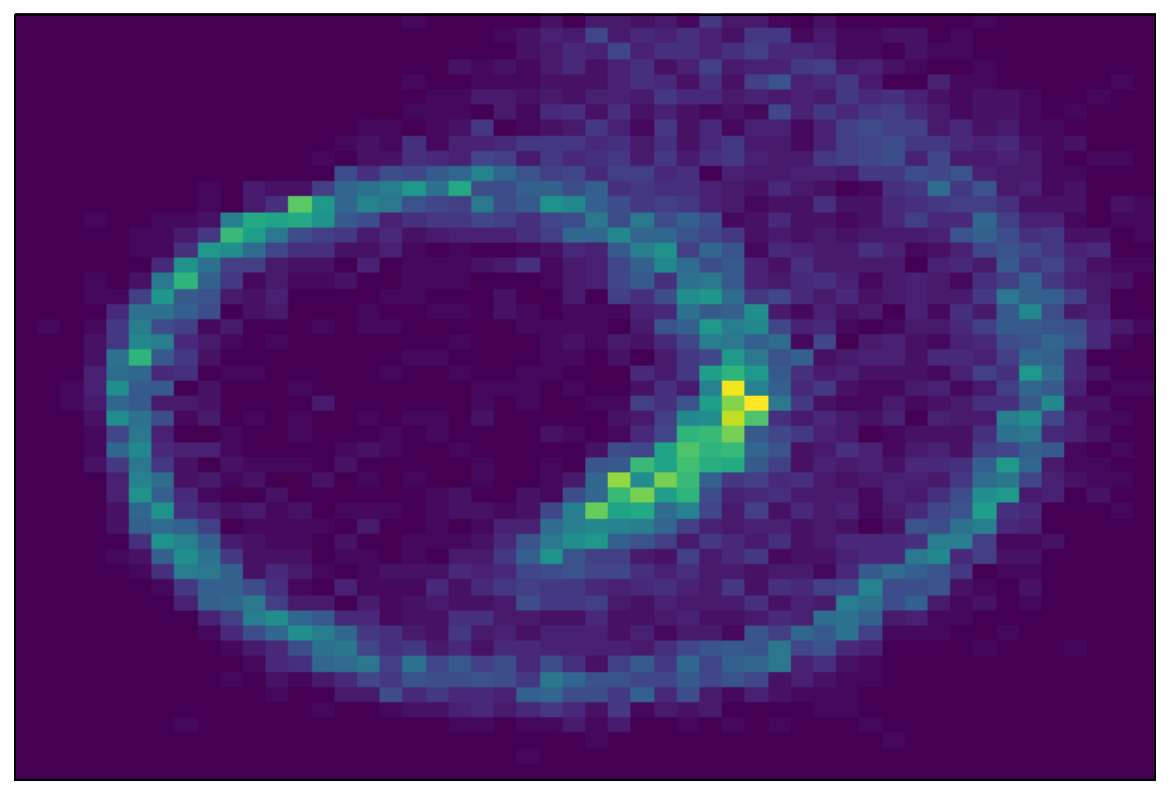}
}
\end{minipage}
\end{minipage}
\begin{minipage}{0.48\linewidth}
\centering
\subfloat[$\mathcal L_\mathrm{ED}$ over training steps.]{%
\includegraphics[width=0.90\linewidth]{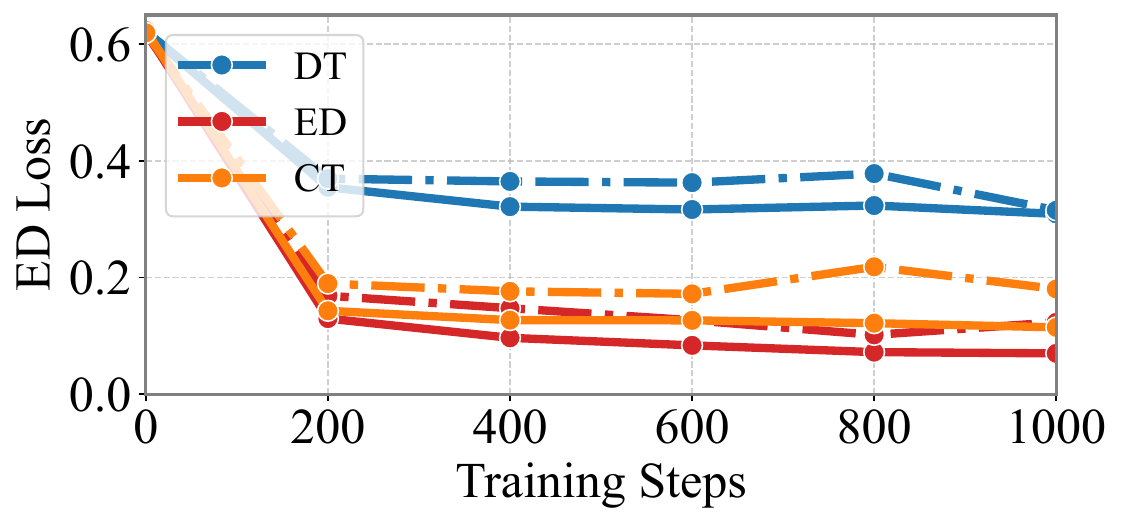}
\label{fig:toy-ed-curve}
}
\end{minipage}

\caption{(Left) Training results across batch sizes. The experiment follows \cref{fig:toy}. Consistency training $\mathcal L_\mathrm{CT}$ exhibits drift toward degenerate distributions as the batch size $B$ decreases. (Right) Estimating the optimality of flow maps using $\mathcal L_\mathrm{ED}$. Solid lines indicate a batch size of 2048, and dash-dot lines indicate 128.}
\label{fig:ct-batchsize}
\end{figure}

\cref{fig:ct-batchsize} shows drift toward a degenerate solution as the batch size decreases, \textit{which is consistent with the finite-batch residual in \cref{prop:instability}}. We also measure $\mathcal L_\mathrm{ED}$ for the toy dataset in \cref{fig:toy-ed-curve}, which serves as a proxy for estimating the optimality of flow maps. While it does not fully reflect generation performance, consistency training with small batch sizes and direct training yield higher $\mathcal L_\mathrm{ED}$ values, which can be interpreted as a signal of a degenerate solution. This helps explain why recent consistency models use large batch sizes.

From this perspective, both the bias term in \cref{prop:suboptimality} and the finite-batch residual in \cref{prop:instability} originate from the same conditional-marginal velocity mismatch. We therefore consider incorporating the marginal velocity as the underlying flow field to reduce this gap in the next section.

\textbf{Weight Initializations.} We further observe that consistency models with $s=0$ are sensitive to weight initialization, which can affect reproducibility. Recent approaches rely on initialization with pretrained flow models, where the network is already approximately aligned with target flows. This reduces certain early-stage instability, and is often referred to as a \textit{preconditioner}~\citep{meanflow}. We find that performance varies with the preconditioner, and reproducing results without known preconditioners can be challenging.

\begin{wraptable}{r}{0.5\linewidth}
\centering
\caption{Consistency training results under different preconditioners. \textit{Preconditioner FID} denotes the FIDs of pretrained networks for given ODE solver and sampling-step pairs. \textit{2-FID} denotes two-step FIDs of consistency models initialized from the corresponding preconditioner (details in \cref{detail:reproducing-ct}).}
\label{tab:reproducibility}
\resizebox{\linewidth}{!}{%
\begin{tabular}{lll}
\toprule
Preconditioner & Preconditioner FID$\downarrow$ & 2-FID$\downarrow$ \\
\midrule 
Multistep Baseline & 1.21 (UCGM-S, 30-step) & 2.52 \\
LightningDiT & 2.17 (Euler, 250-step) & 9.59 \\
In-house Flow Model & 2.41 (UCGM-S, 30-step) & 5.78 \\
w/o Preconditioner & \multicolumn{1}{c}{-} & 419.60 \\
\midrule
Reported Baseline & 1.21 (UCGM-S, 30-step) & 1.42 \\
\bottomrule
\end{tabular}
}
\vspace{-2em}
\end{wraptable}

As shown in \cref{tab:reproducibility}, we evaluated an open-source consistency model~\citep{ucgm} on ImageNet-1K $256\!\times\!256$. With the baseline preconditioner, the model achieves reasonable results but remains above the reported FID. However, with other pretrained flow models, the FIDs become worse.

The resulting few-step FIDs are inconsistent: stronger preconditioners do not necessarily yield better consistency models. Additionally, models diverge when initialized randomly, suggesting that the optimum is inaccessible in this case. 

As a complementary hypothesis, this can be understood from the stop-gradient (fixed-point) updates of consistency models: \textit{initialization matters as training is driven toward nearby fixed points rather than the global optimum}. The result of \cref{tab:reproducibility} suggests the possible existence of attracting basins under different initialization weights. Without the known preconditioner, the training becomes difficult to reproduce.

\textbf{Time Condition.} Some studies~\citep{meanflow, shortcutmodel} enable training from scratch without a preconditioner. The key difference is that they allow $s<t$, while others fix $s=0$. Intuitively, training long-range mappings is more challenging than short-range ones. Under the flow map representation in \cref{eq:gen-flowmap}, Eulerian distillation can be written, especially for linear interpolation, as
\begin{align}
    \mathcal L_\mathrm{ED} = \mathbb E\left[\left\|A''_{t,s}x_t + A'_{t,s}(v^*_t - F_\theta) - A_{t,s}\frac{dF_\theta}{dt}\right\|^2_2\right] = \mathbb E\left[\left\|v^*_t - F_\theta - (t - s)\frac{dF_\theta}{dt}\right\|^2_2\right].
\end{align}
We observe that $s\to t$ amplifies the flow matching term $v^*_t - F_\theta$ since $A''_{t,t}=A_{t,t}=0$, while $s\to 0$ amplifies the linearization term involving the JVP $dF_\theta/dt$. As the JVP is structurally more complex, we hypothesize that it may give rise to more undesired fixed points. This suggests that fixing $s=0$ makes optimization less stable, while relaxing to $s<t$ balances the terms and mitigates instability (see \cref{proof:linearization}).

\begin{figure}[t]
\centering
\subfloat[Consistency Model]{%
\includegraphics[width=0.23\linewidth]{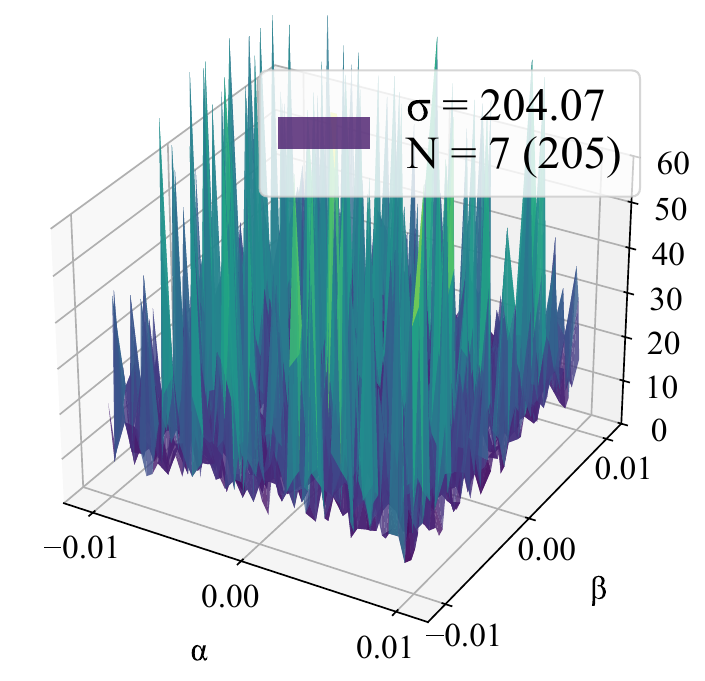}
}
\subfloat[Time Relaxation]{%
\includegraphics[width=0.23\linewidth]{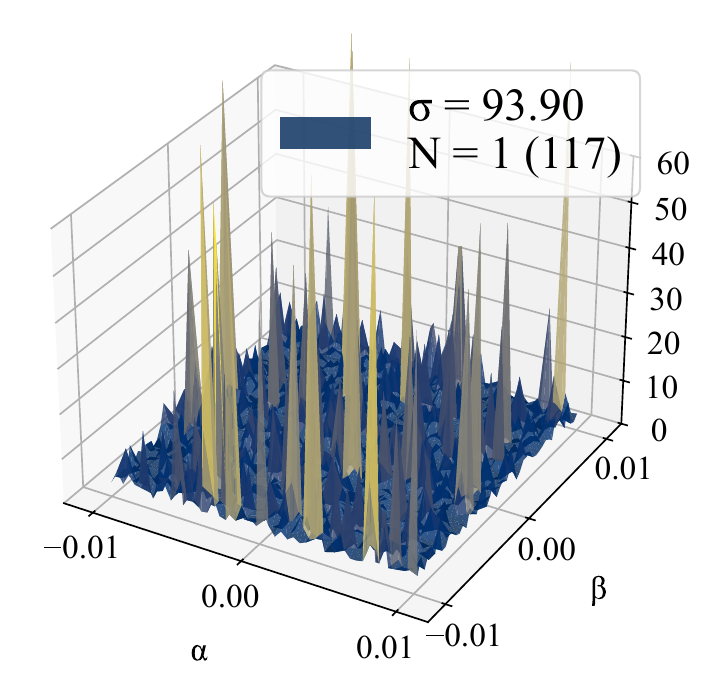}
}
\subfloat[Self-Distillation (ESD)]{%
\includegraphics[width=0.23\linewidth]{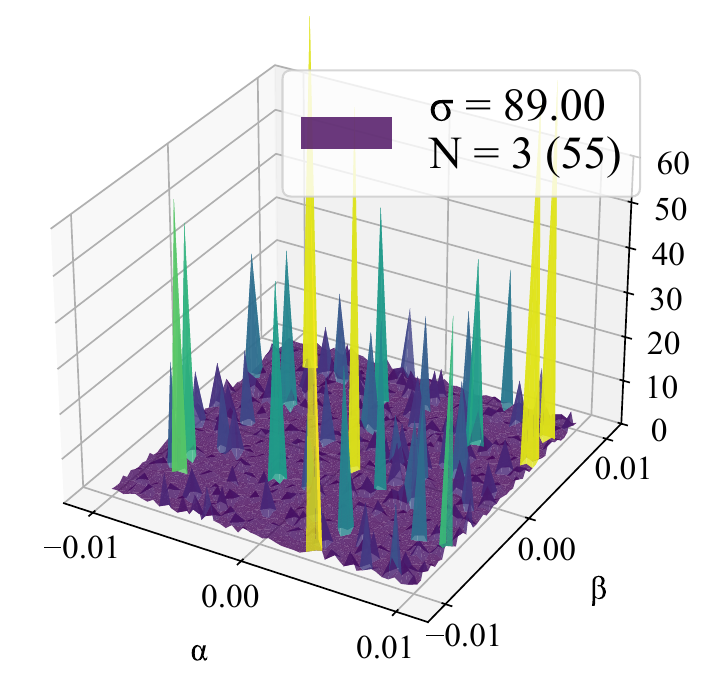}
\label{fig:landscape-sd}
}
\subfloat[iSD (Ours)]{%
\includegraphics[width=0.23\linewidth]{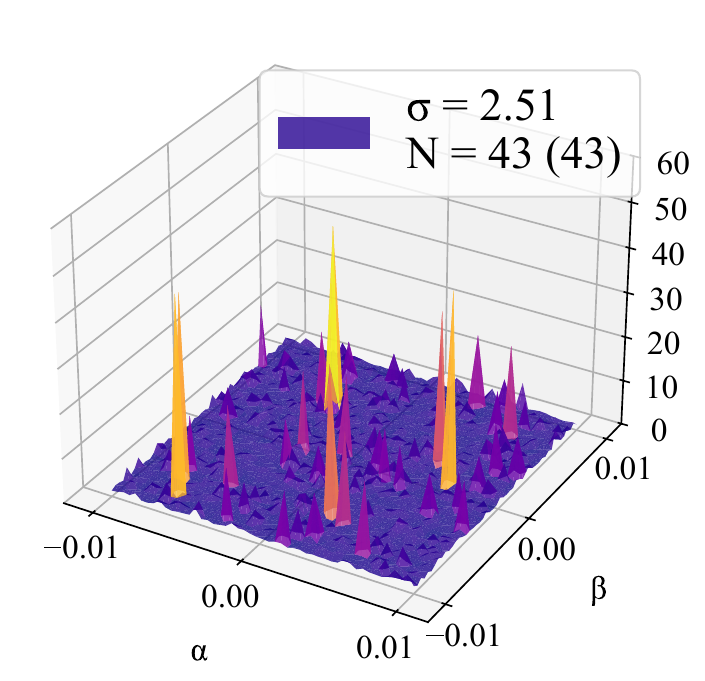}
}
\caption{Loss landscapes of four methods. The standard deviation $\sigma$ and the number of spikes $N$ serve as diagnostic proxies for training stability. iSD shows lower variance and fewer spikes under the common reference range. $\alpha$ and $\beta$ denote the top-2 eigenvectors of the Hessian on ImageNet-1K with DiT-B/4. $N$ denotes the number of samples outside each method's own empirical 95\% range. Values in parentheses report the number of samples exceeding the range defined by iSD, as a common reference (details in \cref{appendix:loss-landscape}).}
\label{fig:landscape-main}
\end{figure}

To support our hypothesis, we examine the loss landscapes of each objective, using their variance and number of spikes as proxies for training stability. As shown in \cref{fig:landscape-main}, time relaxation reduces loss variance and spikes, which can be interpreted as reduced risk of training divergence, supporting our claim. We also note that our method in the next section, iSD, produces even smoother landscapes.

\textbf{Summary.} Consistency training admits the desired flow map only in expectation due to conditional velocity guidance, and small batches can introduce a non-vanishing empirical residual. We observe that performance is sensitive to weight initialization, which can be attributed to fixed-point updates. Time-condition relaxation is associated with training stability and can enable random weight initialization, contributing to reproducibility.

These mechanisms provide an explanation for limited stability and reproducibility across initializations and training settings. Motivated by this observation, we seek to improve them by eliminating reliance on preconditioners through time-condition relaxation and reducing this residual via marginal velocity guidance.

\section{Method}
\label{section:contribution}

From \cref{tab:generalization}, we identify that Eulerian Self-Distillation~\citep[ESD;][]{boffi2025} uses marginal velocity under a relaxed time condition. This can make random initialization feasible and reduce reliance on preconditioners. However, ESD exhibits large gradient norms on high-dimensional data, destabilizing training.

Since ESD is not intended to explicitly address the gradient norm problem, we reformulate self-distillation following the consistency-based formulation of \citet{alignyourflow}. While this is an empirical optimization fix rather than a direct implication of \cref{prop:suboptimality,prop:instability}, we observe that this can reduce gradient norms in \cref{fig:gradnorm}. We call this approach \textit{improved Self-Distillation (iSD)}. We further show that it can be seamlessly combined with classifier-free guidance, yielding iSD-T with performance and reproducibility gains.

\subsection{improved Self-Distillation (iSD)}

\textbf{Eulerian Self-Distillation (ESD)} jointly trains $F_\theta(x_t; t, t)$ via flow matching and $f_\theta(x_t; t, s)$ via Eulerian distillation, guided by its own velocity:
\begin{align}\label{eq:otf-distill}
\mathcal L_\mathrm{SD} &= \mathbb E\left[\|\partial_tf_{\theta}(x_t; t, s) + F_\theta(x_t; t, t)\cdot\nabla_xf_\theta(x_t; t, s)\|^2_2\right].
\end{align}
Compared to direct training, this utilizes the model-induced marginal velocity $v^*_t(x_t)\approx F_\theta(x_t; t, t)$ instead of the per-sample conditional velocity $v_t(x_t|x)$, without requiring teacher networks. This encourages convergence to the marginal flow map and addresses suboptimality in \cref{prop:suboptimality} (see \cref{proof:compatibility}).

\textbf{Reformulation.} To enable ESD while incorporating design choices, we propose to reformulate self-distillation $\mathcal L_\mathrm{SD}$ using the flow map representation (Eq.~\ref{eq:gen-flowmap}). In this case, the guidance velocity $v_\theta(x_t; t) = F_\theta(x_t; t, t)$ corresponds to the marginal velocity under joint training of flow matching:
\begin{align}
    \mathcal L_\mathrm{SD} &= \mathbb E\left[\left\| A''_{t,s}x_t + A'_{t,s}(v_\theta(x_t; t) - F_\theta(x_t; t, s)) - A_{t,s}\frac{dF_\theta(x_t; t, s)}{dt}\right\|^2_2\right].
\end{align}
To reduce gradient norms, we consider two stop-gradient placements: only the spatial derivative and the full derivative. Both yield theoretically well-defined objectives admitting the desired fixed points, and we empirically select the latter for lower gradient norm and computational efficiency (see \cref{fig:gradnorm}):
\begin{align}\label{eq:sd-r}
    \mathcal L_\mathrm{SD\text{-}R} = \mathbb E\left[\|F_\theta(x_t; t, s) - \operatorname{sg}\left[F_\mathrm{tgt}\right]\|^2_2\right],\quad F_\mathrm{tgt} = F_\theta + A''_{t,s}x_t + A'_{t,s}(v_\theta - F_\theta) - A_{t,s}\frac{dF_\theta}{dt}.
\end{align}
Thus, our final objective of improved Self-Distillation follows:
\begin{align}\label{eq:isd}
    \mathcal L_\mathrm{iSD} = \lambda_1\mathcal L_\mathrm{CFM} + \lambda_2\mathcal L_\mathrm{SD\text{-}R},\qquad \mathcal L_\mathrm{CFM} = \mathbb E[\|F_\theta(x_t; t, t) - v_t(x_t|x)\|^2_2].
\end{align}
Consistency training can be viewed as a coupled fixed-point iteration with a parameterized target, whose contraction and global convergence are difficult to establish without additional assumptions. Instead, \cref{proof:compatibility} suggests that the objectives in \cref{eq:isd} are aligned and non-conflicting. We leave contraction and global convergence properties for future work.

\textbf{Comparison to Prior Work.} Compared to consistency models using the per-sample velocity for the flow map target~\citep{sct, meanflow}, iSD utilizes self-contained marginal velocity, which can reduce the mismatch introduced in \cref{prop:instability} and reduce loss variance (see \cref{proof:final-objective}). Recent work addresses related sources of instability from complementary angles. FACM~\citep{facm} introduces flow anchoring, which can be interpreted as a weak form of time-condition relaxation. iMF~\citep{imf} addresses conditional-velocity leakage by using a learned velocity in its JVP computation, which can be interpreted as reducing the conditional-marginal velocity mismatch (see \cref{appendix:comparison}).

\subsection{Implementation}

\textbf{Computing JVP.} To align with the marginal velocity, $dF_\theta/dt$ needs to follow $dF_\theta/dt = \partial_tF_\theta + v_\theta\cdot \nabla_x F_\theta$. This can be implemented using standard JVP APIs with inputs $(x_t, t, s)$ and tangents $(v_\theta(x_t; t), 1, 0)$. This differs from prior work~\citep{sct,meanflow}, which assumes tangents $(v_t(x_t|x), 1, 0)$.

Alternatively, we can approximate it using a finite difference. Given a small step $\epsilon$, we define displaced inputs as $\hat x_{t\pm\epsilon} = x_t \pm \epsilon\cdot v_\theta$, and approximate $dF_\theta/dt\approx [F_\theta(\hat x_{t+\epsilon}; t + \epsilon, s) - F_\theta(\hat x_{t-\epsilon}; t - \epsilon, s)] /(2\epsilon)$.

\textbf{Classifier-free Guidance} (CFG) is a widely used technique for improving diffusion models. However, since flow maps parameterize mappings rather than velocities, directly applying CFG does not guarantee mappings along the CFG velocity (see \cref{proof:cfg-fmm}).

We thus separate two extensions of iSD: Post-CFG and Pre-CFG. Post-CFG operates as a classical CFG:
\begin{align}\label{eq:post-cfg}
    \tilde f_\theta(x_t; t, s, c) = \nu^{-1}(A'_{t,s}x_t - A_{t,s} \tilde F_c),\quad \tilde F_c = F_\theta(x_t; t, s, \varnothing)  + \omega(F_\theta(x_t; t, s, c) - F_\theta(x_t; t, s, \varnothing)),
\end{align}
where $\varnothing$ is the null class label, $c$ is the class label, and $\omega$ is the guidance scale. Although this formulation is not guaranteed to follow the CFG velocity field, it can be readily applied after training.

To ensure that the flow map follows the CFG field, we adopt a training-time CFG formulation~\citep{meanflow}, Pre-CFG. This incorporates CFG directly by replacing flow matching with:
\begin{align}\label{eq:training-time-cfg}
    \mathcal L_\mathrm{CFM}^\omega = \mathbb E\left[\left\|F_\theta(x_t; t, t, c) - \tilde v_t(x_t|x)\right\|^2_2\right],\qquad \tilde v_t(x_t|x) = F_\varnothing + \omega(v_t(x_t|x) - F_\varnothing),
\end{align}
where $F_\varnothing = \operatorname{sg}[F_\theta(x_t; t, t, \varnothing)]$ for brevity. We then train the flow map using the following objective:
\begin{align}\label{eq:isd-t}
    \mathcal L_\mathrm{iSD\text{-}T} &= \lambda_1\mathcal L_\mathrm{CFM}^\omega + \lambda_2\mathcal L_\mathrm{SD\text{-}R}.
\end{align}
This objective encourages the flow map to align with the ground-truth CFG velocity (see \cref{proof:ttcfg}). We refer to this case as iSD-T (\textbf{T}raining-time CFG). We adopt iSD-T for class-conditional settings and vanilla iSD for class-unconditional settings. 

\section{Experiments}
\label{section:experiments}

\begin{figure}[t]
\centering
\subfloat[Design choices]{
\includegraphics[width=0.31\linewidth]{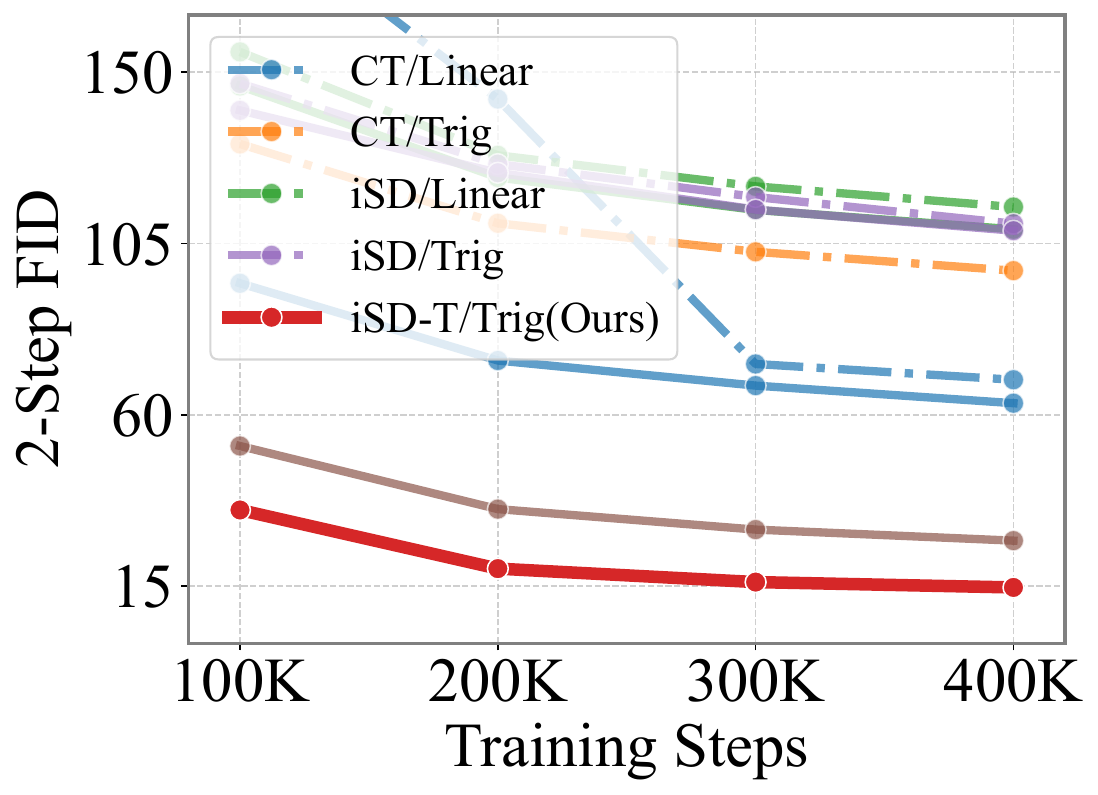}
\label{fig:interp-jvpop}
}
\subfloat[Post-CFG scales]{
\includegraphics[width=0.31\linewidth]{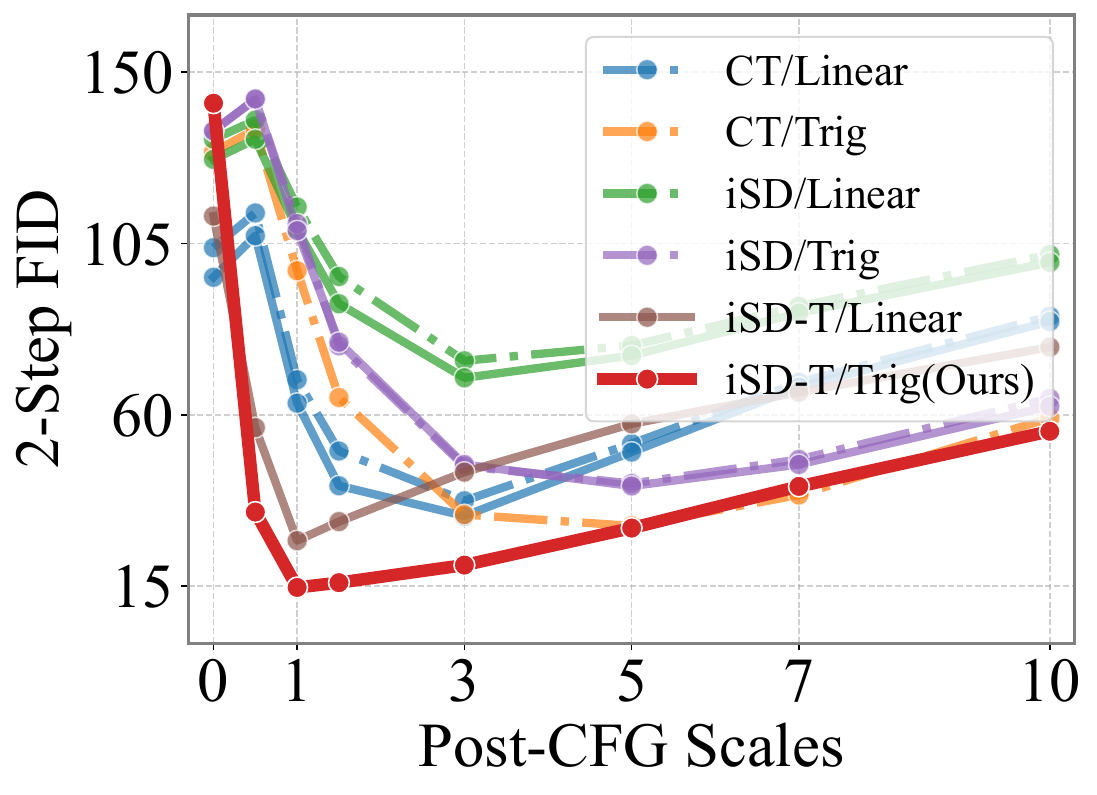}
\label{fig:postcfg}
}
\subfloat[Pre-CFG scales]{
\includegraphics[width=0.31\linewidth]{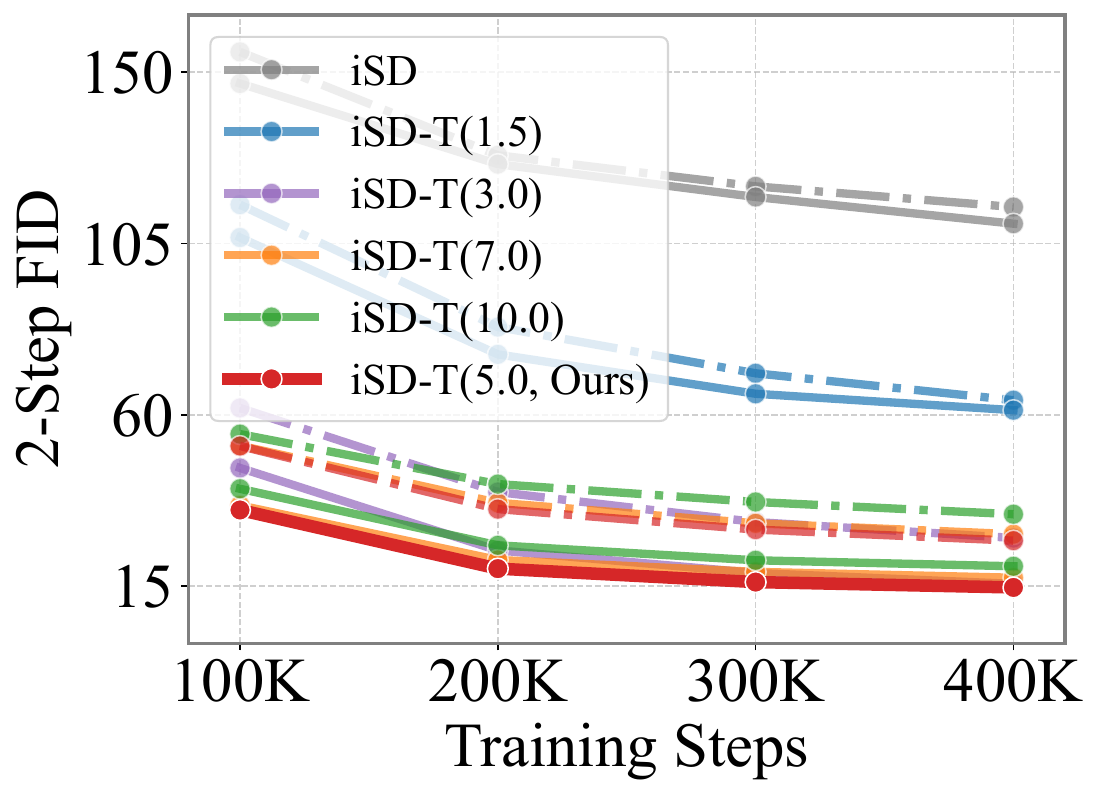}
\label{fig:precfg}
}
\caption{Design choices. (a) FIDs over training steps. Solid lines indicate the JVP approximation, and dash-dot lines indicate the exact JVP. (b) FIDs of Post-CFG over guidance scales. (c) FIDs of Pre-CFG over training steps. Solid lines indicate trigonometric interpolation, and dash-dot lines indicate linear.}
\label{fig:design-choices}
\end{figure}

\textbf{Experimental Settings.} To evaluate our method on image generation, we conduct experiments on the ImageNet-1K~\citep{imagenet} dataset. We use downsampled $32\!\times\!32\!\times\!4$ latent variables from $256\!\times\!256$ images encoded by a VAE~\citep{ldm}, and adopt a DiT~\citep{dit} in BF16 precision.

For unconditional settings, we validate our method on the CelebA-HQ~\citep{pggan} dataset, using the same latent variable settings as in the ImageNet experiments. Sample quality is measured with FID~\citep{fid} and Inception Score~\citep[IS;][]{NIPS2016_inceptionscore}. More details are provided in \cref{detail:implementation}.

We further evaluate our method on diffusion-based policy learning to assess the applicability beyond image generation. We adopt diffusion policy benchmarks as in \citet{chi2023diffusionpolicy}, and report task success rates. This is intended to examine the applicability rather than to establish state-of-the-art results (see \cref{appendix:policy-generation}).

\subsection{Ablation Study}

We conduct our ablation study on ImageNet-1K using DiT-B/4. The model is trained for 400K steps. We set $\omega = 5.0$ for $\mathcal L_\text{iSD-T}$ and use conditional velocity guidance for $\mathcal L_\text{CT}$. All weights are randomly initialized.

\begin{table}[th]
\centering
\caption{Ablation study on ImageNet with DiT-B/4. 2-step evaluation results with standard deviations $\sigma$.}
\label{tab:validation}
\begin{tabular}{lcc}
\toprule
Case & FID$\downarrow$ ($\pm\sigma$) & IS$\uparrow$ ($\pm\sigma$) \\
\midrule
Consistency Training & 120.49 ($\pm$ 3.71) & 9.69 ($\pm$ 0.47) \\
+ Joint training w/ $\mathcal L_\text{CFM}$ & 96.28 ($\pm$ 15.9) & 14.11 ($\pm$ 4.15) \\
+ Relaxing to $s<t$ & 69.18 ($\pm$ 2.84) & 21.45 ($\pm$ 1.21) \\
+ iSD (Eq.~\ref{eq:isd}) & 114.56 ($\pm$ 2.80) & 11.04 ($\pm$ 0.41) \\
+ JVP approximation & 108.70 ($\pm$ 2.83) & 11.80  \textbf{($\pm$ 0.36)} \\
+ iSD-T (Eq.~\ref{eq:isd-t}, $\omega=5.0$) & 26.91 ($\pm$ 0.50) & 74.70 ($\pm$ 2.43) \\
+ Trigonometric Interp. & \textbf{14.62 ($\pm$ 0.06)} & \textbf{184.53} ($\pm$ 0.77) \\
\bottomrule
\end{tabular}
\end{table}

\textbf{Key Factors.} We perform ablation experiments aimed at improving the stability and reproducibility of consistency training, as summarized in \cref{tab:validation}. We measure the variance of quantitative results across five runs with different random initializations, and report mean values.

We begin with consistency training $\mathcal L_\mathrm{CT}$ using linear interpolation. Joint training with flow matching improves results. Relaxing the time condition further improves results and reduces variance, which is consistent with our linearization cost hypothesis.

We observe that self-distillation alone underperforms, while incorporating Pre-CFG $\mathcal L_\text{iSD-T}$ accelerates convergence, improves performance and reproducibility. One possible explanation is that vanilla iSD may trade the finite-batch residual for network approximation error, which can propagate to flow map training. Pre-CFG amplifies the class component and the flow map is trained along this sharpened field, whose task-relevant signal could become more prominent relative to the propagated approximation error.

\begin{table}[th]
\centering
\caption{FIDs across design choices (2-step). The numbers in the header denote Post-CFG scales.}
\label{tab:design-choices}
\begin{tabular}{lllllll}
\toprule
Loss & Interp. & JVP & 1.0 & 3.0 & 5.0 & 7.0 \\
\midrule
$\mathcal L_\text{CT}$ & Linear & Exact & 69.18 & 37.39 & 52.41 & 68.46 \\
$\mathcal L_\text{iSD}$ & Linear & Exact & 114.56 & 74.17 & 78.20 & 88.48 \\
\midrule
$\mathcal L_\text{CT}$ & Linear & Approx. & 63.02 & 33.40 & 50.18 & 67.27 \\
$\mathcal L_\text{iSD}$ & Linear & Approx. & 108.70 & 69.69 & 75.53 & 86.72 \\
\midrule
$\mathcal L_\text{iSD-T}$ & Linear & Approx. & 26.91 & 44.92 & 57.60 & 65.92 \\
$\mathcal L_\text{iSD-T}$ & Trig & Approx. & \textbf{14.62} & 20.49 & 30.21 & 41.15 \\
\bottomrule
\end{tabular}
\end{table}

\textbf{Interpolation.} As shown in \cref{fig:design-choices} and \cref{tab:design-choices}, linear interpolation tends to perform better under conditional velocity guidance, whereas trigonometric interpolation achieves lower FIDs under iSD-T. We observe varying performance across settings and tasks (\cref{appendix:policy-generation}), leaving further analysis to future work.

\textbf{JVP Operation.} We observe that JVP approximation often yields better empirical results than exact computation, as illustrated in \cref{fig:interp-jvpop}. Under BF16 precision, it tends to be more robust, as the exact one is sensitive to numerical precision. The VRAM requirement of the training decreases from 25GB to 18GB on iSD-T, and the training speed improves from 1.48 steps/sec to 4.04 steps/sec on an A100 GPU.

\textbf{Classifier-free Guidance.} As shown in \cref{fig:precfg} and \cref{tab:design-choices}, incorporating Pre-CFG improves FIDs compared to the vanilla $\mathcal L_\text{iSD}$. It also outperforms Post-CFG, which deviates from the intended CFG field, whereas Pre-CFG aligns the flow map with the CFG velocity during training. Combining both often degrades performance, as it results in applying the guidance twice, overly skewing the trajectory. We report the gradient norm and loss curves of iSD-T in \cref{fig:loss-curve}.

\begin{table}[th]
\centering
\caption{Comparison on ImageNet with DiT-B/4. 2-step results with standard deviations $\sigma$ under a controlled setup: the backbone, data, optimization, and training budgets are fixed, while the only difference lies in the training objective. Each method is evaluated using the CFG scale reported as optimal in its work.}
\label{tab:imagenet-reprod}
\begin{tabular}{llr}
\toprule
Case & \multicolumn{1}{c}{FID$\downarrow$ ($\pm\sigma$)} & \multicolumn{1}{c}{IS$\uparrow$ ($\pm\sigma$)} \\
\midrule
MeanFlow ($\omega = 3.0$) & \textbf{11.48} ($\pm$ 1.68) & 167.60 ($\pm$ 20.4) \\
Shortcut Model ($\omega = 1.5$) & 87.52 ($\pm$ 24.3) & 14.94 ($\pm$ 3.68) \\
FACM ($\omega = 1.75$) & 25.52 ($\pm$ 1.73) & 65.42 ($\pm$ 4.84) \\
\midrule
\textbf{iSD-T (Ours, $\omega = 5.0$)} & 14.62 ($\pm$ \textbf{0.06}) & \textbf{184.53 ($\pm$ 0.77)} \\
\bottomrule
\end{tabular}
\end{table}

\textbf{Reproducibility.} We examine the variance of quantitative results to assess reproducibility. In \cref{tab:imagenet-reprod}, our iSD-T exhibits lower variance across runs compared to the baselines. From a performance perspective, we observe an FID gap with MeanFlow. Thus, a main advantage of iSD-T is more reproducible training outcomes with comparable sample quality. See \cref{appendix:add-expr} for further ablation studies.

\subsection{Comparison with Prior Work}

\textbf{Unconditional Generation.} We compare our method with prior approaches on CelebA-HQ $256\!\times\!256$ under a class-unconditional setting, using vanilla iSD without CFG. As shown in \cref{tab:quant-celebahq}, iSD achieves comparable performance in both few-step and multistep generation.

\begin{table}[th]
\centering
\begin{minipage}{0.50\linewidth}
\centering
\caption{Comparison on unconditional CelebA-HQ.}
\label{tab:quant-celebahq}
\resizebox{\linewidth}{!}{%
\begin{tabular}{lcc}
\toprule
Model & 4-FID ($\downarrow$) & 128-FID ($\downarrow$) \\
\midrule
\multicolumn{3}{l}{Distillation-based Methods} \\
\midrule
CD~\citep{consistencymodel} & 39.6 & 59.5 \\
Reflow~\citep{rf} & 18.4 & 16.1 \\
\midrule
\multicolumn{3}{l}{Training from Scratch} \\
\midrule
Flow Matching~\citep{flowmatching}\hspace*{-1.5em} & 63.3 & 7.3 \\
CT~\citep{consistencymodel} & 19.0 & 53.7 \\
Shortcut Model~\citep{shortcutmodel}\hspace*{-1.5em} & 13.8 & \textbf{6.9} \\
\midrule
\textbf{iSD (Ours)} & \textbf{11.3} & 7.1 \\
\bottomrule
\end{tabular}
}
\end{minipage}\hspace*{0.01\linewidth}
\begin{minipage}{0.48\linewidth}
\centering
\caption{Comparison on ImageNet-1K. iSD-T uses linear interpolation with a guidance scale of 5.0.}
\label{tab:imagenet-ditxl}
\resizebox{\linewidth}{!}{%
\begin{tabular}{lcc}
\toprule
Model & NFE ($\downarrow$) & FID ($\downarrow$) \\
\midrule
iCT~\citep{ict} & 2 & 20.3 \\
Shortcut Model~\citep{shortcutmodel} & 1 & 10.6 \\
UCGM~\citep{ucgm} & 1 & 2.10 \\
IMM~\citep{imm} & 2  & 7.77 \\
MeanFlow~\citep{meanflow} & 2 & 2.20 \\
FACM~\citep{facm} & 2 & 1.32 \\
\midrule
\textbf{iSD-T (Ours, SD-VAE)} & 2 & 2.76 \\
\textbf{iSD-T (Ours, VA-VAE)} & 2 & 2.26 \\
\bottomrule
\end{tabular}
}
\end{minipage}
\end{table}

\textbf{Conditional Generation.} We further compare consistency-based approaches on ImageNet-1K under class-conditional settings, as shown in \cref{tab:imagenet-reprod}. Using DiT-B/4, iSD-T achieves performance comparable to the listed methods while exhibiting lower variance, indicating improved reproducibility.

\begin{wrapfigure}{r}{0.48\linewidth}
\vspace{-1.2em}
\includegraphics[width=\linewidth]{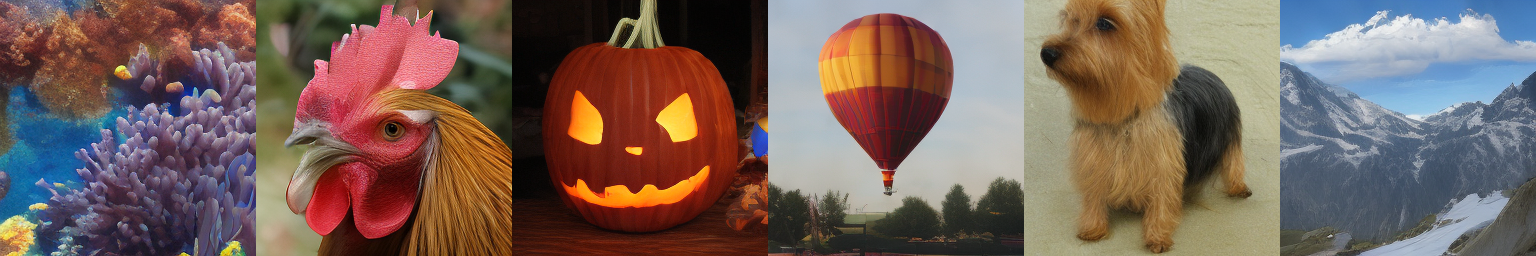}
\caption{2-step results trained on ImageNet, iSD-T.}
\vspace{-2em}
\end{wrapfigure}

To examine scalability, we evaluate iSD-T with DiT-XL. In \cref{tab:imagenet-ditxl}, iSD-T maintains stable training behavior and comparable sample quality at this scale.

We note that achieving strong FID scores at this scale requires incorporating performance-oriented conditioning techniques, such as conditioning on a guidance scale or variations in latent representation. These choices are orthogonal to our analysis and are not required for stability. Importantly, applying such conditioning does not reintroduce instability, indicating that iSD-T provides stable training even when combined with performance-oriented augmentations. This comparison assesses scalability under the combined setting, and the results reflect this configuration rather than the iSD objective alone (see \cref{detail:implementation}).

\subsection{Diffusion-based Policy Learning}

\begin{table}[th]
\centering
\caption{Success rates ($\uparrow$) of robotic control tasks.}
\label{tab:diffusion-policy}
\begin{tabular}{lccc}
\toprule
Model & NFE ($\downarrow$) & Push-T & Transport \\
\midrule
Diffusion Policy~\citep{chi2023diffusionpolicy} & 100 & 0.95 & 1.00 \\
    & 1 & 0.12 & 0.00 \\
\midrule
Shortcut Model~\citep{shortcutmodel} & 1 & 0.87 & 0.80 \\
MeanFlow~\citep{meanflow} & 1 & 0.85 & 0.92 \\
\midrule
\textbf{iSD (Ours)} & 2 & 0.92 & 1.00 \\
    & 1 & 0.86 & 0.94 \\
\bottomrule
\end{tabular}
\end{table}

\begin{wrapfigure}{r}{0.50\linewidth}
\centering
\vspace{-1.2em}
\includegraphics[width=\linewidth]{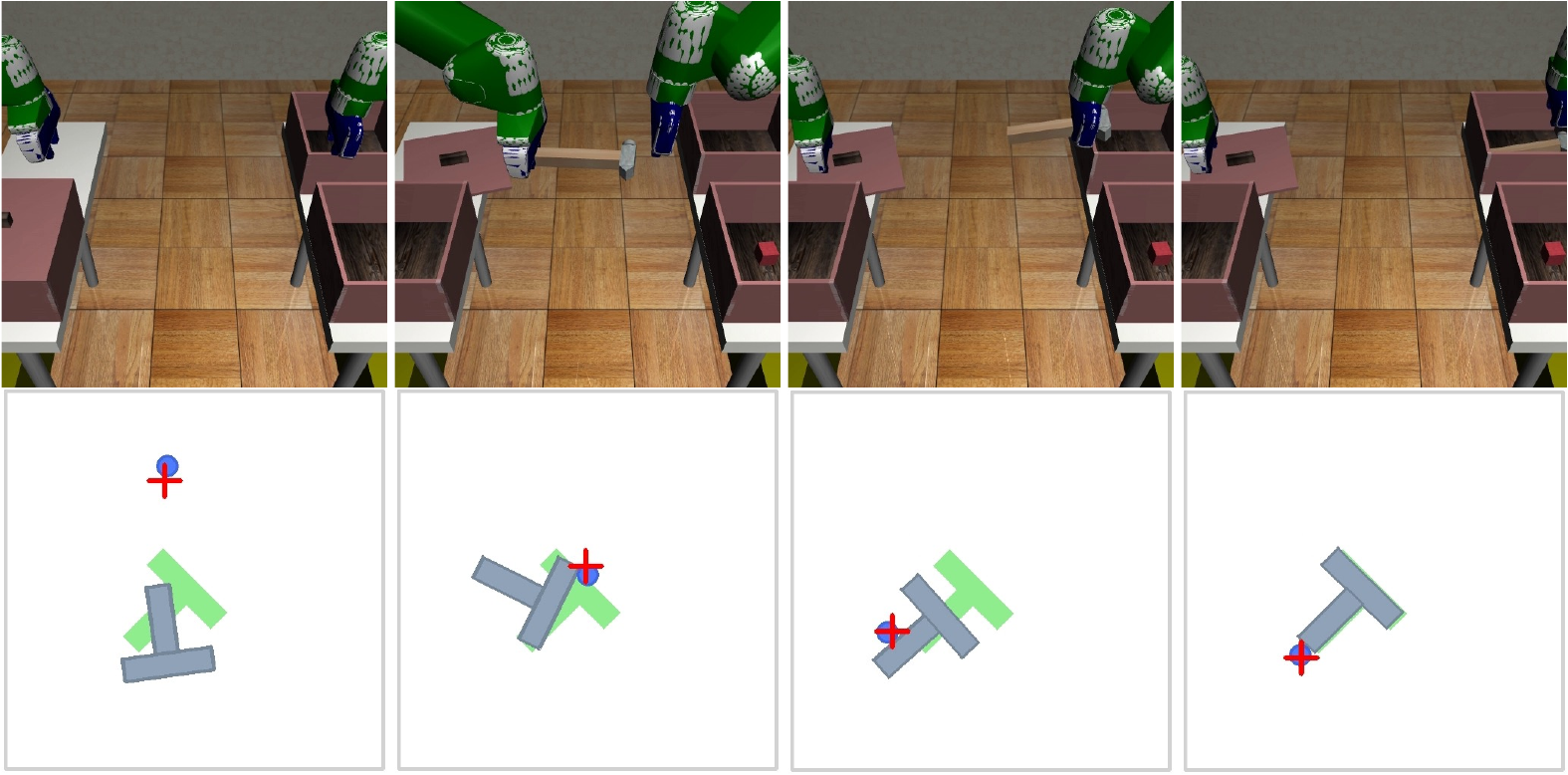}
\captionof{figure}{Simulation results of the iSD policy with 2-NFE on Transport (top) and Push-T (bottom) tasks.}
\vspace{-1em}
\end{wrapfigure}

To assess the applicability of iSD beyond image generation, we adopt it as a diffusion policy objective in robotic control tasks. Experimental settings follow prior work on diffusion-based policies~\citep{chi2023diffusionpolicy} for transformer and state-based architectures. 

As shown in \cref{tab:diffusion-policy}, iSD with linear interpolation yields performance comparable to other few-step methods, under the same evaluation protocol. With 2-step sampling, iSD achieves success rates comparable to those of diffusion policies that require more sampling steps. These results indicate that iSD can serve as a few-step alternative to diffusion-based objectives in policy learning.

\section{Conclusion}
In this work, we conducted a theoretical examination of consistency-based generative models from a flow map perspective, clarifying the sources of suboptimal convergence and training instability that can limit reproducibility. This analysis provides an explanation of how objective design choices influence convergence behavior in practice. Building on these insights, we revisited self-distillation. To reduce its gradient norm problem, we empirically reformulated the self-distillation objective, yielding improved Self-Distillation (iSD). We further integrated training-time classifier-free guidance into iSD, yielding iSD-T with improved performance and reproducibility. Finally, we demonstrated that iSD is applicable to diffusion-based policy learning beyond image generation, without reliance on a pretrained diffusion model.

\subsubsection*{Broader Impact Statement}
This paper presents work aimed at advancing the field of machine learning by improving the theoretical understanding and training stability of consistency-based generative models. The contributions focus on analysis and objective reformulation, without introducing new model architectures, expanding model capacity, or relying on additional data. As such, the potential ethical and societal impacts of this work are aligned with those already well established in generative modeling, including concerns related to data bias, misuse of generated content, and downstream applications. Improving training stability and reproducibility may contribute positively to research efficiency and reduce computational waste, but the work does not directly address broader societal risks, which remain important considerations for future research.

\subsubsection*{Acknowledgments}

This work was supported by the IITP grants (RS-2021-II211343: AI Graduate School Program (Seoul National University) (5\%), RS-2024-00509257: Global AI Frontier Lab (15\%), and RS-2025-25442338: AI Star Fellowship Support Program (30\%)) and the NRF grant (RS-2024-00405857 (50\%)) funded by the Korea government (MSIT). We gratefully acknowledge the generous support of the NVIDIA Academic Grant Program for providing the A100 GPUs and the NVIDIA CUDA Toolkit used in this research.

\bibliography{main}
\bibliographystyle{tmlr}

\newpage
\appendix
\section{Theoretical Analysis of Flow Map Models}
\subsection{Mean Collapse of Diffusion and Flow Matching Models}
\label{proof:mean-collapse}
\textbf{Posterior Distribution.} First, consider the data distribution $p_X(x)=\mathcal N(\mu_X, \sigma_X^2)$ and the interpolation $x_t=\alpha_t x + \sigma_t z$. The conditional distribution is given by $p(x_t=y|x)=\mathcal N(y;\alpha_t x, \sigma_t^2)$. By Bayes' rule,
\begin{align*}
p(x|x_t=y) 
&\propto p(x_t=y|x)p(x) \\
&= \mathcal N(y; \alpha_tx, \sigma_t^2)\mathcal N(x; \mu_X, \sigma_X^2)\\
&\propto\exp\left(- \frac{(y - \alpha_tx)^2}{2\sigma_t^2}-\frac{(x - \mu_X)^2}{2\sigma_X^2}\right) \\
&=\exp\left(-\left(\frac{1}{2\sigma_X^2} + \frac{\alpha_t^2}{2\sigma_t^2}\right)x^2 + \left(\frac{\mu_X}{\sigma_X^2}+ \frac{\alpha_ty}{\sigma_t^2}\right)x - \left(\frac{\mu_X^2}{2\sigma_X^2} + \frac{y^2}{2\sigma_t^2}\right)\right).
\end{align*}
This can be organized as a Gaussian with a closed form $p(x|x_t = y) = \mathcal N(\mu_{x|y,t}, \sigma_{x|y,t}^2)$ where
\begin{align*}
\mu_{x|y,t} = \frac{\alpha_t\sigma^2_Xy + \mu_X\sigma_t^2}{\sigma_t^2 + \sigma^2_X\alpha_t^2},\quad \sigma_{x|y,t}^2 = \frac{\sigma_X^2\sigma_t^2}{\sigma_t^2 + \sigma^2_X\alpha_t^2}.
\end{align*}
Extending the data distribution to a mixture of Gaussians $p_X(x)=\sum_i \pi_i \mathcal N(x;\mu_i,\sigma_i^2)$, we introduce the latent variable $\pi$ for handling $\pi_i$:
\begin{align*}
p(\pi=i)=\pi_i,\quad p(x|\pi=i)=\mathcal N(x;\mu_i,\sigma_i^2).
\end{align*}
Then, the marginal distribution $p(x_t=y|\pi=i)$ can be expressed as
\begin{align*}
p(x_t=y|\pi=i)
&=\int p(x_t=y|x)p(x|\pi=i)dx \\
&= \int \mathcal N(y; \alpha_tx, \sigma_t^2)\mathcal N(x; \mu_i, \sigma_i^2)dx \\
&= \mathcal N(y; \alpha_t\mu_i, \alpha_t^2\sigma_i^2 + \sigma_t^2).
\end{align*}
We define the responsibilities $r_{i,t}(y)$ as the posterior distribution
\begin{align*}
r_{i,t}(y)
= p(\pi=i|x_t=y) 
= \frac{p(x_t=y|\pi=i)p(\pi=i)}{\sum_jp(x_t=y|\pi=j)p(\pi=j)}
= \frac{\pi_i\mathcal N(y; \alpha_t\mu_i, \alpha_t^2\sigma_i^2 + \sigma_t^2)}{\sum_j\pi_j\mathcal N(y; \alpha_t\mu_j, \alpha_t^2\sigma_j^2 + \sigma_t^2)}.
\end{align*}
Therefore, the posterior distribution $p(x|x_t=y)$ is
\begin{align*}
p(x|x_t=y) &= \sum_ip(\pi=i|x_t=y)p(x|x_t=y,\pi=i) \\
&= \sum_ir_{i,t}(y)\mathcal N(x; \mu_{x|i,y,t}, \sigma^2_{x|i,y,t}) \\
&\text{where } \mu_{x|i,y,t} = \frac{\alpha_t\sigma^2_iy + \mu_i\sigma_t^2}{\sigma_t^2 + \sigma^2_i\alpha_t^2},\ \sigma_{x|i,y,t}^2 = \frac{\sigma_i^2\sigma_t^2}{\sigma_t^2 + \sigma^2_i\alpha_t^2}.
\end{align*}
In particular, we observe that $\mu_{x|i,y,1} = \mu_i,\ \sigma^2_{x|i,y,1} = \sigma_i^2$ and $r_{i,1}(y) = \pi_i$ when $\alpha_1 = 0$.

\textbf{One-step Generation.} Under the linear trajectory $x_t=(1-t)x + tz$, the conditional velocity is $v_t(x_t|x) = z - x$. Thus, one-step generation is defined by
\begin{align*}
f_F(x_t; t) &= x_t - tF^*(x_t; t) = x_t - t\,\mathbb E_{x|x_t}[v_t(x_t|x)] \\
&= \mathbb E_{x,z|x_t}\left[x_t - t(z-x)\right] \\
&= \mathbb E_{x|x_t}[x].
\end{align*}
In the unimodal Gaussian case, $\mathbb E_{x|z}[x] = \mu_{x|z,1} =\mu_X$ and the one-step samples generated from $t=1$ collapse to the mean of the data distribution. Similarly, in the mixture of Gaussians case, one-step generated samples collapse to the mixture mean.
\begin{align*}
f_F(z) = \mathbb E_{x|z}[x] = \sum_i r_{i,1}(z)\,\mu_{x|i,z,1} = \sum_i \pi_i\,\mu_i = \mu_X.
\end{align*}
Thus, one-step generation collapses to the data mean $\mu_X$ regardless of the input. \qed

\subsection{Injectivity of Flow Map}
\label{proof:injectivity}

Since the marginal velocity is assumed to be Lipschitz continuous, the Picard-Lindel\"of theorem guarantees a unique solution to the ODE $dx_t = v^*_t(x_t)dt$ for any initial value. The non-crossing property of trajectories follows directly, since any crossing at the same time would contradict uniqueness. Thus, since the flow map is formulated as the solution of the ODE initialized at $x_t$, it is well-defined.

For injectivity, suppose that $f(x; t, s) = f(y; t, s)$ for some $x$ and $y$. By the uniqueness guaranteed by the Picard-Lindel\"of theorem, the two trajectories must coincide along the interval. In particular, $f(x; t, t) = f(y; t, t)$. By the boundary condition of the flow map, $x=y$ and thus the flow map is injective. \qed

\subsection{Eulerian Equation and Uniqueness of the Flow Map}
\label{proof:eulerian-eqn}
Suppose the ground-truth flow map is defined as
\begin{align*}
f^*_{t,s}(x_t) = x_t + \int_t^sv_\tau^*(x_\tau)d\tau = x_s.
\end{align*}
By construction, the identity mapping $f^*_{t,s}(f^*_{s,t}(x_s)) = x_s$ holds. Differentiating both sides w.r.t. $t$ yields
\begin{align*}
\frac{d}{dt}f^*_{t,s}(f^*_{s,t}(x_s)) = \partial_tf^*_{t,s}(f^*_{s,t}(x_s)) + \partial_tf^*_{s,t}(x_s)\cdot\nabla_xf^*_{t,s}(f^*_{s,t}(x_s))= \frac{d}{dt}x_s = 0.
\end{align*}
Using $f^*_{s,t} (x_s)=x_t$ and $\partial_t f^*_{s,t} (x_s)=\partial_t x_t = v^*_t(x_t)$, we obtain the Eulerian equation:
\begin{align*}
\frac{d}{dt}f^*_{t,s}(x_t) = \partial_tf^*_{t,s}(x_t) +v^*_t(x_t)\cdot\nabla_xf^*_{t,s}(x_t) = 0.
\end{align*}
Suppose a trainable network $f_\theta(x; t, s) =  f^\theta_{t,s}(x)$ is sufficiently differentiable, continuous in $x, t, s$, Lipschitz continuous in $x$, and satisfies the boundary condition $f^\theta_{s,s}(x)=x$ for all $s$. If $f_\theta$ satisfies the Eulerian equation, $f^\theta_{t,s}$ remains constant along the characteristic curve induced by $v^*_t(x_t)$. 

Let $\chi_\tau$ denote the characteristic curve defined on $[s, t]$ by $\chi_t = x$ and $\chi'_\tau = v^*_\tau(\chi_\tau)$. Along this curve, $f^\theta_{\tau, s}(\chi_\tau)$ is constant and evaluating at $\tau=t$ and $\tau=s$ yields
\begin{align*}
    f^\theta_{t,s}(x) = f^\theta_{t,s}(\chi_t) =  f^\theta_{s,s}(\chi_s) = \chi_s = f^*_{t,s}(x),
\end{align*}
since $f^*_{t,s}$ generates the characteristic curve $\chi_\tau$ by its definition. Thus, the learned mapping coincides with the exact flow map. \qed

\subsection{Flow Map Formulation and Interpolation Assumption}
\label{proof:interpolation}
We begin by explicitly deriving the one-step Euler solution in \cref{eq:gen-flowmap}. Supposing $x_t = \alpha_t x + \sigma_t z$, the velocity $v^*_t(x_t) = \mathbb E_{x,z|x_t}[\alpha'_tx + \sigma'_t z]$ gives:
\begin{align*}
    v^*_t(x_t) &= \alpha'_t\mathbb E_{x|x_t}[x] + \sigma'_t\mathbb E_{z|x_t}[z] \\
    &= \alpha'_t\mathbb E_{x|x_t}[x] + \frac{\sigma'_t}{\sigma_t} (x_t - \alpha_t\mathbb E_{x|x_t}[x]), \\
    \mathbb E_{x|x_t}[x] &= \left(\alpha'_t - \frac{\sigma'_t}{\sigma_t}\alpha_t\right)^{-1}\left(v^*_t(x_t) - \frac{\sigma'_t}{\sigma_t}x_t\right) \\
    &= \frac{\sigma'_tx_t - \sigma_tv^*_t(x_t)}{\alpha_t\sigma'_t - \sigma_t\alpha'_t} = \nu_t^{-1}(\sigma'_tx_t - \sigma_tv^*_t(x_t)).
\end{align*}
Similarly, the posterior mean of $z$ is given by:
\begin{align*}
    v^*_t(x_t) &= \frac{\alpha'_t}{\alpha_t}(x_t - \sigma_t\mathbb E_{z|x_t}[z]) + \sigma'_t\mathbb E_{z|x_t}[z], \\
    \mathbb E_{z|x_t}[z] &= \left(\sigma'_t - \frac{\alpha'_t}{\alpha_t}\sigma_t\right)^{-1}\left(v^*_t(x_t) - \frac{\alpha'_t}{\alpha_t}x_t\right) \\
    &= \frac{\alpha_t v^*_t(x_t) - \alpha'_t x_t}{\alpha_t\sigma'_t - \sigma_t\alpha'_t} = \nu_t^{-1}(\alpha_t v^*_t(x_t) - \alpha'_t x_t).
\end{align*}
Based on these, we can construct the conditional mean at time $s$:
\begin{align*}
    \mathbb E_{x,z|x_t}[x_s] &= \nu_t^{-1}((\alpha_s\sigma'_t - \sigma_s\alpha'_t)x_t - (\alpha_s\sigma_t - \sigma_s\alpha_t)v^*_t(x_t)) \\
    &= \nu_t^{-1}(A'_{t,s}x_t - A_{t,s}v^*_t(x_t)),
\end{align*}
where $A_{t,s} = \alpha_s\sigma_t - \sigma_s\alpha_t$, $A'_{t,s} = \partial_tA_{t,s}$, and $A''_{t,s} = \partial^2_tA_{t,s}$. Prior work formulates a flow map $f_\theta$ with a pseudo-velocity network $F_\theta$, given by
\begin{align*}
    f_\theta(x_t; t, s) = \nu_t^{-1}(A'_{t,s}x_t - A_{t,s}F_\theta(x_t; t, s)).
\end{align*}
Since it contains $\nu_t$ in its denominator, $\nu_t$ should be non-zero for an interval $t\in [0, T]$. Since $A_{t,t} = 0$ and $A'_{t, t} = \partial_t A_{t,s}|_{s=t} = \nu_t$, this formulation naturally induces the boundary condition $f_\theta(x_t; t, t) = x_t$.

When we assume $\alpha_t = \cos t$ and $\sigma_t = \sin t$ for trigonometric interpolation, $\nu_t = 1$, $A_{t,s} = \sin(t - s)$, and $A'_{t,s} = \cos(t - s)$. For linear interpolation $\alpha_t = 1 - t$ and $\sigma_t = t$, each term reduces to $\nu_t = 1$, $A_{t,s} = t - s$, and $A'_{t,s} = 1$. Based on these settings, prior work formulates flow maps as $f_\theta(x_t; t, s) = \cos(t - s)x_t - \sin(t - s)F_\theta(x_t; t, s)$ for trigonometric interpolation~\citep{sct}, and $f_\theta(x_t; t, s) = x_t - (t - s)F_\theta(x_t; t, s)$ for linear interpolation~\citep{meanflow, consistencyflowmatching,alignyourflow}.

We observe that assuming $\nu_t = \nu\ne 0$ for all $t\in [0, T]$ gives $A'_{t,t} = \partial_t A_{t,s}|_{s=t} = \nu$ and $A''_{t,t} = \partial^2_t A_{t,s}|_{s=t} = 0$, when $\nu$ is a constant. This induces the important property of vanishing linearization, which helps proofs in \cref{proof:linearization} and \cref{proof:compatibility}. Additionally, this assumption can simplify differentiation of the flow map formulation, which is required in the consistency training objective. For these reasons, we assume the \textit{$\nu$-condition}. As we mentioned, both linear interpolation on $[0, 1]$ and trigonometric interpolation on $[0, \pi/2]$ satisfy $\nu=1$.

\textbf{Possible $\nu$-constrained Interpolations.} Suppose that $\gamma_t$ is monotonically increasing over $t\in[0, 1]$ and satisfies the boundary conditions $\gamma_1 = 1$ and $\gamma_0 = 0$. Consider the interpolation defined by $\alpha_t = (1-\gamma_t)^c$ and $\sigma_t = \gamma_t^c$ for some constant $c\in[0.5, 1]$. Then, $\nu_t$ can be written as $\nu_t = c(1-\gamma_t)^{c-1}\gamma_t^{c-1}\gamma'_t$. Imposing $\nu_t = \nu$ gives $\gamma'_t = \nu[c(1-\gamma_t)^{c-1}\gamma_t^{c-1}]^{-1}$ and $c(1-\gamma)^{c-1}\gamma^{c-1}d\gamma = \nu dt$. Integrating both sides yields
\begin{align*}
    c\int(1-\gamma)^{c-1}\gamma^{c-1}d\gamma = \nu\int dt = \nu(t + C),
\end{align*}
where the constant $C$ vanishes due to $\gamma_0 = 0$. For the incomplete beta function $B$, this becomes
\begin{align*}
    cB(\gamma_t; c, c) =  c\int^{\gamma_t}_0(1-\eta)^{c-1}\eta^{c-1}d\eta = \nu\int_0^td\tau = \nu t.
\end{align*}
By the boundary condition, $cB(c, c) = cB(1; c, c) = \nu$, and thus $\gamma_t$ is characterized by
\begin{align*}
    cB(\gamma_t; c, c) = \nu\frac{cB(\gamma_t; c, c)}{cB(c, c)} = \nu I_{\gamma_t}(c, c) = \nu t \implies \gamma_t = I^{-1}_t(c, c),
\end{align*}
where $I$ denotes the regularized incomplete beta function.

In this case, $\gamma_t$ is characterized regardless of $\nu$, and \textit{this is a possible generalization of interpolations satisfying the $\nu$-condition}. In particular, when $c=0.5$,
\begin{align*}
    B\left(\gamma; \frac12, \frac12\right) = \int^\gamma_0 \frac{d\eta}{\sqrt{\eta(1 - \eta)}} = 2\arcsin\sqrt \gamma \implies \gamma_t = \sin^2\left(\frac{\pi}{2}t\right),
\end{align*}
which yields trigonometric interpolation on the interval $[0, 1]$. In this case, the corresponding constant is $\nu=\pi/2$. On the other hand, when $c=1$, we have $\gamma_t = t,\ \alpha_t = 1-t,\ \sigma_t = t$, which reduces to linear interpolation. Interpolating $c$ between 0.5 and 1.0 is a promising direction for exploring interpolations. 

\subsection{Recent Consistency-based Generative Models Are Flow Map Models}
\label{proof:interpretation}
\textbf{sCT.} Under trigonometric interpolation $x_t = \cos(t)x + \sin(t)z$,
\begin{align*}
f_\theta(x_t;t, s) = \cos(t - s)x_t - \sin(t - s)F_\theta(x_t; t, s).
\end{align*}
When $s=0$,
\begin{align*}
f_\theta(x_t; t) = \cos(t)x_t - \sin(t)F_\theta(x_t; t),
\end{align*}
which exactly recovers the sCT formulation. If consistency training is formulated without the stop-gradient operation, then the objective reduces to the direct training objective $\mathcal L_\mathrm{DT}$ in \cref{eq:dt} as $\Delta t \to 0$.
\begin{align*}
    &\frac{1}{\Delta t^2}\cdot\mathbb E\left[\left\|f_\theta(x_t; t, s) - f_{\theta}(x_{t - \Delta t}; t - \Delta t, s)\right\|^2_2\right] \\
    &= \frac{1}{\Delta t^2}\cdot\mathbb E\left[\left\|f_\theta(x_t; t, s) - \left[f_\theta(x_t; t, s) - \partial_t f_\theta(x_t;t, s)\cdot \Delta t - v_t(x_t|x)\cdot \nabla_xf_\theta(x_t; t, s)\cdot \Delta t + O(\Delta t^2)\right]\right\|^2_2\right] \\
    &= \frac{1}{\Delta t^2}\mathbb E\left[\left\|(\partial_tf_\theta(x_t; t, s) + v_t(x_t|x)\cdot\nabla_xf_\theta(x_t; t, s))\cdot \Delta t + O(\Delta t^2)\right\|^2_2\right] \\
    &= \mathbb E\left[\left\|\partial_t f_\theta(x_t; t, s) + v_t(x_t|x)\cdot\nabla_xf_\theta(x_t; t, s)\right\|^2_2\right] + O(\Delta t) \\
    &= \mathcal L_\mathrm{DT} + O(\Delta t).
\end{align*}
When we use the stop-gradient, the continuous-time consistency training objective is given by
\begin{align*}
    &\frac{1}{2\Delta t}\cdot \nabla_\theta\mathbb E\left[\left\|f_\theta(x_t; t, s) - f_{\theta^-}(x_{t - \Delta t}; t - \Delta t, s) \right\|^2_2\right] \\
    &= \frac{1}{\Delta t}\cdot \mathbb E\left[\nabla_\theta f_\theta(x_t; t, s)\cdot\left(f_{\theta^-}(x_t; t, s) - f_{\theta^-}(x_{t - \Delta t}; t - \Delta t, s)\right)\right] \\
    &= \nabla_\theta\mathbb E\left[f_\theta(x_t; t, s)\cdot\frac{f_{\theta^-}(x_t; t, s) - f_{\theta^-}(x_{t - \Delta t}; t - \Delta t, s)}{\Delta t}\right] \\
    &\to \nabla_\theta\mathbb E\left[f_\theta(x_t; t, s)\frac{df_{\theta^-}(x_t; t, s)}{dt}\right] = \nabla_\theta\mathcal L_\mathrm{CT}
\end{align*}
as $\Delta t\to 0$, where $\theta^- = \operatorname{sg}[\theta]$ for the stop-gradient operation $\operatorname{sg}[\cdot]$. Using $\nabla_\theta f_\theta^T y = \frac12\nabla_\theta \|f_\theta - \operatorname{sg}[f_\theta - y]\|^2_2$, the practical implementation of this objective is given by
\begin{align*}
    \nabla_\theta\mathcal L_\mathrm{CT} &= \nabla_\theta\mathbb E\left[f_\theta(x_t; t, s)\frac{df_{\theta^-}(x_t; t, s)}{dt}\right] \\
    &= \nabla_\theta\mathbb E\left[\frac12\left\|f_\theta(x_t; t, s) - f_{\theta^-}(x_t; t, s) + \frac{df_{\theta^-}(x_t; t, s)}{dt}\right\|^2_2\right] \\
    &= \nabla_\theta\mathbb E\left[\frac12\left\|f_\theta(x_t; t, s) - \operatorname{sg}\left[f_\theta(x_t; t, s) - \left[\partial_tf_\theta(x_t; t, s) + v_t(x_t|x)\cdot\nabla_xf_\theta(x_t; t, s)\right]\right]\right\|^2_2\right],
\end{align*}
which coincides with \cref{eq:ct}, since $df_\theta/dt = \partial_tf_\theta + v_t\cdot\nabla_xf_\theta$. We note that the interior of the formulation can be interpreted as a fixed-point-style update, $f_\theta\gets f_\theta - df_\theta/dt$, of the direct training objective. Similarly, consistency distillation can be interpreted as a fixed-point-style update of Eulerian distillation.

From another perspective, since the time derivative of $f_\theta(x_t; t, s)$ is given by
\begin{align*}
    \frac{df_\theta(x_t; t, s)}{dt} &= \partial_t f_\theta(x_t; t, s) + v_t(x_t|x)\cdot \nabla_xf_\theta(x_t; t, s) \\
    &= L_*f_\theta(x_t; t, s) + \Delta v\cdot \nabla_xf_\theta(x_t; t, s),
\end{align*}
where $\Delta v = v_t(x_t|x) - v^*_t(x_t)$ and $L_*f_\theta = \partial_tf_\theta + v^*_t\cdot \nabla_xf_\theta$, the objective can be written as
\begin{align*}
    \mathcal L_\mathrm{CT} = \mathbb E\left[f_\theta\cdot \left(L_*f_{\theta^-}\right) + f_\theta\cdot\left(\Delta v\cdot \nabla_x f_{\theta^-}\right)\right].
\end{align*}
Here and below, we omit $x_t$, $t$, and $s$ for brevity. The first term on the right-hand side corresponds to consistency distillation. By the tower property, the second term vanishes under conditional expectation:
\begin{align*}
    \mathbb E\left[f_\theta\cdot (\Delta v\cdot \nabla_xf_{\theta^-})\right] &= \mathbb E\left[\mathbb E_{x|x_t}\left[f_\theta\cdot (\Delta v\cdot \nabla_xf_{\theta^-})\right]\right] \\
    &= \mathbb E\left[f_\theta\cdot\left(\mathbb E_{x|x_t}\left[\Delta v\right]\cdot \nabla_xf_{\theta^-}\right)\right] \\
    &= \mathbb E\left[f_\theta\cdot\left(0\cdot \nabla_xf_{\theta^-}\right)\right] \\
    &= 0.
\end{align*}
Thus, $\mathcal L_\mathrm{CT}$ reduces to consistency distillation at the expectation level, even along the conditional trajectory. The instability of these gradient dynamics is discussed in \cref{proof:instability}. \qed

\textbf{MeanFlow.} Suppose a flow map model under linear interpolation $x_t = (1 - t)x + tz$:
\begin{align*}
f_\theta(x_t; t, s) = x_t + (s - t)F_\theta(x_t; t, s).
\end{align*}
The corresponding direct training objective is
\begin{align*}
\mathcal L(\theta) = \mathbb E \left[\left\|\partial_tf_\theta(x_t; t, s) + v_t(x_t|x)\cdot\nabla_xf_\theta(x_t; t, s)\right\|^2_2\right] = \mathbb E\left[\left\|\frac{df_\theta(x_t; t, s)}{dt}\right\|^2_2\right],
\end{align*}
where
\begin{align*}
\frac{d}{dt}f_\theta(x_t; t, s) = v_t(x_t|x) - F_\theta(x_t; t, s) + (s - t)\frac{d}{dt}F_\theta(x_t; t, s).
\end{align*}
Recall the MeanFlow objective from \citet{meanflow}
\begin{align}\label{eq:meanflow-training}
\mathcal L(\theta) = \mathbb E[\|u_\theta(z_t; r, t) - \operatorname{sg}[v_t - (t - r)(v_t\cdot\partial_zu_\theta + \partial_tu_\theta)]\|^2_2].
\end{align}
Rewriting the MeanFlow objective using flow map notation gives
\begin{align*}
& \nabla_\theta\mathbb E\left[\|F_\theta(x_t; t, s) - \operatorname{sg}[v_t(x_t|x) - (t - s)(v_t(x_t|x)\cdot\nabla_xF_\theta + \partial_tF_\theta)]\|^2_2\right] \\
&= \nabla_\theta\mathbb E\left[\left\|F_\theta(x_t; t, s) - v_t(x_t|x) + (t - s)\cdot\frac{d}{dt}F_{\theta^-}(x_t; t, s)\right\|^2_2\right] \\ 
&= \nabla_\theta\mathbb E\left[\left\|F_\theta(x_t; t, s) - F_{\theta^-}(x_t; t, s) - \left[v_t(x_t|x) - F_{\theta^-}(x_t; t, s) + (s - t)\frac{d}{dt}F_{\theta^-}(x_t; t, s)\right]\right\|^2_2\right] \\
&= \nabla_\theta\mathbb E\left[\frac{2}{(t-s)}f_\theta(x_t; t, s)\frac{df_{\theta^-}(x_t; t, s)}{dt}\right].
\end{align*}
Thus, the MeanFlow objective is a weighted instance of the continuous-time consistency training under linear interpolation. We note that their ideal MeanFlow identity coincides with Eulerian distillation in \cref{eq:ed}. However, their practical implementation in \cref{eq:meanflow-training} uses stop-gradient and conditional velocity, rather than marginal velocity, thereby coinciding with the continuous-time consistency training objective. \qed

\textbf{Shortcut Model.}
From \citet{shortcutmodel}, the Shortcut Model objective consists of the flow matching objective and the consistency objective. 
\begin{align*}
\mathcal L_\mathrm{SC} = \mathbb E[\|s_\theta(x_t; t, 0) - v_t\|^2_2 + \|s_\theta(x_t;t, 2d) - [s_\theta(x_t; t, d) + s_\theta(x'_{t+d}; t+d, d)] / 2\|^2_2],
\end{align*}
with $x'_{t+d} = x_t + d\cdot s_\theta(x_t; t, d)$. By setting $d=s-t$ and $F_\theta(x_t;t,s)=s_\theta(x_t;t,s-t)$, sampling $t \sim \mathcal U[0,1]$, and choosing $s=t-2^{-d'}$ for $d' \sim \mathrm{Cat}[1,7]$, we obtain the flow map under linear interpolation
\begin{align*}
f_\theta(x_t; t, s) = x_t + (s-t)F_\theta(x_t; t, s) = x'_{t+d}.
\end{align*}
We can rewrite the flow matching objective of the Shortcut Model as
\begin{align*}
\|s_\theta(x_t; t, 0) - v_t\|^2_2 = \|F_\theta(x_t; t, t) - v_t(x_t|x)\|^2_2.
\end{align*}
For $r=s+d$, the consistency objective of the Shortcut Model can be written in the form of the semigroup condition, $f_{t,r}(x_t) = f_{s,r}(f_{t,s}(x_t))$:
\begin{align*}
& \|s_\theta(x_t;t, 2d) - [s_\theta(x_t; t, d) + s_\theta(x'_{t+d}; t+d, d)] / 2\|^2_2 \\
&= \|F_\theta(x_t; t, r) - [F_\theta(x_t; t, s) + F_\theta(f_\theta(x_t; t, s); s, r)]/2\|^2_2 \\
&= \frac{1}{4d^2}\|x_t + 2d\cdot F_\theta(x_t; t, r) - x_t - 2d[F_\theta(x_t; t, s) + F_\theta(f_\theta(x_t; t, s); s, r)]/2\|^2_2 \\ 
&= \frac{1}{4d^2}\|x_t + 2d\cdot F_\theta(x_t; t, r) - [x_t + d\cdot F_\theta(x_t; t, s) + d\cdot F_\theta(f_\theta(x_t; t, s); s, r)]\|^2_2 \\
&= \frac{1}{4d^2}\|f_\theta(x_t; t, r) - [f_\theta(x_t; t, s) + d\cdot F_\theta(f_\theta(x_t; t, s); s, r)]\|^2_2 \\
&= \frac{1}{4d^2}\|f_\theta(x_t; t, r) - f_\theta(f_\theta(x_t; t, s); s, r)\|^2_2.
\end{align*}
Hence, the objective of the Shortcut Model is
\begin{align*}
\mathcal L(\theta) = \mathbb E \left[\|F_\theta(x_t; t, t) - v_t\|^2_2 + \frac{1}{4d^2}\|f_\theta(x_t; t, r) - f_\theta(f_\theta(x_t; t, s); s, r)\|^2_2 \right].
\end{align*}
With the Taylor approximation of $F_{t,s} = F_\theta(x_t; t, s)$
\begin{align*}
    F_{t,r} = F_{t,s} + d\cdot \partial_s F_{t,s} + O(d^2),\quad F_{s,r} = F_{t,s} + d\cdot\partial_tF_{t,s} + d\cdot\partial_sF_{t,s} + d\cdot F_{t,s}\cdot \nabla_xF_{t,s} + O(d^2),
\end{align*}
we obtain
\begin{align*}
    &d[2F_{t,r} - F_{t,s} - F_{s,r}] \\
    & = d[2[F_{t,s} + d\cdot\partial_sF_{t,s}] - F_{t,s} - [F_{t,s} + d\cdot\partial_tF_{t,s} + d\cdot\partial_sF_{t,s} + d\cdot F_{t,s}\cdot\nabla_xF_{t,s}]] + O(d^3)\\
    & = d^2[\partial_sF_{t,s} - \partial_tF_{t,s} - F_{t,s}\cdot \nabla_xF_{t,s}] + O(d^3).
\end{align*}
Thus,
\begin{align*}
    \frac{1}{4d^2}\|f_{t,r}(x_t) - f_{s,r}(f_{t,s}(x_t))\|^2_2 = \frac{d^2}4\|\partial_sF_{t,s} - \partial_tF_{t,s} - F_{t,s}\cdot\nabla_xF_{t,s}\|^2_2 + O(d^3).
\end{align*}
The differentiation of the linear flow map with respect to timestep $t$ is given by
\begin{align*}
    \frac{df_{t,s}}{dt} = v_t^* - F_{t,s} + (s-t)\cdot(\partial_tF_{t,s} + v_t^*\cdot\nabla_x F_{t,s}).
\end{align*}
With the Taylor approximation and the relation $F_{t,t}\approx v^*_t$ obtained from $\mathcal L_\mathrm{CFM}$, we have
\begin{align*}
    F_{t,s} = F_{t,t} + d\cdot \partial_sF_{t,t} + O(d^2) \approx v_t^* + d\cdot\partial_sF_{t,t} + O(d^2).
\end{align*}
The identity $\partial_sF_{t,s} = \partial_sF_{t,t} + O(d)$ implies
\begin{align*}
    v_t^* - F_{t,s}(x_t) = -d\cdot\partial_sF_{t,s} + O(d^2).
\end{align*}
Hence,
\begin{align*}
    \frac{df_{t,s}}{dt} = d[\partial_tF_{t,s} + v_t^*\cdot\nabla_xF_{t,s} - \partial_sF_{t,s}] + O(d^2).
\end{align*}
Since $v_t^* \approx F_{t,t} = F_{t,s} + O(d)$, we further obtain 
\begin{align*}
    \frac{df_{t,s}}{dt} = d[\partial_tF_{t,s} + F_{t,s}\cdot\nabla_xF_{t,s} - \partial_sF_{t,s}] + O(d^2)
\end{align*}
and
\begin{align*}
    \mathcal L = \mathbb E_{x,z,t,s}\left[\|F_\theta(x_t; t, t) - v_t\|^2_2 + \frac1{4}\left\|\frac{d}{dt}f_\theta(x_t; t, s)\right\|^2_2 + O(d^3)\right].
\end{align*}

The model learns $F_\theta(x_t; t, t)\approx v^*_t(x_t)$ due to the flow matching loss term. This can be interpreted as the model learning a flow map corresponding to the trajectory induced by an approximated marginal velocity. Therefore, the Shortcut Model can be seen as Eulerian self-distillation under an $O(d^3)$ bound. \qed

\textbf{Consistency Trajectory Model.} Consistency Trajectory Model~\citep[CTM;][]{ctm} can be interpreted as training a network to satisfy the semigroup condition of the flow map, $f_{t,r}(x_t) = f_{s,r}(f_{t,s}(x_t))$, using the following objective:
\begin{align*}
    \mathcal L_\mathrm{CTM} = \mathbb E[\|f_{\theta^-}(f_\theta(x_t; t, r); r, 0) - f_{\theta^-}(f_{\theta^-}(\Phi(x_t; t, s); s, r); r, 0)\|^2_2],
\end{align*}
where $\Phi$ denotes a flow map along the learned flow of the teacher network, and timesteps satisfy $0<r<s<t$. They show that, under suitable assumptions, it converges to the teacher flow map in the version without the stop-gradient operation:
\begin{align*}
    \mathbb E[\|f_\theta(f_{\theta}(x_t; t, r); r, 0) - f_\theta(f_\theta(\Phi(x_t; t, s); s, r); r, 0)\|^2_2].
\end{align*}
In the flow map perspective, while they assume a variance-exploding process and an $x$-prediction network, their velocity model follows the same formulation as linear flow (i.e., $v_t(x_t|x) = t^{-1}(x_t - x)$), and the flow map formulation is also consistent with it. Therefore, for analytical convenience, we adopt a linear interpolation-based formulation.

As shown in relation to the Shortcut Model, the semigroup-based objective, $\|f_\theta(x_t; t, r) - f_\theta(f_\theta(x_t; t, s); s, r)\|^2_2$, can be interpreted as a weighted Eulerian self-distillation framework. In addition, \citet{boffi2025} proves that the semigroup-based objective, denoted as progressive self-distillation, encourages the network to learn the desired flow map.

Since CTM aims to distill the knowledge of the teacher velocity network, it replaces the inner flow map $f_\theta(x_t; t, s)$ with the teacher flow $\Phi(x_t; t, s)$. If we assume that CTM uses the metric function $d_r(x, y) = \|f_{\theta^-}(x; r, 0) - f_{\theta^-}(y; r, 0)\|^2_2$, this objective reduces to $d_r(f_\theta(x_t; t, r), f_{\theta^-}(\Phi(x_t; t, s); s, r))$. We can view it as a teacher-guided, metricized semigroup-based flow map model, which admits an Eulerian distillation interpretation under an asymptotic gap. \qed

\textbf{Consistency Flow Matching.}
Suppose a flow map under linear interpolation with fixed $s = 0$:
\begin{align*}
f_\theta(x_t; t) = x_t - tF_\theta(x_t; t).
\end{align*}
Then, the Consistency Flow Matching objective from \citet{consistencyflowmatching} becomes
\begin{align*}
\mathcal L(\theta) = \mathbb E \left[\|f_\theta(x_t; t) - f_{\theta^-}(x_{t - \Delta t}; t - \Delta t)\|^2_2 + \alpha\|F_\theta(x_t; t) - F_{\theta^-}(x_{t-\Delta t}; t - \Delta t)\|^2_2\right].
\end{align*}
The first term on the right-hand side corresponds to the consistency training objective, and the second term is the regularizer. Hence, we can interpret Consistency Flow Matching as training a flow map model via consistency training with regularization. \qed

\textbf{UCGM.}
For arbitrary interpolation of $\alpha_t, \sigma_t$, setting $s=0$ yields
\begin{align*}
f_\theta(x_t;t)  = \nu^{-1}_t(\sigma'_tx_t - \sigma_t F_\theta).
\end{align*}
We can reformulate the objective while preserving the gradient direction up to a positive scalar weight:
\begin{align*}
&\nabla_\theta\|f_\theta(x_t; t)- f_{\theta^-}(x_{\lambda t}; \lambda t)\|^2_2 \\
&= 2\cdot\nabla_\theta f_\theta(x_t; t)\cdot(t-\lambda t)\frac{f_{\theta^-}(x_t; t) - f_{\theta^-}(x_{\lambda t}; \lambda t)}{t - \lambda t} \\
&\propto 2\cdot\nabla_\theta f_{\theta}(x_t; t)\cdot\frac{f_{\theta^-}(x_t; t) - f_{\theta^-}(x_{\lambda t}; \lambda t)}{t - \lambda t} \\
&= -2\frac{\sigma_t}{\nu_t}\nabla_\theta F_\theta(x_t; t)\cdot\frac{f_{\theta^-}(x_t; t) - f_{\theta^-}(x_{\lambda t}; \lambda t)}{t - \lambda t} \\
& = \nabla_\theta\left\|F_\theta(x_t; t) - F_{\theta^-}(x_t; t) - \frac{\sigma_t[f_{\theta^-}(x_t; t) - f_{\theta^-}(x_{\lambda t}; \lambda t)]}{\nu_t(t - \lambda t)}\right\|^2_2,
\end{align*}
which gives the objective of UCGM.
\label{proof:gap-between-fm-fmm}
When $\lambda=0$, this reduces to the flow matching objective since $\lambda t = 0$ collapses $f_{\theta^-}(x_0; 0) = x_0$. In this case, the objective becomes origin prediction, which in turn yields $F_\theta$ with a velocity-matching objective. Otherwise, setting $\lambda\to 1$ reduces the objective to consistency training by $\Delta = F_\theta(x_t; t) - F_{\theta^-}(x_t; t) - \frac{\sigma_t}{\nu_t}\frac{df_{\theta^-}(x_t; t)}{dt}$. For $\lambda\in(0, 1)$, the objective $\|f_\theta(x_t; t) - f_{\theta^-}(x_{\lambda t}; \lambda t)\|^2_2$ yields consistency along the geometric sequence $\mathcal T_\lambda(t) = \{\lambda^kt\}^N_{k=0}$.

Define
\begin{align*}
    g_t(x_t) = (x_t - \alpha_tf_t(x_t))\sigma_t^{-1}.
\end{align*}
Then, when $f_t(x_t) = x$, it follows that $g_t(x_t) = z$ for $x_t = \alpha_tx + \sigma_t z$. Using this, the flow map can be formulated in a DDIM-like manner as
\begin{align*}
    f_{t,s}(x_t) = \alpha_sf_t(x_t) + \sigma_s g_t(x_t) = \frac{\sigma_s}{\sigma_t}x_t + (\alpha_s - \frac{\sigma_s}{\sigma_t}\alpha_t)f_t(x_t).
\end{align*}
Assuming the composition chain
\begin{align*}
    f_{s,r}(f_{t,s}(x_t)) = f_{s,r}(\tilde x_s) = \frac{\sigma_r}{\sigma_s}\tilde x_s + (\alpha_r - \frac{\sigma_r}{\sigma_s}\alpha_s)f_s(\tilde x_s),
\end{align*}
for $\tilde x_s = f_{t,s}(x_t)$, we obtain
\begin{align*}
    f_{t,r}(x_t) - f_{s,r}(f_{t,s}(x_t)) = (\alpha_r - \frac{\sigma_r}{\sigma_s}\alpha_s)(f_t(x_t) - f_s(\tilde x_s)).
\end{align*}
For $s = \lambda^kt$ for some $k\in\mathbb N$, if it follows that $\tilde x_s = f_{t,s}(x_t) \approx x_s$, then $f_s(x_s) = f_t(x_t)$ and $f_{t,r} = f_{s,r}\circ f_{t,s}$. In this case, the flow map can be constructed along the geometric sequence $\mathcal T_\lambda(t)$.

In general, the velocity of the DDIM map is given by $\frac{d}{ds}f_{t,s} = \alpha'_sf_t + \sigma'_sg_t$. Since the marginal velocity is
\begin{align*}
    v^*_t(x_t) = \alpha'_t\mathbb E_{x|x_t}[x] + \sigma'_t\mathbb E_{x|x_t}[(x_t - \alpha_tx)\sigma_t^{-1}],
\end{align*}
the DDIM map coincides with the flow map only when $v^*_s(f_{t,s}(x_t)) = \frac{d}{ds}f_{t,s}(x_t)$, which implies $\mathbb E_{x|\tilde x_s}[x] = f_t(x_t)$. Setting $s\to t$ reduces this condition to $\mathbb E_{x|x_t}[x] = f_t(x_t)$ by the identity assumption. However, this condition fails to preserve the injectivity of the flow map at $t=1$ due to the mean collapse problem. Therefore, the DDIM-style map does not generally coincide with the flow map. \qed

\textbf{Reflow.}
Rectified flows introduce Reflow to straighten trajectories after training. In Reflow, sampling from the trained model is performed via 
\begin{align*}
x_0=z + \int_1^0v_\theta(x_t;t)dt\approx \mathrm{ODESolver}(v_\theta, z, 1, 0),
\end{align*}
followed by finetuning w.r.t. the coupling $\Pi_{Z,\theta} = p_Z(z)p_{v_\theta}(x|z)$. The velocity $\hat v_t$ of the trajectory induced by the coupling $\Pi_{Z,\theta}$ is given by
\begin{align*}
\hat v_t = z - \left(z + \int^0_1v_\theta(x_t; t)dt\right) = \int^1_0v_\theta(x_t; t)dt,
\end{align*}
which corresponds to the displacement of the flow map. Therefore, Reflow can be interpreted as direct supervision of the flow map under linear interpolation. \qed

\subsection{Suboptimality of Direct Training}
\label{proof:suboptimality}
Unlike Eulerian distillation, direct training does not guarantee convergence to the optimal flow map. Consider the direct training objective using the conditional velocity:
\begin{align*}
\mathcal{L}_\mathrm{DT}=\mathbb{E}_{x,z,t,s}\left[\|\partial_tf_\theta(x_t; t, s)+v_t(x_t|x)\cdot\nabla_xf_\theta(x_t; t, s)\|^2_2 \right].
\end{align*}
By defining the velocity error as $\Delta v=v_t(x_t|x)-v^*_t(x_t)$, we can rewrite the objective in the form of Eulerian distillation:
\begin{align*}
\mathcal{L}_\mathrm{DT}
    &=\mathbb{E}\left[\|\partial_tf_\theta(x_t; t, s)+(\Delta v+v^*_t(x_t))\cdot\nabla_xf_\theta(x_t; t, s)\|^2_2 \right]\\
    &=\mathbb{E}\left[\|\partial_tf_\theta(x_t; t, s)+\Delta v\cdot\nabla_xf_\theta(x_t; t, s)+v^*_t(x_t)\cdot\nabla_xf_\theta(x_t; t, s)\|^2_2 \right] \\
    &=\mathbb{E}\left[\|\partial_tf_\theta(x_t; t, s)+v^*_t(x_t)\cdot\nabla_xf_\theta(x_t; t, s)\|^2_2 \right] + \mathbb{E}\left[\|\Delta v\cdot\nabla_xf_\theta(x_t; t, s)\|^2_2 \right],
\end{align*}
where the cross term vanishes, since $\mathbb E_{x|x_t}[\Delta v] = 0$ and the remaining terms are fixed given $x_t$. In this case, the second term can be represented as
\begin{align*}
\mathbb E\left[\mathrm{Var}_{x|x_t}\left[\Delta v \cdot \nabla_xf_\theta(x_t; t, s) \right]\right].
\end{align*}
Under an independent coupling, the velocity error $\Delta v=v_t(x_t|x)-v^*_t(x_t)$ is typically nonzero. Consequently, unless $\|\nabla_xf_\theta(x_t; t, s)\|$ collapses to zero, the objective function inherently contains a larger variance term compared to that of Eulerian distillation. To minimize the overall loss, the optimizer faces a trade-off involving this variance.

Note that for flow matching, the loss term $\mathbb E_{x,z,t}[\|v_t(x_t|x) - F_\theta(x_t; t)\|^2_2]$ is decomposed as follows:
\begin{align*}
\mathbb E\left[\|\Delta v + v^*_t(x_t) - F_\theta(x_t; t)\|^2_2 \right] = \mathbb E \left[\|v^*_t(x_t) - F_\theta(x_t; t)\|^2_2 \right] + \mathrm{Var}[\Delta v].
\end{align*}
The variance term, $\mathrm{Var}[\Delta v]$, is independent of the network. Therefore, it does not affect convergence to the global optimum.

\textbf{Stationary Condition.} Since the gradients of Eulerian distillation and direct training differ, the optima obtained by gradient-based optimization can also differ. To find and compare the stationary points of the direct training objective, we apply the Euler-Lagrange equation. The objective can be represented in the vector form as
\begin{align*}
    \mathcal L_\mathrm{DT} = \mathbb E_{x,z,t,s}\left[\|\partial_tf_{t,s}(x_t) + v_t(x_t|x)^T\nabla_xf_{t,s}(x_t)\|^2_2\right] = \iiint_\Omega\rho_tL(f, \partial_t f, \nabla_xf) \,dx_t\,dt\,ds.
\end{align*}
We set the conditional expectation as the Lagrangian,
\begin{align*}
    \rho_tL(f, \partial_t f, \nabla_x f) = \rho_t(x_t)\cdot \mathbb E_{x|x_t}[\|\partial_tf_{t,s}(x_t) + v_t(x_t|x)^T\nabla_xf_{t,s}(x_t)\|^2_2].
\end{align*}
The corresponding Euler-Lagrange equation is
\begin{align*}
    \frac{\partial (\rho_tL)}{\partial f} - \partial_t\left(\frac{\partial (\rho_tL)}{\partial (\partial_tf)}\right) - \nabla_x\cdot\left(\frac{\partial (\rho_tL)}{\partial(\nabla_xf)}\right) = 0 \iff \mathcal S_\mathrm{DT} = \partial_t(\rho_t\mathbb E_{x|x_t}[R]) + \nabla_x\cdot(\rho_t\mathbb E_{x|x_t}[v_tR]) = 0,
\end{align*}
where the residual is defined as 
\begin{align*}
    R = \partial_t f_{t,s}(x_t) + v_t(x_t|x)^T\nabla_xf_{t,s}(x_t).
\end{align*}
Letting $\Delta v = v_t(x_t|x) - v^*_t(x_t)$ and introducing the operator $L_*f = \partial_t f + (v^*_t)^T\nabla_xf$, the residual can be rewritten as $R = L_*f_{t,s} + \Delta v^T\nabla_xf_{t,s}$, so that
\begin{align*}
    &\partial_t\left(\rho_t\mathbb E_{x|x_t}[R]\right) = \partial_t(\rho_tL_*f_{t,s}), \\
    &\begin{aligned}
        \nabla_x\cdot (\rho_t\mathbb E_{x|x_t}[v_tR]) &= \nabla_x \cdot(\rho_t\mathbb E_{x|x_t}[(v^*_t + \Delta v)(L_*f_{t,s} + \Delta v^T\nabla_xf_{t,s})]) \\
        &= \nabla_x\cdot \left(\rho_tv^*_tL_*f_{t,s} + \rho_t\mathbb E_{x|x_t}[\Delta v(\Delta v^T\nabla_xf_{t,s})]\right) \\
        &= \nabla_x\cdot \left(\rho_tv^*_tL_*f_{t,s} + \rho_t\Sigma_{\Delta v|x_t}\nabla_xf_{t,s}\right).
    \end{aligned}
\end{align*}
Therefore, the optimality condition becomes
\begin{align*}
    \mathcal S_\mathrm{DT} = \partial_t(\rho_tL_*f_{t,s}) + \nabla_x\cdot (\rho_tv^*_t L_*f_{t,s}) + \nabla_x\cdot (\rho_t\Sigma_{\Delta v|x_t}\nabla_xf_{t,s}) = 0,
\end{align*}
and using the continuity equation $\partial_t\rho_t + \nabla\cdot(\rho_tv^*_t) = 0$ further simplifies to
\begin{align*}
    \mathcal S_\mathrm{DT} &= (L_*f_{t,s})\partial_t\rho_t + \rho_t(\partial_t L_*f_{t,s}) + (L_*f_{t,s})\nabla\cdot (\rho_t v^*_t) + (\rho_tv^*_t)^T\nabla(L_*f_{t,s}) + \nabla\cdot(\rho_t\Sigma_{\Delta v|x_t}\nabla_xf_{t,s}) \\
    &= \cancel{(L_*f_{t,s})(\partial_t\rho_t + \nabla\cdot (\rho_tv^*_t))} + \rho_t(\partial_t L_*f_{t,s} + (v^*_t)^T\nabla(L_* f_{t,s})) + \nabla\cdot(\rho_t\Sigma_{\Delta v|x_t}\nabla_xf_{t,s}) \\
    &= \rho_t L_*(L_*f_{t,s}) + \nabla\cdot (\rho_t\Sigma_{\Delta v|x_t}\nabla_xf_{t,s})
\end{align*}
When $\Sigma_{\Delta v|x_t} = \mathrm{Cov}_{x|x_t}[v_t(x_t|x)]\to 0$, the condition reduces to $\rho_tL_*(L_*f_{t,s}) = 0$, which coincides with the stationary condition of Eulerian distillation. Thus, in \cref{prop:suboptimality}, we denote it by $\mathcal S_\mathrm{ED} = \rho_tL_*(L_*f_{t,s})$, and the bias term by $C = \nabla\cdot(\rho_t\Sigma_{\Delta v|x_t}\nabla_x f_{t,s})$, such that $\mathcal S_\mathrm{DT} = \mathcal S_\mathrm{ED} + C = 0$. Therefore, the stationary point of direct training can fail to satisfy the stationary condition of Eulerian distillation: $\mathcal S_\mathrm{ED} = -C\ne 0$. \qed

\subsection{Instability of Consistency Training}
\label{proof:instability}

The continuous-time consistency training objective employs a stop-gradient operation. In this case, the objective is defined as 
\begin{align*}
    \mathcal L_\mathrm{CT} &= \mathbb E\left[f_{t,s}(x_t)^T\operatorname{sg}\left[\frac{df_{t,s}(x_t)}{dt}\right]\right] \\
    &= \mathbb E\left[f_{t,s}(x_t)^T\operatorname{sg}\left[L_*f_{t,s} + \Delta v^T\nabla_x f_{t,s}\right]\right].
\end{align*}
Following \cref{proof:suboptimality}, the stationary condition induced by the stop-gradient objective is:
\begin{align*}
    \mathcal S_\mathrm{CT} &= \rho_t(x_t)\cdot\mathbb E_{x|x_t}\left[L_*f_{t,s} + \Delta v^T\nabla_xf_{t,s} \right] \\
    &= \rho_t(x_t)\cdot\left(L_*f_{t,s} + \mathbb E_{x|x_t}\left[\Delta v^T\nabla_xf_{t,s}\right]\right) \\
    &= \rho_t(x_t)\cdot L_*f_{t,s} = 0,
\end{align*}
and this stationary condition is equivalent to the Eulerian condition in \cref{eq:eulerian} when $\rho_t(x_t) > 0$. The stationary condition of $\mathcal L_\mathrm{CT}$ reduces to that of consistency distillation at the expectation level, which is consistent with the observation in \cref{proof:interpretation}. However, when the expectation is replaced by a finite-batch estimate, the empirical residual term corresponding to $\mathbb E[\Delta v^T\nabla_xf_{t,s}]$ may not vanish and may contribute to degenerate solutions in small batch settings (\cref{fig:ct-batchsize}). \qed

\textbf{Fixed-point Property.} We also note that since the objective is expressed as a linear term, the positive-semidefinite curvature that encourages stable convergence at the optimum is absent. It only specifies the fixed point on $L_*f_{t,s} = 0$, and the gradient dynamics alone may fail to converge. Empirically, we show that different initializations can result in varying performance (\cref{tab:reproducibility}). In some cases, training diverges to degenerate solutions and fails to generate meaningful images (details in \cref{detail:reproducing-ct}).

\textbf{Stop-gradient Placement.} \citet{flowmapmatching} analyzes the formulation that applies the stop-gradient operation only to the spatial gradient term:
\begin{align*}
    \mathbb E\left[\left\|\partial_tf_{t,s} - \operatorname{sg}\left[v_t^T\nabla_xf_{t,s}\right]\right\|^2_2\right].
\end{align*}
This formulation differs in structure from the consistency-style objectives discussed above. 

While they show that the functional gradient of this formulation can align with that of Eulerian distillation and admit fixed-point solutions, this analysis does not directly extend to prior consistency models, whose numerical realizations and optimization dynamics differ.

Motivated by this distinction, we formally analyze consistency-style objectives and show that they also admit fixed-point solutions satisfying the Eulerian equation, without implying stable or global convergence of the resulting optimization dynamics.

\begin{figure}[th]
\centering
\subfloat[Toy setting in \cref{fig:toy}]{%
\includegraphics[width=0.45\linewidth]{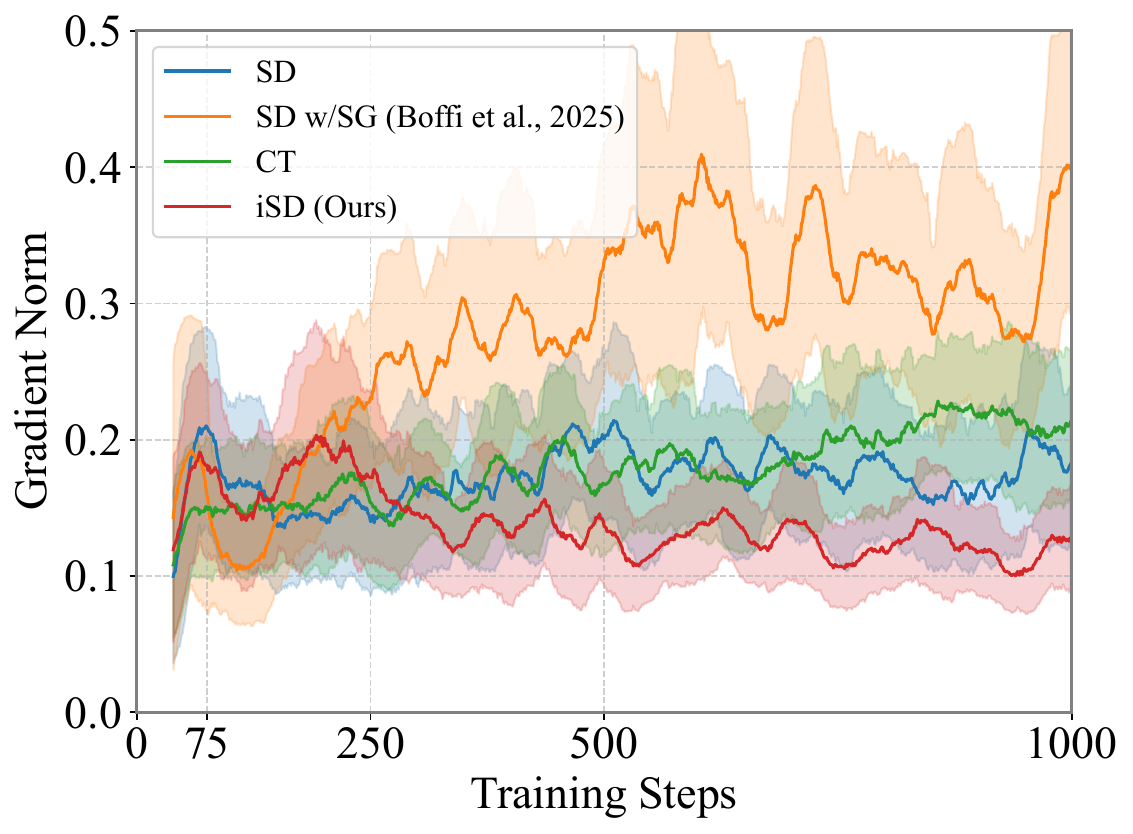}
}
\subfloat[ImageNet ablation setting (Linear, Pre-CFG 5.0)]{%
\includegraphics[width=0.45\linewidth]{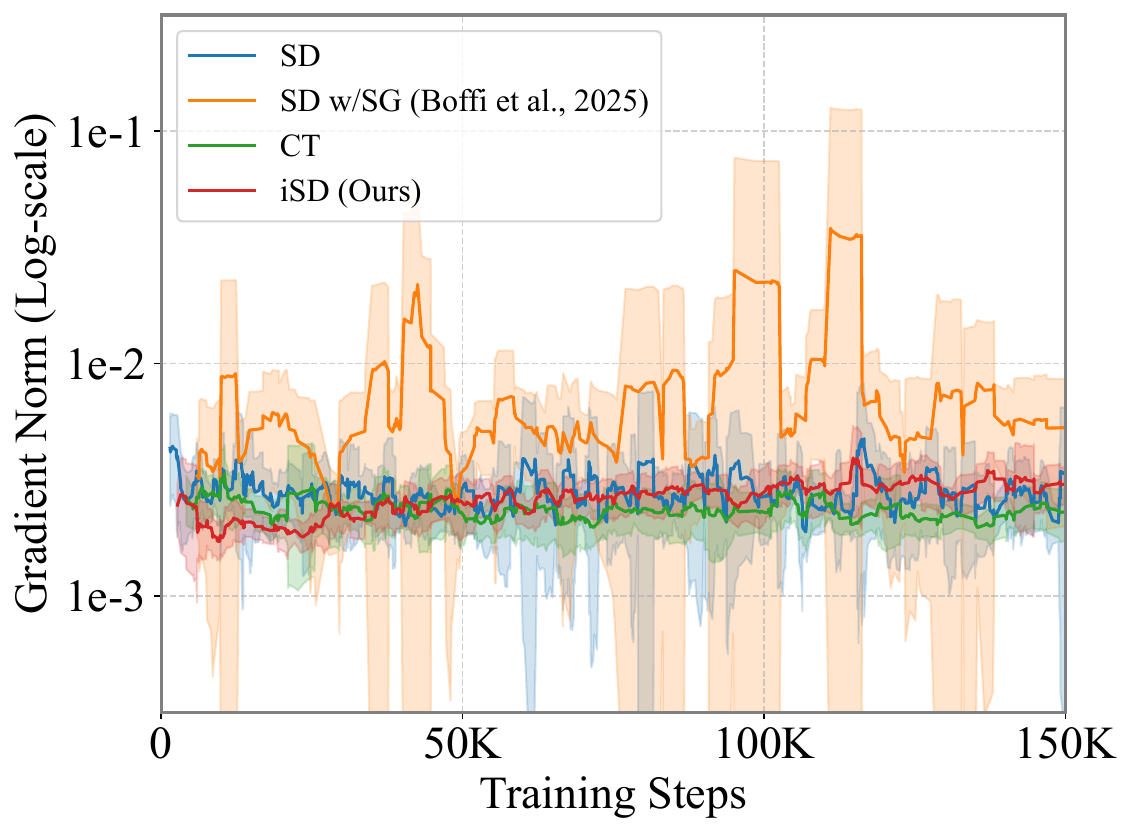}
}
\caption{Gradient norms of each method. The colored regions indicate 1 standard deviation. \textit{SD} denotes Eulerian self-distillation (ESD), \textit{SD w/SG} denotes the stop-gradient version of ESD introduced by \citet{boffi2025}, \textit{CT} denotes consistency training, and \textit{iSD} denotes our improved Self-Distillation.}
\label{fig:gradnorm}
\end{figure}

To highlight differences in numerical realization, we compare the gradient norms of each method, as shown in \cref{fig:gradnorm}. We observe that self-distillation with a stop-gradient applied to the spatial derivative~\citep{boffi2025}, $\mathbb E[\|\partial_t f_{t,s} + \operatorname{sg}[F_{t,t}^T\nabla_x f_{t,s}]\|^2_2]$, can yield larger gradient norms than the original formulation, whereas our iSD yields consistently smaller gradient norms.

\textbf{Network-induced Coupling.} Prior work~\citep{gc, vct} has introduced network-induced couplings (NIC) to reduce loss variance. For $x$ and $z$, NIC replaces one of them with a network prediction. Suppose that $\hat x = f_{t,0}(x_t)$ with stop-gradient: $\hat x_t = \alpha_t\operatorname{sg}[f_{t,0}(x_t)] + \sigma_t z$ for $x_t = \alpha_t x + \sigma_t z$. The conditional velocity of this coupling is $\hat x_t' = \alpha'_t\operatorname{sg}[f_{t,0}(x_t)] + \sigma'_t z$ and the marginal velocity becomes:
\begin{align*}
    v^\mathrm{NIC}_{t}(\hat x_t) &= \mathbb E_{x,z|\hat x_t}[\alpha'_t\operatorname{sg}[f_{t,0}(x_t)] + \sigma'_tz].
\end{align*}
This differs from the marginal velocity $v^*_t$ induced by the standard independent coupling, $v^\mathrm{NIC}_t(x_t)\ne v^*_t(x_t)$, in general. Therefore, the resulting flow map does not necessarily need to follow the standard flow field. We note that these couplings are known to reduce loss variance; however, we do not primarily focus on them, as the flow field is inconsistent with our targets.

\textbf{Discrete-time models} can be viewed as numerical approximations of corresponding continuous systems~\citep{consistencymodel}. We therefore primarily focus on the behavior of continuous systems that discrete models approximate. The identified instability arises from the structure of the objective, rather than time discretization. Thus, our analysis provides plausible explanations for similar instabilities observed in practice.

\subsection{Linearization Cost Hypothesis}
\label{proof:linearization}

Recall that $f_{t,s}(x_t) = \nu^{-1}(A_{t,s}'x_t - A_{t,s}F_{t,s}(x_t))$ for $A_{t,s} = \alpha_s\sigma_t - \sigma_s\alpha_t$. Under the given formulation of flow maps, a pseudo-velocity network $F_\theta$ learns the weighted displacement between $x_t$ and $x_s$. From the perspective of $F_\theta$, this process can be seen as shifting the training target from a classical instantaneous velocity field to a weighted displacement field, which transforms the path between $x_t$ and $x_s$ into the linear form of $x_t$ and $F_\theta$. We refer to it as \textit{linearization}.

Differentiating with respect to the starting point $t$ gives
\begin{align*}
    \frac{df_\theta(x_t; t, s)}{dt} = \nu^{-1}\left(A''_{t,s}x_t + A'_{t,s}(v_t^*(x_t) - F_\theta(x_t; t, s)) - A_{t,s}\frac{dF_\theta(x_t; t, s)}{dt}\right).
\end{align*}
Following \citet{alignyourflow}, the gradient of the consistency training objective can be written as 
\begin{align*}
    &\nabla_\theta\mathbb E\left[2f^T_\theta(x_t; t, s)\frac{df_{\theta^-}(x_t; t, s)}{dt}\right] \\
    &\propto \nabla_\theta\mathbb E\left[-A_{t,s}\nu^{-2} F_\theta(x_t; t, s)\cdot \left(A''_{t,s}x_t + A'_{t,s}(v_t^*(x_t) - F_{\theta^-}(x_t; t, s)) - A_{t,s}\frac{dF_{\theta^-}(x_t; t, s)}{dt}\right)\right].
\end{align*}
In this case, $v^*_t - F_\theta$ can be interpreted as the flow matching term weighted by $A'_{t,s}$, and $A''_{t,s}x_t - A_{t,s}dF_\theta/dt$ as a linearization term involving the JVP, which penalizes the $t$-dependent outputs of $f_\theta$. We note that $A_{t,t} = 0$ by definition, and $A''_{t,t} = \partial^2_tA_{t,s}|_{s=t} = 0$ since the $\nu$-condition gives $A'_{t,t} = \partial_t A_{t,s}|_{s=t} = \nu$, where $\nu$ is the time-independent constant. Thus, the linearization term vanishes when $s\to t$, and the objective becomes flow matching. Conversely, $s\to 0$ for fixed $t$ increases the step size and relatively strengthens the linearization term.

We hypothesize that the linearization cost increases with step size, making optimization more challenging. This is because the linearization term involves a complex structure induced by the JVP, whereas the flow matching term requires only a simple forward pass. Such a complex structure may introduce additional fixed points or spikes, hindering convergence to the desired optimum. To support this hypothesis, we conduct an ablation study over the step size parameter $d$ in our ImageNet-1K ablation settings (\cref{appendix:add-expr}), and observe that small step sizes are associated with lower loss values and better quantitative results. Moreover, we construct loss landscapes before and after time-condition relaxation in \cref{appendix:loss-landscape}, and observe that the relaxation reduces loss spikes and variances empirically.

\subsection{Training-time Classifier-free Guidance}
\label{proof:ttcfg}
MeanFlow suggests classifier-free guidance~\citep[CFG;][]{cfg} for flow maps:
\begin{align*}
    &\nabla_\theta\mathbb E\left[\left\|F_\theta(x_t; t, s, c) - \operatorname{sg}\left[\tilde v_t(x_t|x) - (t-s)\left(\tilde v_t(x_t|x)\cdot\nabla_xF_\theta + \partial_tF_\theta\right)\right]\right\|^2_2\right] \\
    &\mathrm{where}\quad \tilde v_t(x_t|x) = F_{\theta^-}(x_t; t, t, \varnothing) + \omega (v_t(x_t|x) - F_{\theta^-}(x_t; t, t, \varnothing)),
\end{align*}
for a conditional class label $c$, the null class label $\varnothing$ of unconditional generation, and a guidance scale $\omega$. We find that this formulation encourages the flow map to follow the ground-truth CFG trajectory. The ground-truth CFG trajectory is given by $\tilde v_t^*(x_t; c) = v^*_t(x_t) + \omega(v^*_t(x_t; c) - v^*_t(x_t))$ where  $v^*_t(x_t; c)$ is the class-conditional marginal velocity $v^*_t(x_t; c) = \mathbb E_{x|x_t, c}[v_t(x_t|x)]$ and $v^*_t(x_t)$ is the marginal velocity $v^*_t(x_t) = \mathbb E_{x|x_t}[v_t(x_t|x)] = \mathbb E_{c|x_t}[v^*_t(x_t; c)]$.

To prove this, we first consider only the flow matching setting:
\begin{align*}
    &\nabla_\theta\mathbb E\left[\left\|F_\theta(x_t; t, c) - \tilde v_t(x_t|x)\right\|^2_2\right] \\
    &= \nabla_\theta\mathbb E\left[\left\|F_\theta(x_t; t, c) - \left(F_{\theta^-}(x_t; t, \varnothing)  + \omega\left(v_t(x_t|x) - F_{\theta^-}(x_t; t, \varnothing)\right)\right)\right\|^2_2\right].
\end{align*}
With label dropout $c=\varnothing$, the objective reduces to weighted unconditional flow matching that converges to the unconditional marginal velocity:
\begin{align*}
    &\nabla_\theta\mathbb E\left[\left\|F_\theta(x_t; t, \varnothing) - (F_{\theta^-}(x_t; t, \varnothing) + \omega(v_t(x_t|x) - F_{\theta^-}(x_t; t, \varnothing)))\right\|^2_2\right] \\
    &= \mathbb E\left[2\omega \cdot \nabla_\theta F_\theta(x_t; t, \varnothing)\cdot (F_{\theta^-}(x_t; t, \varnothing) - v_t(x_t|x))\right] \\
    &=\nabla_\theta\mathbb E\left[\omega \left\|F_{\theta}(x_t; t, \varnothing) - v_t(x_t|x)\right\|^2_2\right].
\end{align*}
In this case, we can set $F_\theta(x_t; t, \varnothing)\approx v^*_t(x_t)$ if the network has sufficient capacity. For the conditional case $c\ne\varnothing$, applying the tower property yields:
\begin{align*}
    &\nabla_\theta\mathbb E\left[\left\|F_\theta(x_t; t, c) - (F_\varnothing(x_t; t) + \omega (v_t(x_t|x) - F_\varnothing(x_t; t)))\right\|^2_2\right] \\
    &= \nabla_\theta\mathbb E\left[\mathbb E_{x|x_t, c}\left[\left\| F_\theta(x_t; t, c) - (F_\varnothing(x_t; t) + \omega(v_t(x_t|x) - F_\varnothing(x_t; t)))\right\|^2_2\right]\right] \\
    &= \nabla_\theta\mathbb E\left[\mathbb E_{x|x_t, c}\left[\left\| F_\theta(x_t; t, c) - (F_\varnothing(x_t; t) + \omega(v^*_t(x_t; c) - v^*_t(x_t; c) +  v_t(x_t|x) - F_\varnothing(x_t; t)))\right\|^2_2\right]\right] \\
    &= \nabla_\theta\mathbb E\left[\mathbb E_{x|x_t, c}\left[\left\|F_\theta(x_t; t, c) - (F_\varnothing(x_t; t) + \omega(v^*_t(x_t; c) - F_\varnothing(x_t; t))) + \omega(v^*_t(x_t; c) - v_t(x_t|x))\right\|^2_2\right] \right] \\
    &\approx \nabla_\theta\mathbb E\left[\left\|F_\theta(x_t; t, c) - (v^*_t(x_t) + \omega (v^*_t(x_t; c) - v^*_t(x_t)))\right\|^2_2\right] + \omega^2\cdot \nabla_\theta\mathbb E\left[\mathrm{Var}_{x|x_t,c}\left[v_t(x_t|x)\right]\right] \\
    &=\nabla_\theta\mathbb E\left[\left\|F_\theta(x_t; t, c) - \tilde v^*_t(x_t; c)\right\|^2_2\right],
\end{align*}
where $F_\varnothing(x_t; t) = F_{\theta^-}(x_t; t, \varnothing)$ for brevity. Thus, the self-referential target $\tilde v_t(x_t|x)$ with label dropout encourages convergence to the ground-truth CFG velocity.

Consider the convergence of the flow map to the CFG velocity field. Following \cref{proof:interpretation}, let $L_*f_{t,s} = \partial_tf_{t,s} + \tilde v^*_t\cdot \nabla_xf_{t,s}$ and $\Delta v = \tilde v_t(x_t|x) - \tilde v^*_t(x_t; c)$. It is sufficient to show that $\mathbb E_{x|x_t, c}[\Delta v] = 0$, which is encouraged when label dropout is applied and the flow matching objective is jointly optimized. Thus, the objective encourages convergence to the flow map induced by the CFG velocity. \qed

\section{Improving Self-Distillation}
\label{appendix:isd}
\subsection{Compatibility of the Joint Objective}
\label{proof:compatibility}

Revisit our objective
\begin{align*}
    \mathcal L = \mathbb E_{x,z,t,s}\left[\left\|F_\theta(x_t; t, t) - v_t(x_t|x)\right\|^2_2 + \left\|\partial_tf_\theta(x_t; t, s) + F_{\theta^-}(x_t; t, t)\cdot \nabla_xf_\theta(x_t; t, s)\right\|^2_2\right].
\end{align*}
The first term on the right-hand side trains $F_\theta(x_t; t, t)$ to approximate the marginal velocity via the flow matching objective, while the second term learns the flow map $f_{t,s}$ along the trajectory of $F_{\theta^-}(x_t; t, t)$ in a self-distillation manner.

Individually, each term is encouraged to converge to its desired optimum: the marginal velocity and the flow map by the Eulerian equation (\cref{proof:eulerian-eqn}). From a joint perspective, we need to consider $t=s$, since the network is forced to optimize both terms simultaneously at this point. As Eulerian distillation collapses to the flow matching objective when $s\to t$ under the $\nu$-condition, the second term trains the model to learn the instantaneous velocity of the trajectory (\cref{proof:linearization}). In this case, $F_\theta(x_t; t, t)$ learns from $F_{\theta^-}(x_t; t, t)$, and inductively approximates $v^*_t(x_t)$ through the first term. This naturally reduces to non-conflicting joint training. For $t\ne s$, the network is conditioned differently in the two terms, and it can learn the proper mapping provided that the network capacity is sufficient. Consequently, the overall objective naturally trains the network to follow the marginal velocity as the trajectory of the flow map. \qed

\subsection{Deriving Final Objective}
\label{proof:final-objective}

Recall the consistency training objective under the flow map representation (\cref{proof:linearization}):
\begin{align*}
    &\nabla_\theta\mathbb E\left[f^T_\theta(x_t; t, s)\frac{df_{\theta^-}(x_t; t, s)}{dt}\right] \\
    &\propto \nabla_\theta\mathbb E\left[-A_{t,s}\nu^{-2} F_\theta(x_t; t, s)\cdot \left(A''_{t,s}x_t + A'_{t,s}(v_t^*(x_t) - F_{\theta^-}(x_t; t, s)) - A_{t,s}\frac{dF_{\theta^-}(x_t; t, s)}{dt}\right)\right] \\
    &= \frac12\cdot \nabla_\theta\mathbb E\left[A_{t,s}\nu^{-2}\|F_\theta(x_t; t, s) - \operatorname{sg}[F_\mathrm{tgt}(x_t; t, s)]\|^2_2\right],
\end{align*}
where $F_\mathrm{tgt}(x_t; t, s) = F_{\theta}(x_t; t, s) + \left(A''_{t,s}x_t + A'_{t,s}(v_t^*(x_t) - F_{\theta}(x_t; t, s)) - A_{t,s}\frac{dF_{\theta}(x_t; t, s)}{dt}\right)$.

To follow the marginal velocity, we replace $v^*_t(x_t)$ with the instantaneous velocity $v_\theta(x_t; t) = F_\theta(x_t; t, t)$ while jointly training with $\mathcal L_\mathrm{CFM}$. Particularly, for linear interpolation, we have $A_{t,s} = t-s,\ A'_{t,s} = 1$ and $A''_{t,s} = 0$. This simplifies the target to $F^\mathrm{lin}_\mathrm{tgt}(x_t; t, s) = v_\theta(x_t; t) - (t-s)F'_\theta(x_t; t, s)$, which coincides with the regression target of MeanFlow. For trigonometric interpolation, we have $A_{t,s} = \sin(t-s),\ A'_{t,s} = \cos(t-s)$ and $A''_{t,s} = -\sin(t-s)$. Thus, the target becomes $F_\mathrm{tgt}^\mathrm{tri}= F_\theta(x_t; t, s) + \cos(t-s)\cdot \left(v_\theta(x_t; t) - F_\theta(x_t; t, s)\right)- \sin(t-s)\cdot\left(x_t + F'_\theta(x_t; t, s)\right)$.

Although consistency training already admits the marginal flow map at its fixed point, the gradient in practice can exhibit a gap expressed as:
\begin{align*}
    \mathbb E_{x,z,t,s}[f_{t,s}\cdot(\Delta v\cdot \nabla_xf_{t,s})].
\end{align*}
When self-distillation is combined with flow matching, $\mathbb E\left[\|F_\theta(x_t; t, t) - v_t(x_t|x)\|^2_2\right]$, the network-estimated velocity can reduce the residual $\Delta v = F_\theta(x_t; t, t) - v^*_t(x_t)\approx v^*_t(x_t) - v^*_t(x_t) = 0$ relative to the per-sample conditional guidance $\Delta v = v_t(x_t|x) - v^*_t(x_t)$. In \cref{appendix:loss-landscape}, self-distillation is associated with lower loss landscape variance in the evaluated setting.

For JVP approximation, to ensure that $dx_t/dt$ follows the velocity $v_\theta(x_t; t) = F_\theta(x_t; t, t)$, we approximate
\begin{align*}
    \frac{dx_t}{dt} \approx \frac{[x_t + \epsilon\cdot v_\theta(x_t; t)] - [x_t - \epsilon\cdot v_\theta(x_t; t)]}{2\epsilon} = v_\theta(x_t; t).
\end{align*}
Thus, the full JVP approximation becomes
\begin{align*}
    \frac{dF_\theta(x_t; t, s)}{dt} = \frac{F_\theta(x_t + \epsilon\cdot v_\theta(x_t; t), t + \epsilon, s) - F_\theta(x_t - \epsilon\cdot v_\theta(x_t; t), t - \epsilon, s)}{2\epsilon} + O(\epsilon^2).
\end{align*}
Our final objective is
\begin{align*}
    \mathcal L_\mathrm{iSD} &= \lambda_1\mathcal L_\mathrm{CFM} + \lambda_2\mathcal L_\mathrm{SD\text{-}R}, \\
    \mathcal L_\mathrm{CFM} &= \mathbb E\left[\|F_\theta(x_t; t, t) - v_t(x_t|x)\|^2_2\right],\quad \mathcal L_\mathrm{SD\text{-}R} = \mathbb E\left[\|F_\theta(x_t; t, s) - \operatorname{sg}\left[F_\mathrm{tgt}(x_t; t, s)\right]\|^2_2\right].
\end{align*}
We consider two weighting methods: (i) cosine weighting $\lambda_1 = \lambda_2 = \cos(\pi t / 2)$ from \citet{ucgm}, and (ii) adaptive weighting $\lambda_1 = \lambda_2 = (\operatorname{sg}[\tilde{\mathcal L}_{t,s}(x_t, x)] + \eta)^{-p}$ where $\tilde{\mathcal L}_{t,s}(x_t, x) = \|F_\theta(x_t; t, t) - v_t(x_t|x)\|^2_2 + \|F_\theta(x_t; t, s) - \operatorname{sg}[F_\mathrm{tgt}(x_t; t, s)]\|^2_2$ from \citet{meanflow}. While our method enables training from scratch without additional weighting, we found that adaptive weighting can further improve performance in the SD-VAE setting, and cosine weighting in the VA-VAE setting. We use adaptive weighting by default, except when using VA-VAE.

When using JVP approximation, the truncation error is $O(\epsilon^2)$. This error appears acceptable under BF16 precision, as the approximation shows better results than direct JVP computation. The complete training and sampling algorithms are provided in \cref{alg:isd-training} and \cref{alg:isd-sampling}.

We note that the update in consistency training can be viewed as a coupled fixed-point iteration with a parameterized target. In this setting, contraction and global convergence properties are generally difficult to establish without additional assumptions. Instead, \cref{proof:compatibility} suggests that the two objectives are aligned and do not introduce conflicts. We leave deriving global convergence rates for future work.

\newcommand{\COMMENTa}[1]{\hfill $\triangleright$ {\small #1}}
\begin{algorithm}[th]
    \caption{(iSD Training) Training algorithm of vanilla iSD}
    \label{alg:isd-training}
    \begin{algorithmic}
        \State {\bfseries Input:} Noise distribution $p_Z$, data distribution $p_X$, model $F_\theta$, learning rate $\mu$, time distribution $\tau$, adaptive weighting $(\eta, p)$, JVP approximation step size $\epsilon$, class labels $c$.
        \Repeat
            \State $z \sim p_Z,\quad x \sim p_X,\quad t,s \gets \tau$
            \State $x_t\gets \alpha_t x + \sigma_t z,\quad v_t\gets \alpha'_t x + \sigma'_tz$
            \State $F_{t,s}\gets F_\theta(x_t; t, s, c),\quad F_{t,t}\gets F_\theta(x_t; t, t, c)$
            \State $F'_{t,s}\gets \left[F_\theta(x_t + \epsilon F_{t,t}; t + \epsilon, s, c) - F_\theta(x_t - \epsilon F_{t,t};t - \epsilon, s, c)\right] / (2\epsilon)$ \COMMENTa{JVP-Approx.}
            \State $F_\mathrm{tgt}\gets   F_{t,s} + \left(A''_{t,s}x_t + A'_{t,s}(F_{t,t} - F_{t,s}) - A_{t,s}F'_{t,s}\right)$
            \State $\mathcal L \gets  \lambda_1\|F_{t,t} - v_t\|^2_2 + \lambda_2\|F_{t,s} - \operatorname{sg}[F_\mathrm{tgt}]\|^2_2 $  \COMMENTa{Optimization Target}
            \State $\theta\gets \theta-\mu\nabla_\theta \mathcal L$  \COMMENTa{Model Update}
        \Until Convergence
    \end{algorithmic}
\end{algorithm}
\begin{algorithm}[th]
    \caption{(iSD Sampling) Sampling algorithm of vanilla iSD}
    \label{alg:isd-sampling}
    \begin{algorithmic}
        \State {\bfseries Input:} Initial noise $z\sim p_Z$, trained model $F_\theta$, class labels $c$, sampling time steps $\{t_i\}_{i=1}^{N+1}$.
        \State $x\gets z$
        \For{$i\gets 1$ to $N$}
            \State $ x \gets \nu^{-1}(A'_{t_i,t_{i+1}}x - A_{t_i,t_{i+1}}F_\theta(x; t_i, t_{i+1}, c))$
        \EndFor
    \end{algorithmic}
\end{algorithm}

\subsection{Random Initialization instead of Preconditioners}

When constructing the iSD framework, we focus on random initializations rather than preconditioners. As shown in \cref{tab:reproducibility}, pretrained networks with better FIDs do not necessarily yield better FIDs in the resulting consistency models. With random initialization, one can explore different random seeds and retrain the network to obtain improved results. There also remains potential to improve performance using alternative initialization strategies, such as leveraging preconditioners. However, when adopting preconditioners as baselines, it becomes difficult to systematically design experiments to discover \textit{better preconditioners}: which data, architectures, and hyperparameters should be used, how to quantify the quality of preconditioners, and whether perturbing pretrained weights is necessary. Therefore, we conclude that random initialization makes the reproducibility problem more tractable, and we construct iSD as a randomly initializable framework. 

\subsection{Classifier-free Guidance of Flow Map Models}
\label{proof:cfg-fmm}

\textbf{Pre-CFG.} By abstracting the guiding trajectory to $v_\theta$, the flow map model can naturally be trained to follow the specific trajectory as long as it is Lipschitz continuous. To achieve this, we replace the guiding trajectory and refer to this training scheme as \textit{Pre-CFG}. We consider three variants of objectives: iSD-U, iSD-C, and iSD-T.

\textbf{iSD-U} (Guidance-\textbf{U}nconditional). Given a label $c$ and a null class label $\varnothing$, let the corresponding velocity fields be $F_\theta(x_t; t, t, c)$ and $F_\theta(x_t; t, t, \varnothing)$. If both are globally Lipschitz continuous, then the CFG trajectory $\tilde v_\theta(x_t; t, c)$ is also globally Lipschitz continuous, since any linear combination of Lipschitz continuous functions remains Lipschitz continuous.
\begin{align*}
    \tilde v_\theta(x_t; t, c) = F_\theta(x_t; t, t, \varnothing) + \omega(F_\theta(x_t; t, t, c) - F_\theta(x_t; t, t, \varnothing)).
\end{align*}
Thus, the flow map can be trained to follow the CFG velocity field. As a naive approach, we first train the network $F_\theta(x_t; t, s, c)$ to align with $\tilde v_\theta$, while leaving the remaining components unchanged. This shares a similar scheme with the Shortcut Model, which trains the marginal velocity and guides the shortcut mapping with a constructed CFG trajectory $\tilde v_\theta$.

In this case, however, we need to address a conflict: $\mathcal L_\mathrm{CFM}$ enforces $F_{t,t}\approx v^*_t$ while $\mathcal L_\mathrm{SD\text{-}R}$ enforces $F_{t,t}\approx \tilde v_{\theta}$.

\textbf{iSD-C} (Guidance-\textbf{C}onditional). To resolve this conflict, we append the guidance scale $\omega$ as an additional condition, $F_\theta(x_t; t, s, c, \omega)$. The modified objectives are given by:
\begin{align*}
    \mathcal L_\mathrm{CFM} &= \mathbb E\left[\|F_\theta(x_t; t, t, c, 1.0) - v_t(x_t|x)\|^2_2\right], \\
    \mathcal L_\mathrm{iSD\text{-}C} &= \mathbb E\left[\left\|F_\theta(x_t; t, s, c, \omega) - \operatorname{sg}[F_\mathrm{tgt}(x_t; t, s, c, \omega)]\right\|^2_2\right],
\end{align*}
where
\begin{align*}
    F_\mathrm{tgt} &= F_\theta(x_t; t, s, c, \omega) + \left(A''_{t,s}x_t + A'_{t,s}\left(\tilde v_\theta(x_t; t, c) - F_\theta(x_t; t, s, c, \omega)\right) - A_{t,s}F'_\theta(x_t; t, s, c, \omega)\right), \\
    \tilde v_\theta(x_t; t, c) &= F_\theta(x_t; t, t, \varnothing, 1.0) + \omega (F_\theta(x_t; t, t, c, 1.0) - F_\theta(x_t; t, t, \varnothing, 1.0)).
\end{align*}
Hence, $\mathcal L_\mathrm{CFM}$ ensures $F_\theta(x_t; t, t, c, 1.0)\approx v^*_t(x_t; c)$, while $\mathcal L_\mathrm{iSD\text{-}C}$ ensures $F_\theta(x_t; t, t, c, \omega)\approx \tilde v_\theta(x_t; t, c)$. This choice is natural, as $\tilde v_\theta(x_t; t, c) \approx v^*_t(x_t; c)$ when $\omega = 1$.

\textbf{iSD-T} (\textbf{T}raining-time CFG). To resolve the conflict from another perspective, we adopt the training-time CFG described in \cref{proof:ttcfg}. As proven there, the training target $\tilde v_t(x_t|x) = (1-\omega) F_{\theta^-}(x_t; t, t,\varnothing) + \omega v_t(x_t|x)$ leads the flow matching objective toward the CFG velocity field. Accordingly, we modify the flow matching objective to follow the CFG velocity, distilling the approximated velocity into flow mappings:
\begin{align*}
    \mathcal L_\mathrm{CFM}^\omega &= \mathbb E\left[\left\|F_\theta(x_t; t, t, c) - \tilde v_t(x_t|x)\right\|^2_2\right], \\
    \mathcal L_\mathrm{iSD\text{-}T} &= \mathbb E\left[\left\|F_\theta(x_t; t, s, c) - \operatorname{sg}\left[F_\mathrm{tgt}(x_t; t, s, c)\right]\right\|^2_2\right],
\end{align*}
where $v_\theta(x_t; t) = F_\theta(x_t; t, t, c)$. In this case, the network approximates the CFG velocity, and the resulting flow mappings naturally follow the CFG trajectory. Unlike iSD-U, iSD-T preserves the objective alignment at the boundary $s=t$. We therefore adopt iSD-T as the default configuration for class-conditional experiments. The ablation studies are provided in \cref{appendix:add-expr}, and the training algorithm is described in \cref{alg:isdt-training}.

\textbf{Post-CFG.} However, \textit{Post-CFG}, defined as
\begin{align*}
    \tilde F_\theta(x_t; t, s, c) = F_\theta(x_t; t, s, \varnothing) + \omega(F_\theta(x_t; t, s, c) - F_\theta(x_t; t, s, \varnothing)),
\end{align*}
following \cref{alg:postcfg}, does not generally follow the exact CFG trajectory. This discrepancy arises from the definition of the flow map,
\begin{align*}
f_\theta(x_t; t, s) \approx x_t + \int_t^sv^*_\tau(x_\tau)d\tau,
\end{align*}
which performs the path integral along a specific trajectory induced by $v^*_\tau$. For a CFG trajectory, the path integral should be taken along $\tilde v_\tau^*$. In contrast, Post-CFG computes two separate forward passes, integrating along $v_\tau^*(x_\tau; c)$ and $v_\tau^*(x_\tau; \varnothing)$, rather than along $\tilde v_\tau^*$. As a result, the resulting update can differ from the expected CFG trajectory.
\begin{algorithm}[th]
    \caption{(iSD-T Training) Training algorithm of iSD-T}
    \label{alg:isdt-training}
    \begin{algorithmic}
        \State {\bfseries Input:} Noise distribution $p_Z$, data distribution $p_X$, model $F_\theta$, learning rate $\mu$, time distribution $\tau$, adaptive weighting $(\eta, p)$, JVP approximation step size $\epsilon$, Pre-CFG scale $\omega$, class labels $c$.
        \Repeat
            \State $z \sim p_Z,\quad x \sim p_X,\quad t, s\gets \tau$
            \State $x_t\gets \alpha_t x + \sigma_t z,\quad v_t\gets \alpha'_tx + \sigma'_tz$
            \State $\tilde v_t \gets \omega v_t + (1-\omega)\operatorname{sg}[F_{\theta}(x_t; t, t, \varnothing)]$
            \State $F_{t,s}\gets F_\theta(x_t; t, s, c),\quad F_{t,t}\gets F_\theta(x_t; t, t, c)$
            \State $F'_{t,s}\gets \left[F_\theta(x_t + \epsilon F_{t,t}; t + \epsilon, s, c) - F_\theta(x_t - \epsilon F_{t,t}; t - \epsilon, s, c)\right] / (2\epsilon)$ \COMMENTa{JVP-Approx.}
            \State $F_\mathrm{tgt}\gets F_{t,s} + \left(A''_{t,s}x_t + A'_{t,s}(F_{t,t} - F_{t,s}) - A_{t,s}F'_{t,s}\right)$
            \State $\mathcal L \gets \lambda_1\|F_{t,t} - \tilde v_t\|^2_2 + \lambda_2\|F_{t,s} - \operatorname{sg}[F_\mathrm{tgt}]\|^2_2$  \COMMENTa{Optimization Target}
            \State $\theta\gets \theta-\mu\nabla_\theta \mathcal L$  \COMMENTa{Model Update}
        \Until Convergence
    \end{algorithmic}
\end{algorithm}
\begin{algorithm}[ht]
    \caption{(Post-CFG Sampling) Sampling algorithm of iSD with Post-CFG}
    \label{alg:postcfg}
    \begin{algorithmic}
        \State {\bfseries Input:} Initial noise $z\sim p_Z$, model $F_\theta$, Post-CFG scale $\omega$, class labels $c$, sampling steps $\{t_i\}_{i=1}^{N+1}$.
        \State $x\gets z$
        \For{$i\gets 1$ to $N$}
            \State $\tilde F_{t,s}\gets (1-\omega )F_{\theta} (x;t_i,t_{i+1},\varnothing)+\omega F_\theta(x;t_i,t_{i+1},c)$ \COMMENTa{Post-CFG}
            \State $ x \gets \nu^{-1}(A'_{t_{i}, t_{i+1}} x - A_{t_i, t_{i+1}} \tilde F_{t,s})$
        \EndFor
    \end{algorithmic}
\end{algorithm}

\section{Experimental Details}
\subsection{Reproducibility of Consistency Training}
\label{detail:reproducing-ct}
To evaluate the reproducibility of consistency training, we conducted experiments within the UCGM~\citep{ucgm} framework. We compared the FID scores of several models trained under different initialization conditions. Following UCGM, we first extract latent representations of ImageNet-1K $256\times 256$ using VA-VAE~\citep{lightningdit}. All few-step models were trained with the same hyperparameters and settings: RAdam optimizer with a learning rate of 1e-4, weight decay of 0.0, $\beta_1=0.9,\ \beta_2 = 0.999$, batch size of 1024, gradient clipping at 0.1, and timestep $t$ sampled from Beta(0.8, 1.0). For enhancement, we applied a label drop ratio of 0.1, an enhancement range of $(0, 0.75)$, and a ratio of 2.0. We also used the cosine function as the loss weighting function and trained all models with linear interpolation for 40K iterations.

\textbf{Multistep Baseline.} We trained the DiT-XL/1 architecture initialized from the publicly released multistep checkpoint of UCGM. This configuration achieved a 2-step FID of 2.52, which is reasonable but still falls short of the reported FID of 1.42.

\textbf{LightningDiT.} We trained the LightningDiT-XL/1 architecture from its released pretrained model. In this setting, the model achieved a 2-step FID of 9.59, which is worse than the reported FID.

\textbf{In-house Multistep Model.} We trained the DiT-XL/2 architecture from scratch. For training, we used AdamW~\citep{adamw}  with a learning rate of 0.0002, $\beta_1 = 0.9,\ \beta_2 = 0.95$, EMA decay weight of 0.999, and timestep $t$ sampled from Beta(1, 1). We used an enhancement ratio of 0.47 and a cosine weighting function. After training the multistep model for 800K iterations, we conducted consistency training under the same few-step settings. This resulted in a 2-step FID of 5.78.

\begin{figure}[th]
\centering
\subfloat{
\includegraphics[width=0.31\linewidth]{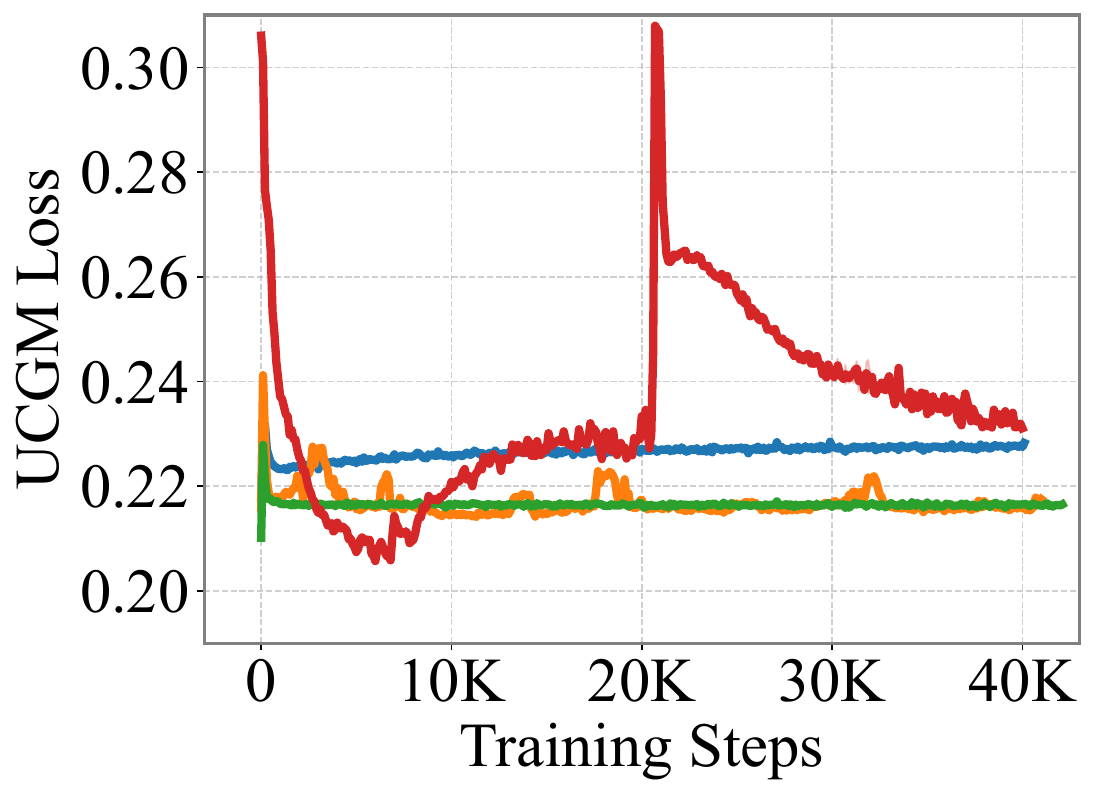}
}
\subfloat{
\includegraphics[width=0.31\linewidth]{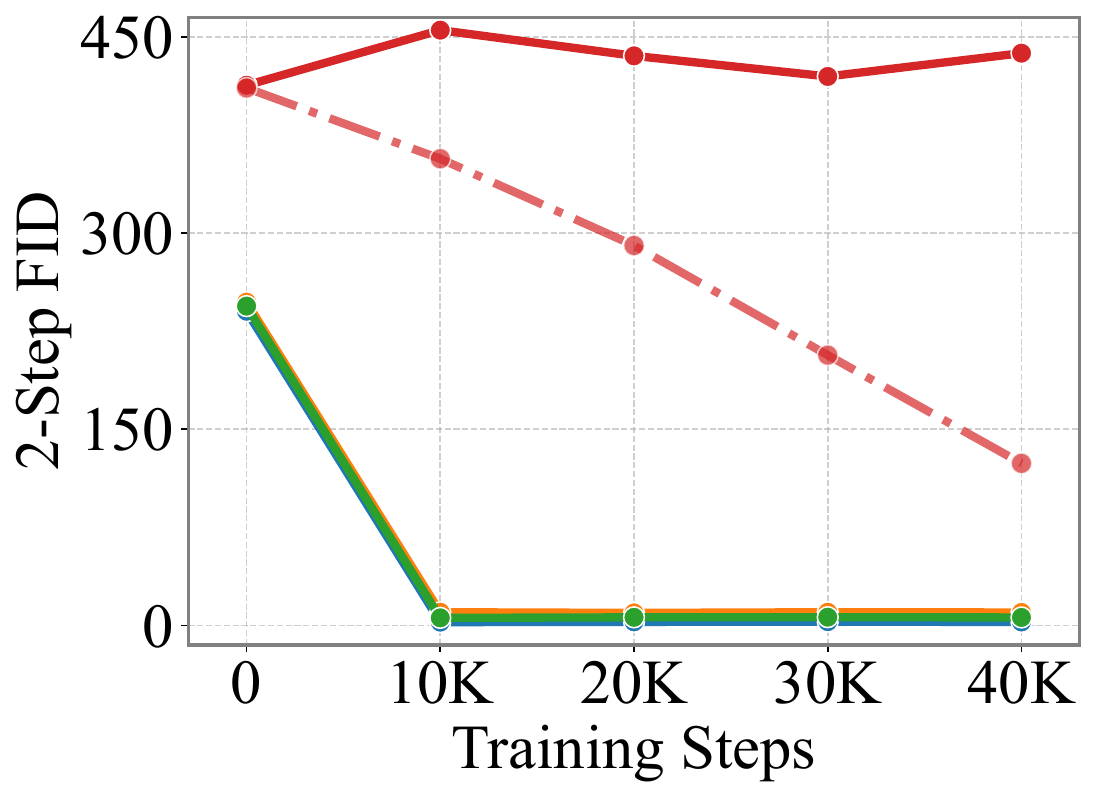}
}
\subfloat{
\includegraphics[width=0.31\linewidth]{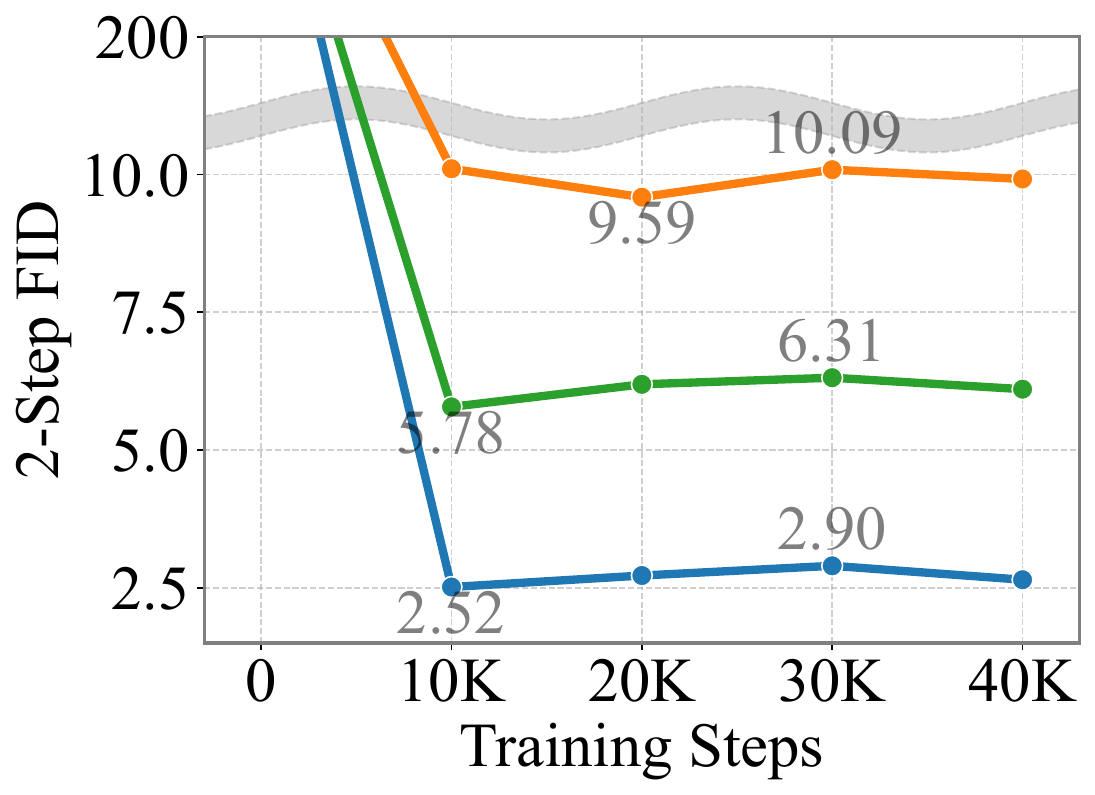}
}

\begin{minipage}{0.75\linewidth}
    \includegraphics[width=\linewidth]{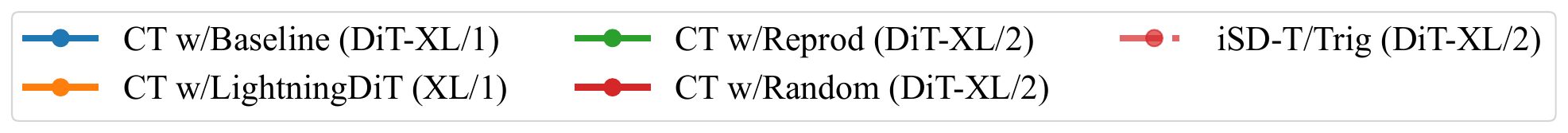}
\end{minipage}
\caption{Training curves of consistency training. (a) Training loss over training steps. (b) Two-step FIDs over training steps. (c) Rescaled Y-axis for the consistency model with a preconditioner.}
\label{fig:instability-fig}
\end{figure}

\textbf{Without Preconditioner.} We train a DiT-XL/2 architecture from randomly initialized weights without any preconditioner. In this case, training consistently failed, with the loss diverging and no meaningful samples being generated. While 40K steps may appear insufficient for scratch training, other scratch training methods already show a rapid decrease by 40K steps (\cref{fig:instability-fig}). We consider that this is enough to check the unstable dynamics compared to other models.

These results suggest that consistency training can be sensitive to initialization and the choice of preconditioner. It becomes unstable under random initialization and appears to substantially benefit from a well-trained multistep model. Even with pretrained initializations, consistency training demonstrates limited robustness and reproducibility across different architectures and setups.

\subsection{Implementation}
\label{detail:implementation}

\begin{table}[th]
\centering
\caption{Experimental Settings.}
\label{tab:experimental-setting}
\resizebox{1.0\linewidth}{!}{%
\begin{tabular}{lccccccc}
\toprule
Dataset & \multicolumn{6}{c}{ImageNet-1K $256\!\times\!256$} & CelebA-HQ $256\!\times\!256$ \\
\midrule
Preprocessor & \multicolumn{4}{c}{SD-VAE~\citep{ldm}} & SD-VAE & VA-VAE~\citep{lightningdit} & SD-VAE \\
Input size & \multicolumn{4}{c}{$32\times 32\times 4$} & $32\times32\times4$ & $16\times 16\times 32$ & $32\times 32\times 4$ \\
Condition & \multicolumn{4}{c}{Class-conditional} & Conditional & Conditional & Unconditional \\
\midrule
Backbone & DiT-B/4 & DiT-B/2 & DiT-L/2 & DiT-XL/2 & DiT-XL/2$^\dagger$ & DiT-XL/1$^\dagger$ & DiT-B/2 \\
\midrule
Params (M)  & 131.33 & 131.33 & 459.42 & 676.78 & 682.02 & 681.77 & 131.33 \\
Depth       & 12 & 12 & 24 & 28 & 28 & 28 & 12 \\
Hidden dim  & 768 & 768 & 1024 & 1152 & 1152 & 1152 & 768 \\
Heads       & 12 & 12 & 16 & 16 & 16 & 16 & 12 \\
Patch size  & $4{\times}4$ & $2{\times}2$ & $2{\times}2$ & $2{\times}2$ & $2{\times}2$ & $1{\times}1$ & $2{\times}2$\\
\midrule
Interpolation & \multicolumn{4}{c}{Trigonometric} & Linear & Linear & Trigonometric \\
Self-distillation & \multicolumn{4}{c}{$\mathcal L_\mathrm{iSD\text{-}T}$} & $\mathcal L_\mathrm{iSD\text{-}T}$ & $\mathcal L_\mathrm{iSD\text{-}T}$ & $\mathcal L_\mathrm{iSD}$ \\
Joint training & \multicolumn{4}{c}{Enabled} & Enabled & Enabled & Enabled \\
JVP & \multicolumn{4}{c}{Approximation} & Approx. & Approx. & Approx. \\
Weighting & \multicolumn{4}{c}{Adaptive} & Adaptive & Cosine & Adaptive \\
$\epsilon$ & \multicolumn{4}{c}{0.005} & 0.005 &  0.005 & 0.005 \\
$p$ & \multicolumn{4}{c}{1.0} & 1.0 & - & 1.0 \\
$\eta$ & \multicolumn{4}{c}{0.01} & 0.01 & - & 0.01 \\
Pre-CFG $\omega$ & \multicolumn{4}{c}{5.0} & Randomly sampled & Randomly sampled & - \\
\midrule
FP Precision & \multicolumn{4}{c}{BF16} & BF16 & BF16 & BF16 \\
Batch size  & \multicolumn{4}{c}{256} & 256 & 256 & 256 \\
Label dropout & \multicolumn{4}{c}{0.1} & 0.1 & 0.1 & - \\
Optimizer   & \multicolumn{4}{c}{AdamW~\citep{adamw}} & NorMuon~\citep{normuon} & NorMuon & AdamW \\
LR Scheduler   & \multicolumn{4}{c}{Constant} & Constant & Constant & Constant \\
$\beta_1$   & \multicolumn{4}{c}{0.9} & - & - & 0.9 \\
$\beta_2$  & \multicolumn{4}{c}{0.999} & 0.95 & 0.95 & 0.999 \\
Momentum & \multicolumn{4}{c}{-} & 0.95 & 0.95 & - \\
Learning rate & \multicolumn{4}{c}{1e-4} & 5e-4 & 5e-4 & 1e-4 \\
Training steps & \multicolumn{3}{c}{400K} & 800K & 800K & 600K & 200K \\
Weight decay & \multicolumn{4}{c}{0} & 0 & 0 & 0 \\
EMA decay   & \multicolumn{4}{c}{0.99995} & 0.99995 & 0.99995 & 0.99995 \\
\bottomrule
\end{tabular}
}
\end{table}

\textbf{ImageNet 256$\times$256.} SD-VAE~\citep{ldm} was used to encode images into a $32\times32\times4$ latent representation. For DiT~\citep{dit} models, we employed RMSNorm~\citep{rmsnorm}, QK normalization~\citep{qknorm}, and RoPE~\citep{rope}. Each model was scaled by depth and hidden dimension. We sampled $t$ and $s$ independently from Beta(0.8, 1.0), and set $t, s := \max(t, s), \min(t, s)$. For trigonometric interpolation, we additionally scale the timesteps as $t \gets \frac{\pi}{2}t$ and $s\gets \frac{\pi}{2}s$. For generations, we uniformly sampled the intermediate timesteps from the interval without additional engineering.

In ablation studies, DiT-B/4 was trained for 400K steps with a batch size of 256 and BF16 precision. For iSD-T, the corresponding wall-clock time is 32 GPU hours on a single A100 GPU. For large-scale experiments, DiT-XL/2 was trained for up to 800K steps, corresponding to 216 A100 GPU hours. We note that prior work~\citep{ucgm, imm, meanflow, facm} reports training from scratch for 800 or more epochs, corresponding to 1M or more steps with a larger batch size (e.g., 1024).

We also report results obtained with additional performance-oriented architectural choices. Specifically, we augment the input space for the guidance scale and the CFG range by sampling them during training rather than using fixed values, following the design choices introduced in \citet{imf}. The guidance scale is sampled from a truncated exponential distribution, and the CFG range is sampled from a uniform distribution. These modifications are orthogonal to our analysis and are used only to demonstrate that iSD-T remains stable under commonly adopted performance-oriented settings. For conditioning, we concatenate conditions in the input token space. We denote this configuration with SD-VAE by DiT-XL/2$^\dagger$ and VA-VAE by DiT-XL/1$^\dagger$. Detailed training parameters are provided in \cref{tab:experimental-setting}, and the corresponding results in \cref{tab:imagenet-ditxl} are reported under this setting.

\textbf{CelebA-HQ.} For CelebA-HQ, we use SD-VAE to encode images into $32\!\times\!32\!\times\!4$ latent representations. The model is based on DiT-B/2 in an unconditional setting, as in the Shortcut Model. It is trained on four RTX 3090 GPUs with a global batch size of 256. It takes 1.83 days for 200K steps (176 RTX 3090 GPU hours). Detailed settings are provided in \cref{tab:experimental-setting}.

\textbf{The Choice of Interpolation} reflects different geometric assumptions. Linear interpolation is aligned with the optimal transport path under a Euclidean cost. This is well-suited for tasks such as Push-T (\cref{appendix:policy-generation}), where the end-effector coordinates are defined in Euclidean space. In contrast, trigonometric interpolation is associated with a geodesic on a spherical manifold. It is more appropriate for tasks involving normalized representations~\citep{chen2024flow}, such as image generation in certain latent spaces.

We note that these choices are not strict requirements. In practice, sufficiently expressive networks can compensate for geometric mismatch~\citep{zheng2026diffusion}. Therefore, we view the choice of interpolation primarily as an inductive bias, rather than a hard constraint.

\begin{figure}[th]
\centering
\subfloat[Gradient norm (log-scale)]{
\includegraphics[width=0.31\linewidth]{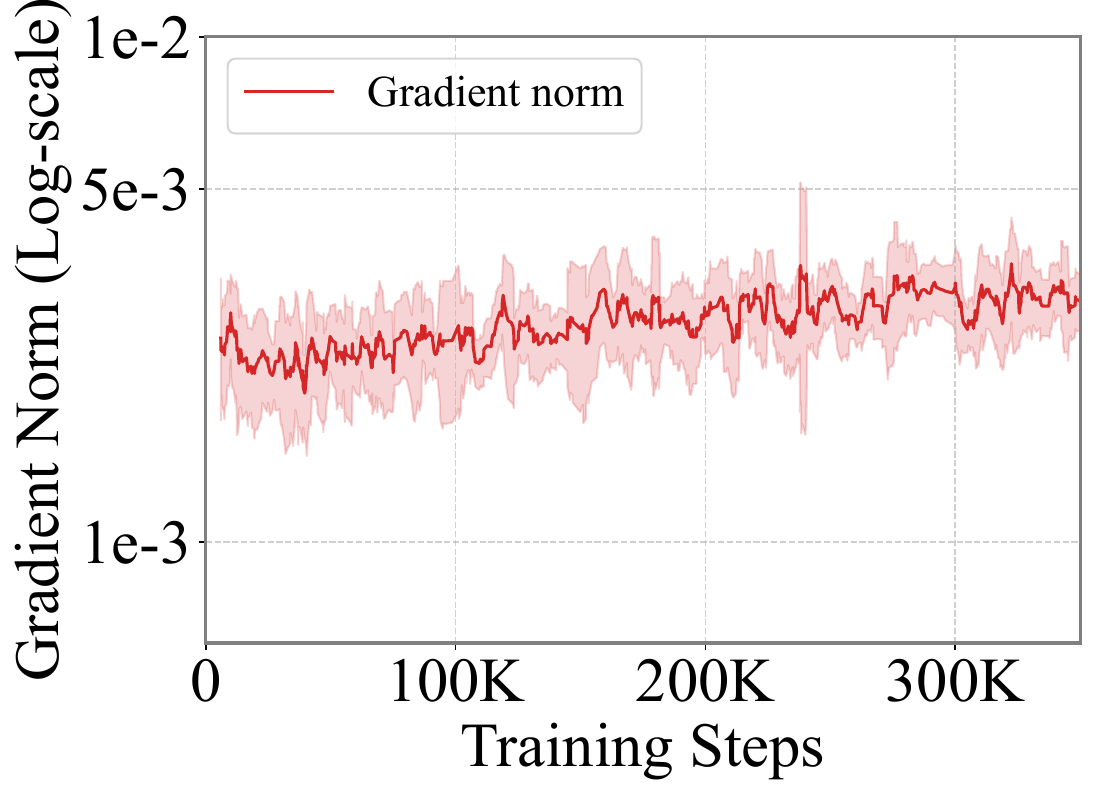}
}
\subfloat[Flow matching loss $\mathcal L_\mathrm{CFM}^\omega$ in \cref{eq:training-time-cfg}]{
\includegraphics[width=0.31\linewidth]{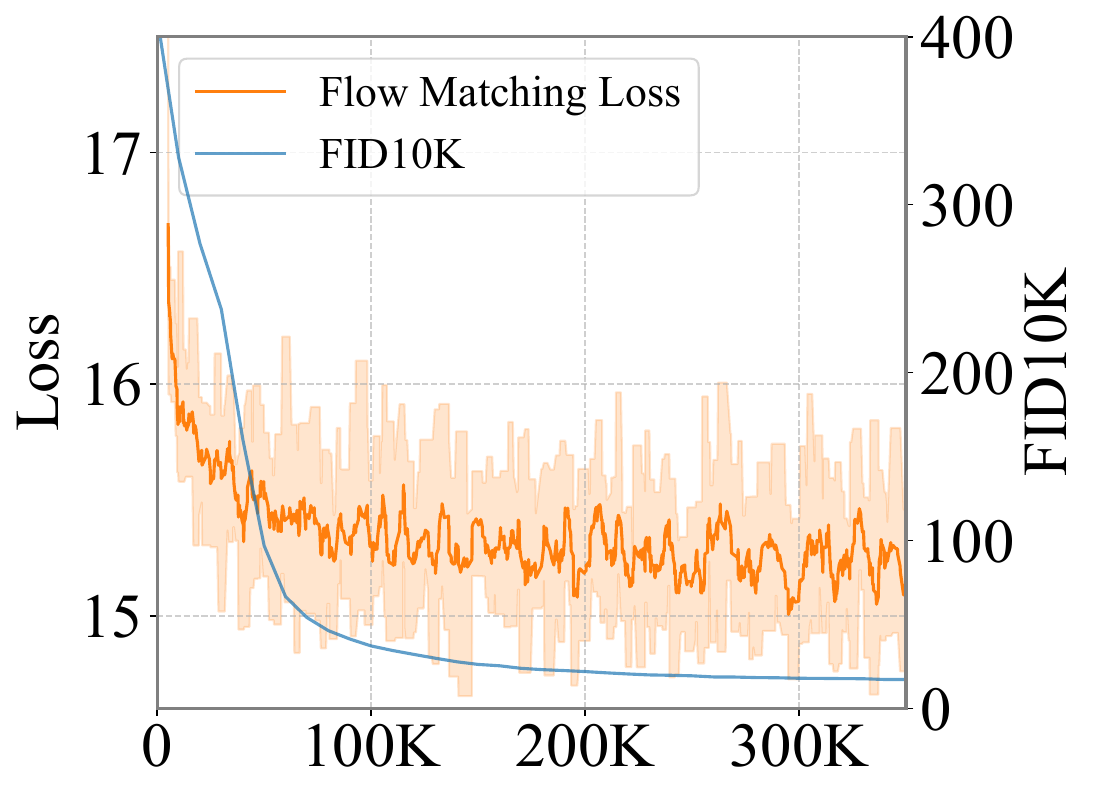}
\label{fig:loss-curve-cfm}
}
\subfloat[SD-R loss $\mathcal L_\mathrm{SD\text{-}R}$ in \cref{eq:sd-r}]{
\includegraphics[width=0.31\linewidth]{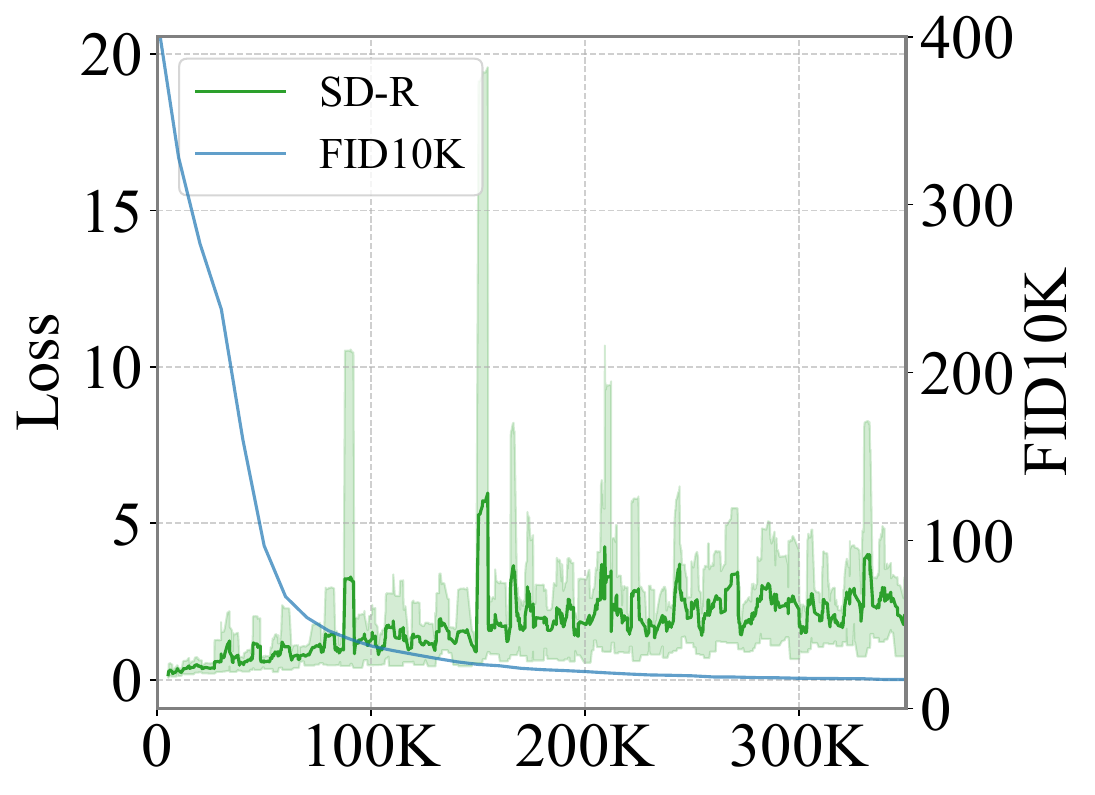}
\label{fig:loss-curve-sd-r}
}
\caption{Gradient norm and loss curves of iSD-T compared to FID10K. FID and the flow matching loss decrease together over training, consistent with the role of velocity estimation in flow map training.}
\label{fig:loss-curve}
\end{figure}

\subsection{Additional Ablation Study}
\label{appendix:add-expr}

\begin{table}
\centering
\captionsetup{position=top}
\caption{Direct intervention on guiding velocity by interpolating marginal velocity ($\gamma=0$) and conditional velocity ($\gamma=1$). The marginal velocity guidance shows lower FID and lower loss variance. (Left) Toy experiments following \cref{fig:toy}. We measure the optimality of flow maps using the proxy measure $\mathcal L_\mathrm{ED}$. (Right) Experiments in the ImageNet ablation setting. We report FIDs and the standard deviations $\sigma$ of each loss landscape, following \cref{fig:landscape-main}.}
\subfloat[$\mathcal L_\mathrm{ED}$ in toy experiments]{%
\label{tab:gamma-sweep-toy}
\begin{tabular}{c|ccccc}
\toprule
$\gamma$ & 1.00 & 0.75 & 0.50 & 0.25 & 0.00 \\
\midrule
$\mathcal L_\mathrm{ED}$ & 0.34 & 0.24 & 0.17 & 0.12 & 0.10 \\
$\mathcal L_\mathrm{CT}$ & 0.16 & 0.15 & 0.15 & 0.12 & 0.10 \\
\bottomrule
\end{tabular}
}\hspace{0.01\linewidth}
\subfloat[FID and $\sigma$ in ablation settings]{%
\label{tab:gamma-sweep-dit}
\begin{tabular}{c|ccc}
\toprule
$\gamma$ & 1.0 & 0.5 & 0.0 \\
\midrule
FID & 120.49 & 112.42 & 110.38 \\
$\sigma$ & 204.07 & 92.43 & 61.94 \\
\bottomrule
\end{tabular}
}
\end{table}

\textbf{Direct Intervention.} To demonstrate the role of guiding velocities, we directly perturb the guiding velocity and observe the resulting flow maps. We use marginal velocity as the baseline guidance, and control a proxy for the noise level by defining $v^\gamma_t(x_t) = (1-\gamma)v^*_t(x_t) + \gamma v_t(x_t|x)$, interpolating between marginal velocity $v^*_t$ and noisy conditional velocity $v_t(x_t|x)$.

In a toy setting, we measure optimality as $\mathcal L_\mathrm{ED}$ with $v^*_t$ following \cref{fig:toy-ed-curve}, as shown in \cref{tab:gamma-sweep-toy}. For Eulerian distillation $\mathcal L_\mathrm{ED}$ and consistency training $\mathcal L_\mathrm{CT}$ along $v^\gamma_t$, optimality improves monotonically as $\gamma$ decreases, i.e., as the guiding velocity becomes closer to the marginal velocity. On ImageNet-1K with DiT-B/4, reducing $\gamma$ improves FID. We measure instability via the standard deviation of each loss landscape, and it also decreases as $\gamma$ decreases.

This provides empirical evidence consistent with the view: the gradient difference induced by the guiding velocity can affect optimization and is associated with suboptimality and instability (\cref{prop:suboptimality,prop:instability}).

\begin{figure}[th]
\centering
\subfloat[Loss]{
\includegraphics[width=0.31\linewidth]{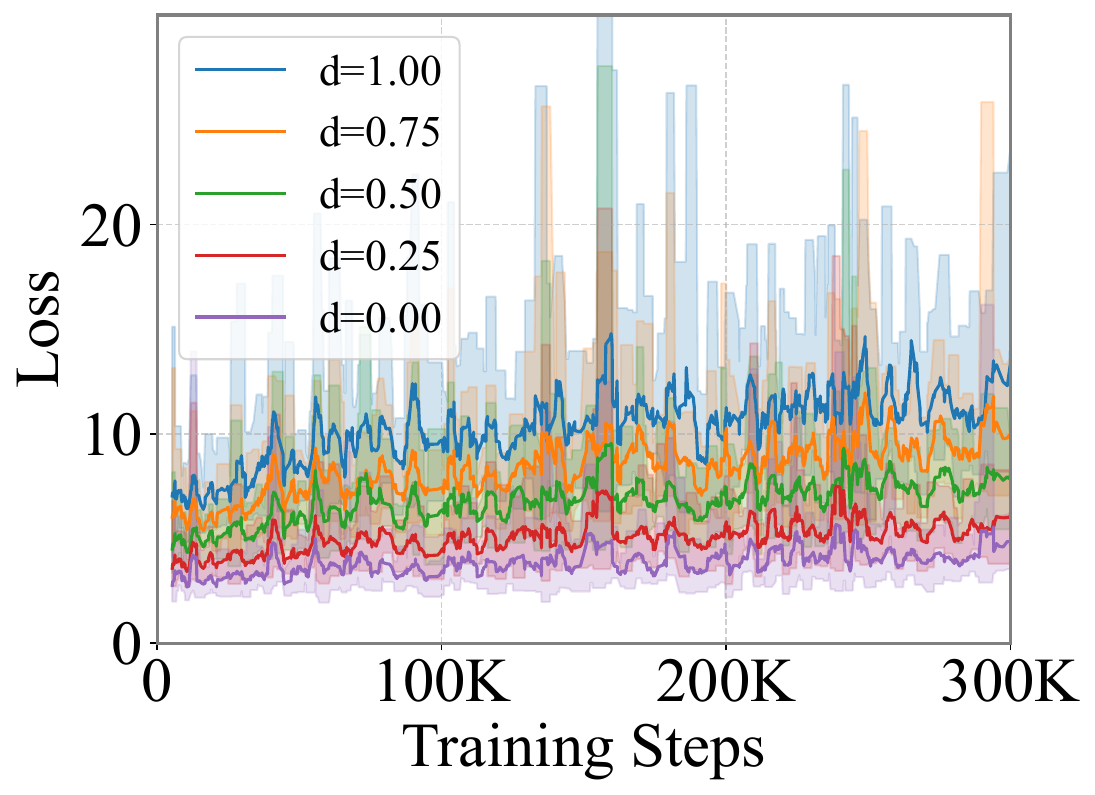}
}
\subfloat[Gradient norm (log-scale)]{
\includegraphics[width=0.31\linewidth]{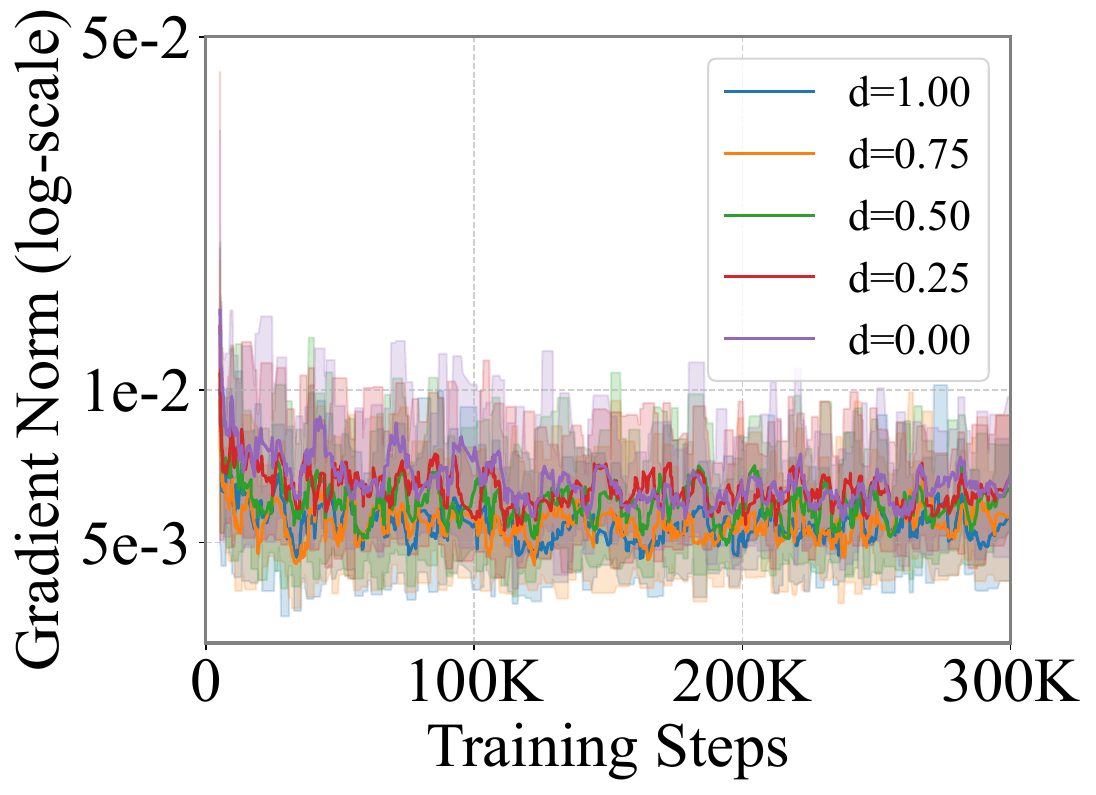}
}
\subfloat[Quantitative metrics]{
\includegraphics[width=0.31\linewidth]{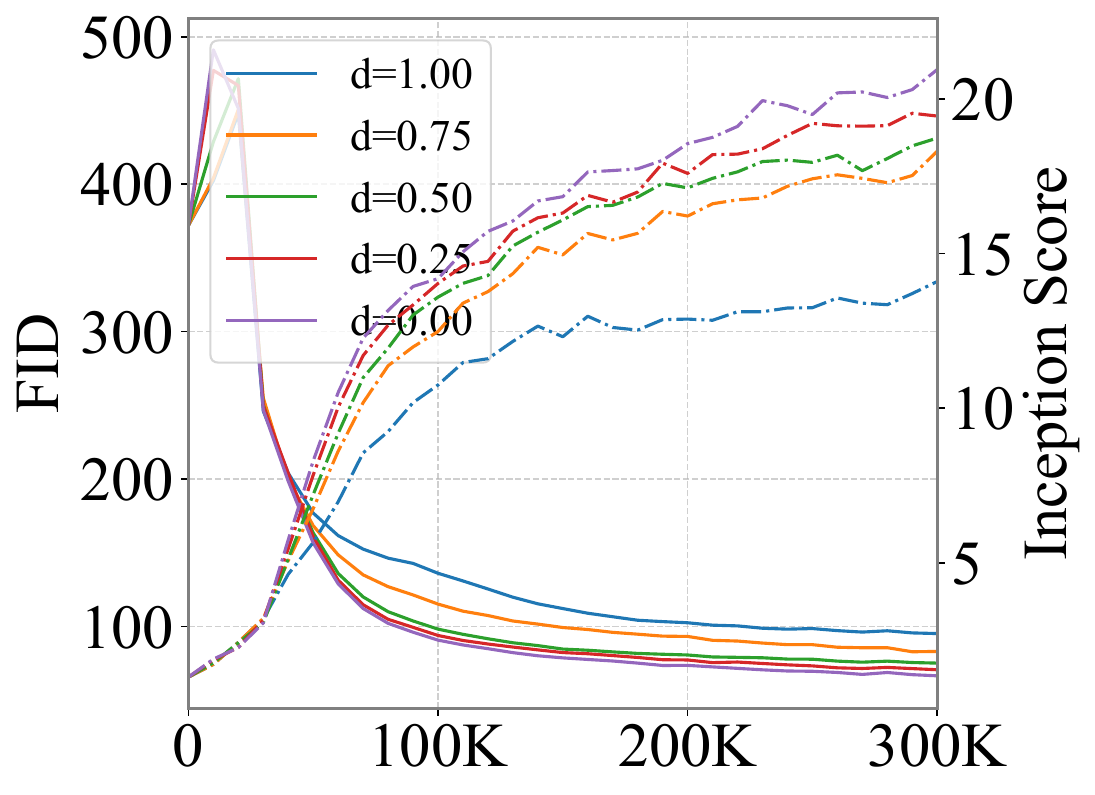}
}
\caption{Gradient norm and loss curves of consistency training, varying step size parameter $d$ in \cref{eq:step-size-ablation}. Smaller $d$ values are associated with lower loss values and better quantitative metrics in this setting.}
\label{fig:step-size-ablation}
\end{figure}

\textbf{Step Size.} To examine the role of the JVP term from the linearization-cost perspective, we vary the step size parameter $d$. We sample $s$ as follows:
\begin{align}\label{eq:step-size-ablation}
    s = u\cdot \max\{t - d, 0\},\quad t\sim \mathrm{Beta}(0.8, 1.0),\quad u\sim \mathcal U[0, 1),
\end{align}
which ensures $t - s \ge d$ when $t \ge d$. Accordingly, $d=0$ recovers the relaxed time condition $s<t$, whereas $d=1$ recovers fixed $s=0$; thus, varying $d$ interpolates between the two time-condition setups. We conduct experiments under the ablation setting, using consistency training with conditional velocity guidance and joint training with flow matching. We report loss values, gradient norms, and quantitative metrics while varying $d\in\{0, 0.25, 0.5, 0.75, 1\}$.

As shown in \cref{fig:step-size-ablation}, the distillation loss increases with $d$, consistent with an increased contribution of the term $(t-s)\cdot dF_\theta(x_t; t, s)/dt$. Smaller $d$ values are also associated with better quantitative metrics. These observations are consistent with our hypothesis that a linearization cost proportional to the step size may introduce additional optimization complexity. The gradient norms remain at comparable levels across the sweep, with $d=0$ being slightly higher than the other settings. We note that these results are limited to a single-seed evaluation, and we leave validation across multiple runs for future work.

\begin{table}[th]
\centering
\captionsetup{position=top}
\caption{Ablation study for hyperparameters. We set $\epsilon = 0.005$, $\eta = 0.01$, $p=1.0$, and NFE=2 by default.}
\label{tab:hparams-ablation}
\subfloat[JVP approximation step size $\epsilon$]{%
\begin{tabular}{c|cccc}
\toprule
$\epsilon$ & 0.01 & 0.005 & 0.001 & 0.0001 \\
\midrule
FID & 14.74 & 14.62 & 15.03 & 29.29 \\
\bottomrule
\end{tabular}
}\hspace{0.01\linewidth}
\subfloat[Adaptive weighting bias $\eta$]{%
\begin{tabular}{c|ccccc}
\toprule
$\eta$ & 1.0 & 0.1 & 0.01 & 0.001 & 0.0001 \\
\midrule
FID & 14.68 & 14.58 & 14.62 & 14.34 & 14.53 \\
\bottomrule
\end{tabular}
}

\subfloat[Adaptive weighting exponent $p$]{%
\begin{tabular}{c|ccccc}
\toprule
$p$ & 0.0 & 0.5 & 1.0 & 1.5 & 2.0 \\
\midrule
FID & 31.85 & 18.03 & 14.62 & 13.66 & 13.49 \\
\bottomrule
\end{tabular}
}\hspace{0.01\linewidth}
\subfloat[Sampling steps (NFE)]{%
\begin{tabular}{c|cccc}
\toprule
NFE & 1 & 2 & 4 & 8 \\
\midrule
FID & 24.18 & 14.62 & 14.31 & 14.43 \\
\bottomrule
\end{tabular}
}
\end{table}

\textbf{Hyperparameters.} In \cref{tab:hparams-ablation}, we explore hyperparameters in the ablation setting: iSD-T on DiT-B/4 with trigonometric interpolation and $\omega=5.0$. The default value of $\epsilon=0.005$ follows \citet{ucgm}, and default values of $\eta=0.01$ with $p=1.0$ follow \citet{meanflow}. We observe that FID remains stable for $\epsilon\in [0.001, 0.01]$, while extremely small values can degrade performance, likely due to numerical instability in lower-precision settings. For adaptive weighting parameters, performance also varies smoothly across $\eta$ and for nonzero $p$. We also observe that performance improves from 1 to 2 steps, and remains nearly plateaued.

This suggests that the method may not be highly sensitive within a reasonable range of hyperparameters. We use a single default setting for $\epsilon$, $\eta$, and $p$ across all experiments without additional tuning, including both image and policy generation. This indicates that the method may not require extensive hyperparameter search.

\textbf{JVP Approximation.} To assess the generality of JVP approximation, we additionally apply it to MeanFlow, beyond its use in iSD. Under the ablation setting, training MeanFlow with the exact JVP requires approximately 25GB of VRAM and 58 A100 GPU hours. With the approximation, the VRAM usage decreases to 11GB and the wall-clock time to 21.5 GPU hours. Under this setting, the FID decreases to 10.74. Note that in Fully Sharded Data Parallel setups, while integrating exact JVP may require additional engineering effort, this approximation can be incorporated with lower implementation overhead.

\textbf{Time-condition Annealing.} Since our training dynamics can be divided into two stages, learning marginal velocity and flow mapping guided by itself, we also consider annealing of the time condition $s$, sampled near $t$ for the initial stage and then warmed up to $0\le s<t$ as training continues. To linearly warm up the distribution of $s$, we first sample $s$ and $t$ following the baseline and reformulate $s$ as:
\begin{align*}
    s = t - (t-s)\times \min\left(1.0, \frac{\mathrm{steps}}{\mathrm{max\_warmup}}\right).
\end{align*}
We set \texttt{max\_warmup} to 10K steps, corresponding to the first knee point of the flow matching term. However, time-condition annealing does not noticeably affect the training dynamics, and the resulting FID of 29.65 is worse than the baseline.

To interpret this phenomenon, we consider two points: (i) we adopt adaptive weighting, which covers both flow matching and self-distillation objectives, and (ii) in the initial training stages, the loss of flow matching remains around 15, while that of self-distillation is near 0.5. This suggests that the flow matching objective already dominates the training signal, and thus the intended effect of annealing becomes diluted.

\begin{figure}[t]
\centering
\subfloat[Variants of Pre-CFG methods]{
\includegraphics[width=0.31\linewidth]{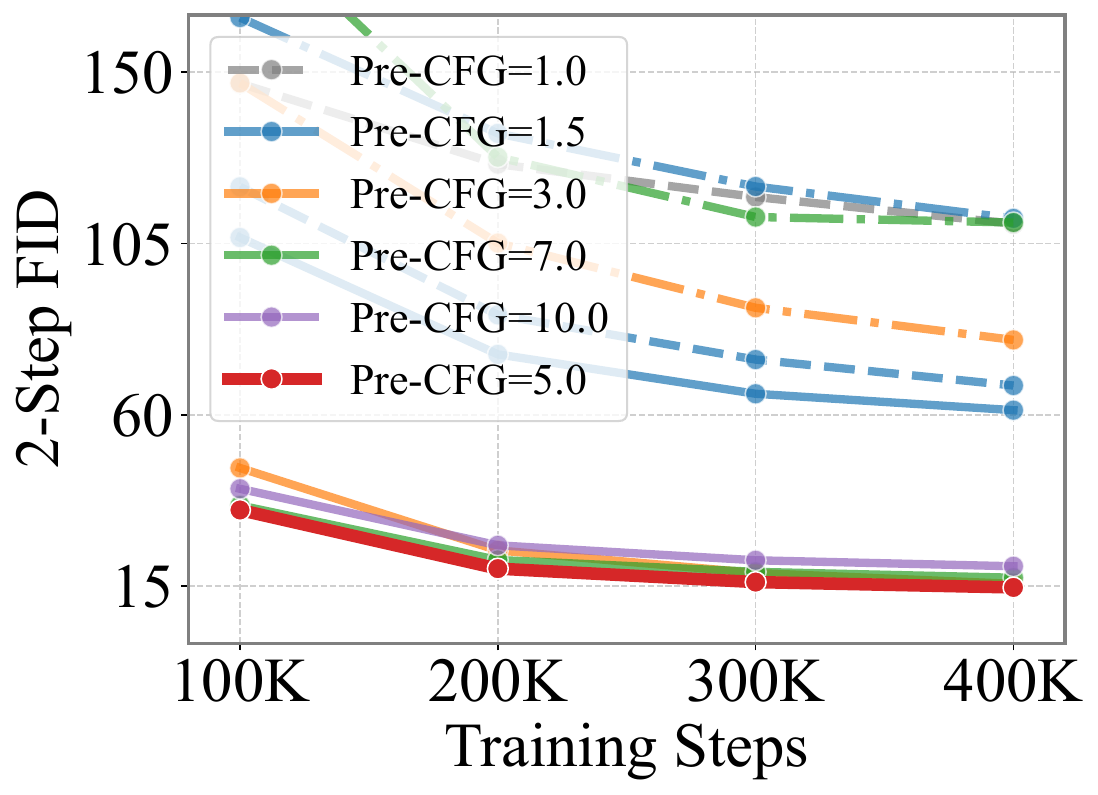}
\label{fig:precfg-variants}
}
\subfloat[Post-CFG scales]{
\includegraphics[width=0.31\linewidth]{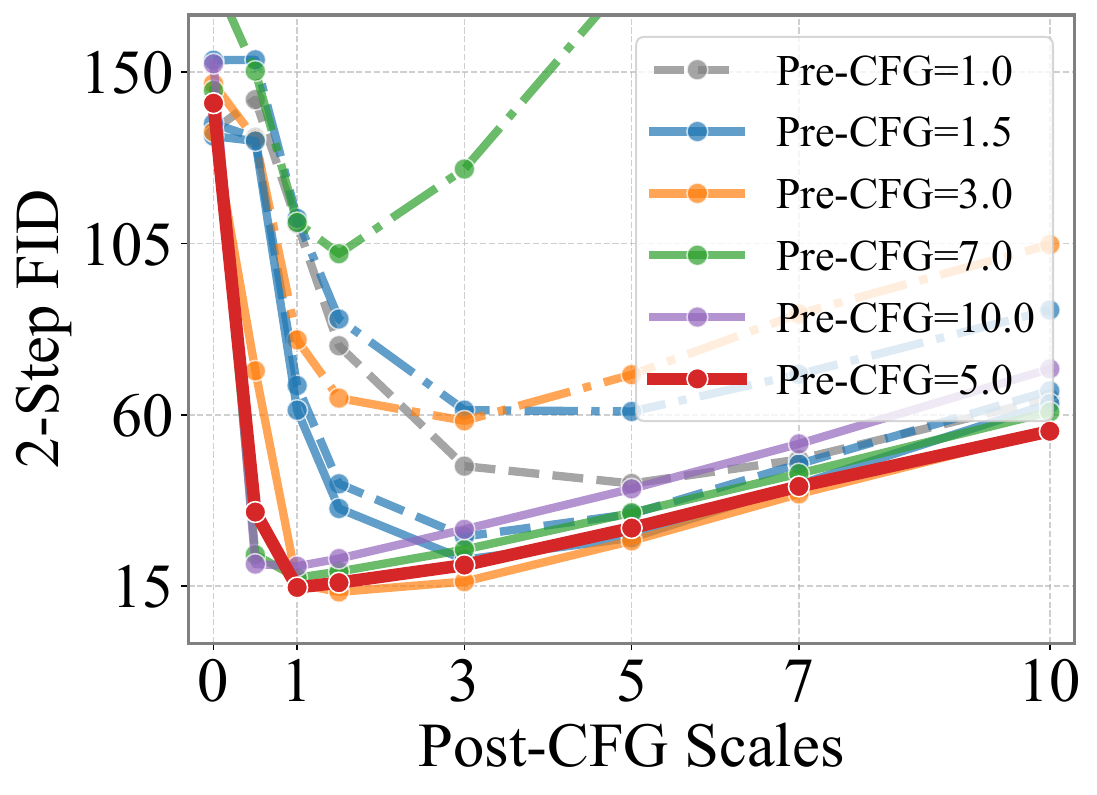}
\label{fig:postcfg-of-variants}
}
\subfloat[Multiplied CFG scales]{
\includegraphics[width=0.31\linewidth]{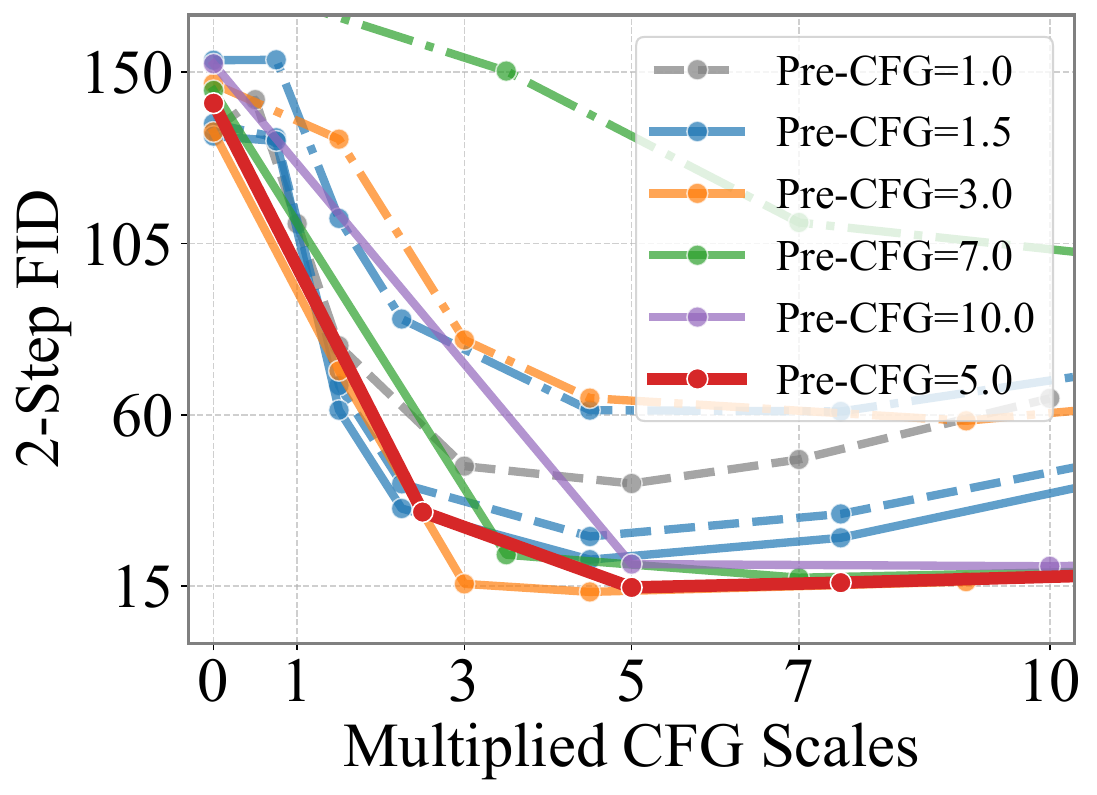}
\label{fig:multiplied-cfgs}
}
\caption{Variants of Pre-CFG. Dashed, dash-dot, and solid lines indicate iSD-U, iSD-C, and iSD-T, respectively. Each color represents a different Pre-CFG scale. Trigonometric interpolation and the JVP approximation are used by default on DiT-B/4. (a) FIDs of Pre-CFG variants over training steps. (b) FIDs of Post-CFG over guidance scales. (c) FIDs over multiplied guidance scales.}
\end{figure}

\textbf{Variants of Pre-CFG.} In the ablation setting, as shown in \cref{fig:precfg-variants} and \cref{tab:suppl-training}, applying Pre-CFG $\mathcal L_\mathrm{iSD\text{-}U}$ with $\omega = 1.5$ improves FIDs compared to vanilla $\mathcal L_\mathrm{iSD}$. However, training diverges when $\omega > 3.0$. This instability is consistent with the conflict discussed in \cref{proof:cfg-fmm}, where increasing the guidance scale can make the self-distillation target less aligned with the flow matching objective in iSD-U settings.

When the guidance scale is appended as a condition, $\mathcal L_\mathrm{iSD\text{-}C}$ enables training at higher guidance scales, outperforming $\mathcal L_\mathrm{iSD}$. However, $\mathcal L_\mathrm{iSD\text{-}C}$ consistently underperforms $\mathcal L_\mathrm{iSD\text{-}U}$. Intuitively, the additional condition forces the network to learn both CFG and non-CFG mappings, which may impose an extra burden on the network. Even though $\mathcal L_\mathrm{iSD\text{-}U}$ does not preserve objective alignment at the boundary $s=t$, few-step generation commonly assumes $s\ll t$, and iSD-U shows performance comparable to iSD-T in some evaluated settings. On the other hand, iSD-T neither introduces the conflict nor imposes an additional burden on the network. It consistently outperforms both variants. Therefore, we adopt iSD-T as the default for class-conditional experiments.

We also examined Post-CFG, as illustrated in \cref{fig:postcfg-of-variants}. iSD-U shows improved results when Post-CFG is applied, particularly at $\omega = 3.0$. Applying Post-CFG on top of Pre-CFG can be interpreted as applying CFG twice, leading to a multiplied guidance scale $4.5 = 1.5\times3.0$. Similarly, iSD-T achieves its lowest FIDs around a Pre-CFG scale of $\omega=5.0$, while Post-CFG on it underperforms even when the scale is close to one. When applying iSD-T with $\omega=3.0$, Post-CFG outperforms at $\omega=1.5$, corresponding to a total scale of 4.5. \textit{These consistent observations suggest that the effective CFG scale lies near 5.0 in this ablation setting.} We refer to this empirical tendency as effective-scale behavior.

We note that iSD-C shows effective CFG scales between 7.5 and 10.5, distinct from the $\omega$-unconditional settings (iSD-U and iSD-T). However, even within its optimal range, iSD-C underperforms them.

\textbf{Pre-CFG Scale for Larger Networks.} For iSD-T, we investigate the effective guidance scale for the larger network, DiT-XL/2, as shown in \cref{tab:suppl-training}. We train DiT-XL/2 with $\omega=5.0$ and measure metrics across Post-CFG scales (\cref{tab:ditxl2-postcfg}).

\begin{table}[th]
\centering
\caption{FIDs and ISs across Post-CFG scales for DiT-XL/2 with iSD-T ($\omega=5.0$, 700K steps).}
\label{tab:ditxl2-postcfg}
\begin{tabular}{c|cccccccc}
\toprule
Post-CFG $\omega$ & 0.0 & 0.5 & 1.0 & 1.5 & 3.0 & 5.0 & 7.0 & 10.0 \\
\midrule
FID & 100.89 & \textbf{11.19} & 16.96 & 18.26 & 19.33 & 25.83 & 35.59 & 48.50 \\
IS & 13.53 & 164.48 & \textbf{234.85} & 229.35 & 181.94 & 116.25 & 85.75 & 65.36 \\
\bottomrule
\end{tabular}
\end{table}

DiT-XL/2 shows better results at the Post-CFG scale of $\omega=0.5$, suggesting an effective Pre-CFG scale near 2.5. As reported in \cref{tab:suppl-training}, the Pre-CFG scale of $\omega=3.0$ outperforms $\omega=5.0$, supporting the effective-scale behavior. This indicates that \textit{the effective Pre-CFG scale may not generalize to larger networks}. A similar trend has been observed in prior work, where different CFG scales were used across architectures.

\begin{figure}[th]
\centering
\includegraphics[width=0.4\linewidth]{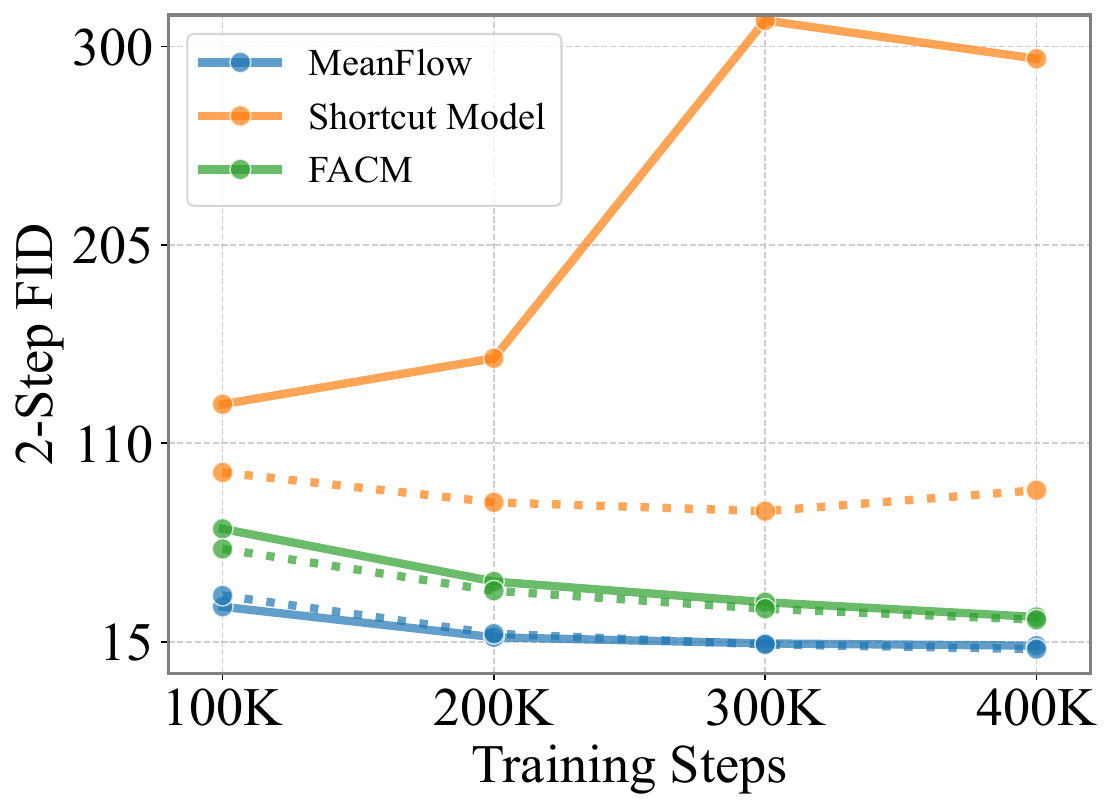}
\caption{FIDs on ImageNet. Solid lines indicate $\omega=5.0$, and dotted lines indicate the $\omega$ values in \cref{tab:imagenet-reprod}.}
\label{fig:prior_cfg50}
\end{figure}

\begin{wraptable}{r}{0.33\linewidth}
\centering
\vspace{-1.4em}
\caption{Comparison at $\omega=5.0$.}
\label{tab:imagenet-reprod-omega50}
\begin{tabular}{llr}
\toprule
Case & FID$\downarrow$ & IS$\uparrow$ \\
\midrule
MeanFlow & 13.12 & 220.9 \\
Shortcut Model & 128.7 & 6.57 \\
FACM & 26.85 & 85.05 \\
\bottomrule
\end{tabular}
\vspace{-1em}
\end{wraptable}

\textbf{Pre-CFG Scale for Prior Work.} In \cref{tab:imagenet-reprod}, we present results using the CFG scale reported as optimal in each work. For direct comparison, we train these methods with a fixed scale of $\omega=5.0$, as shown in \cref{tab:imagenet-reprod-omega50} and \cref{fig:prior_cfg50}. Under this setting, the FIDs of MeanFlow and FACM increase relative to their reported scales. For the Shortcut Model, we did not observe stable convergence at larger guidance scales in our setup, and therefore report results using the recommended scale.

This behavior is consistent with our observations for iSD-U. As discussed in \cref{proof:cfg-fmm}, both CFG formulations share a common structure: they replace the guidance velocity with its CFG counterpart, while leaving the flow matching objective unchanged. At large guidance scales, this structural discrepancy can introduce conflicts between the consistency and flow matching objectives, potentially degrading training stability.

\begin{wraptable}{r}{0.45\textwidth}
\centering
\vspace{-1.4em}
\caption{FIDs across batch sizes.}
\label{tab:fid-batch}
\begin{tabular}{lrrr}
\toprule
Case & \multicolumn{1}{c}{32} & \multicolumn{1}{c}{256} & \multicolumn{1}{c}{1024} \\
\midrule
Baseline & 137.35 & 120.49 & 115.80 \\
MeanFlow & 31.63 & 11.48 & 8.84 \\
iSD-T & 31.47 & 14.62 & 13.10 \\
\midrule
iSD-T (DiT-B/2$^\dagger$) & \multicolumn{1}{c}{-} & 8.76 & 8.79 \\
\bottomrule
\end{tabular}
\vspace{-2em}
\end{wraptable}

\textbf{Batch Size.} To extend our batch size experiments from the toy setting in \cref{fig:ct-batchsize} to the ImageNet scale, we vary the batch size across the listed methods in \cref{tab:fid-batch}.

For the baseline and MeanFlow, larger batch sizes improve FIDs. This is consistent with our analysis: increasing the batch size reduces the residual in \cref{prop:instability}.

For iSD-T, performance appears less sensitive to batch size. One possible explanation is that the guidance velocity is given by a network approximation of the marginal velocity, which could reduce reliance on accurate batch statistics. In this case, performance may depend more on the quality of the network estimation. This reduced sensitivity is also observed in the last row, where DiT-B/2 yields similar FIDs across batch sizes. We acknowledge that these experiments are limited to a single run, and we leave validation across multiple seeds for future work.

We note that the default batch size of 256 in our ablation setting follows the configurations of MeanFlow~\citep{meanflow} and Shortcut Model~\citep{shortcutmodel}.

\subsection{Comparison with Other Work}
\label{appendix:comparison}

\begin{figure}[th]
\centering
\includegraphics[width=0.45\linewidth]{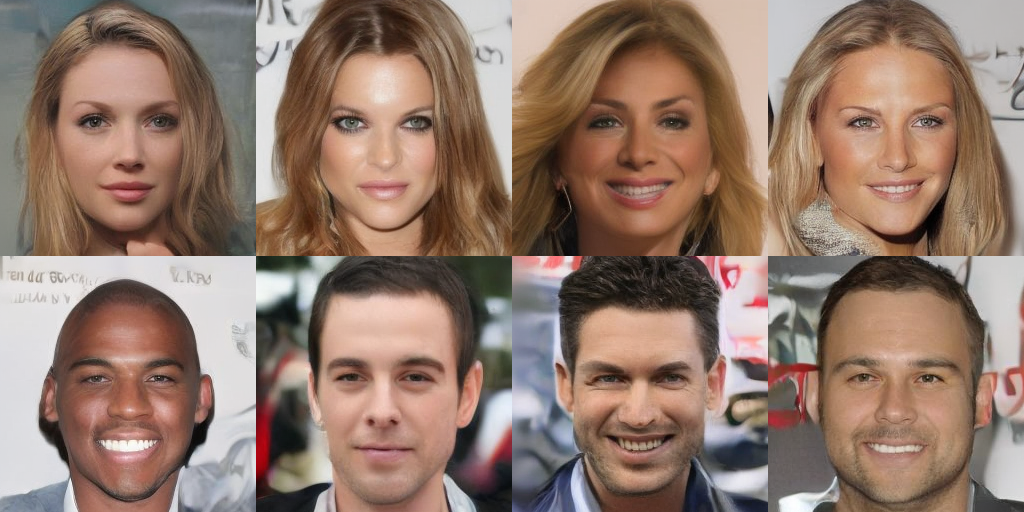}
\caption{4-step samples from iSD on CelebA-HQ $256\!\times\!256$.}
\label{expr:celebahq}
\end{figure}

\textbf{CelebA-HQ.} We conduct additional comparisons on CelebA-HQ to validate our method in an unconditional setting. For fair comparison, all methods in \cref{tab:quant-celebahq} are trained with the DiT-B/2 architecture. As shown in \cref{tab:quant-celebahq}, iSD achieves competitive results on both few-step (4-step) and multistep (128-step) generation. Since we train for 200K steps in this case, iSD provides an efficient training configuration, while the Shortcut Model uses 800K steps.

\textbf{Shortcut Model.} To compare the reproducibility with the Shortcut Model, we train the network under the same ablation settings. We utilize an EMA network as a teacher with its CFG approach at $\omega=1.5$. Note that its guidance method shares an idea with iSD-U (\cref{proof:cfg-fmm}). Under this setup, iSD-U attains stronger quantitative results, as shown in \cref{tab:suppl-training}.

We hypothesize that the performance gap between the Shortcut Model and iSD-U may arise from the step-size discretization in the Shortcut Model, which introduces an asymptotic gap relative to continuous-time formulations. Given the behaviors seen in iSD-U under different $\omega$ values, it is plausible that the Shortcut Model may experience increased training sensitivity as $\omega$ becomes larger. As iSD-T shows the best results among the evaluated variants and preserves objective alignment, applying its Pre-CFG scheme to shortcut approaches may be a promising future direction.

\begin{figure}[th]
\centering
\subfloat[Gradient norms]{
\includegraphics[width=0.45\linewidth]{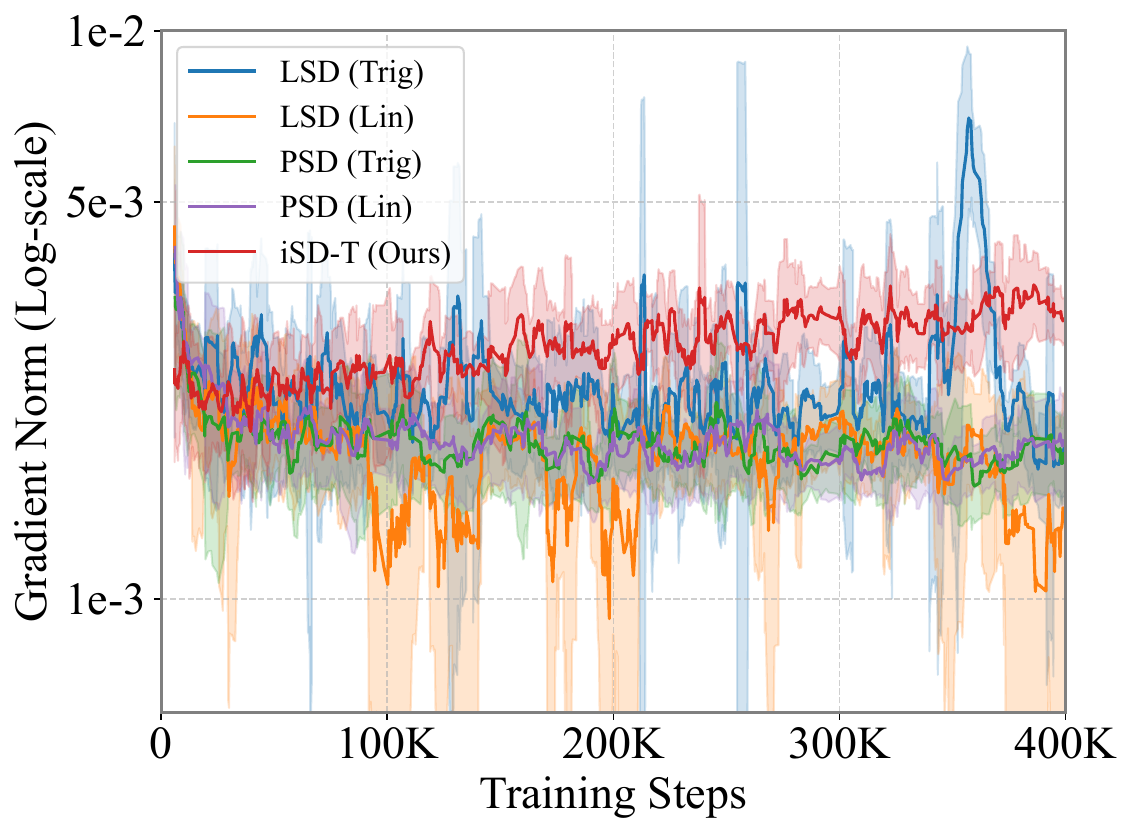}
}
\subfloat[FIDs]{
\includegraphics[width=0.45\linewidth]{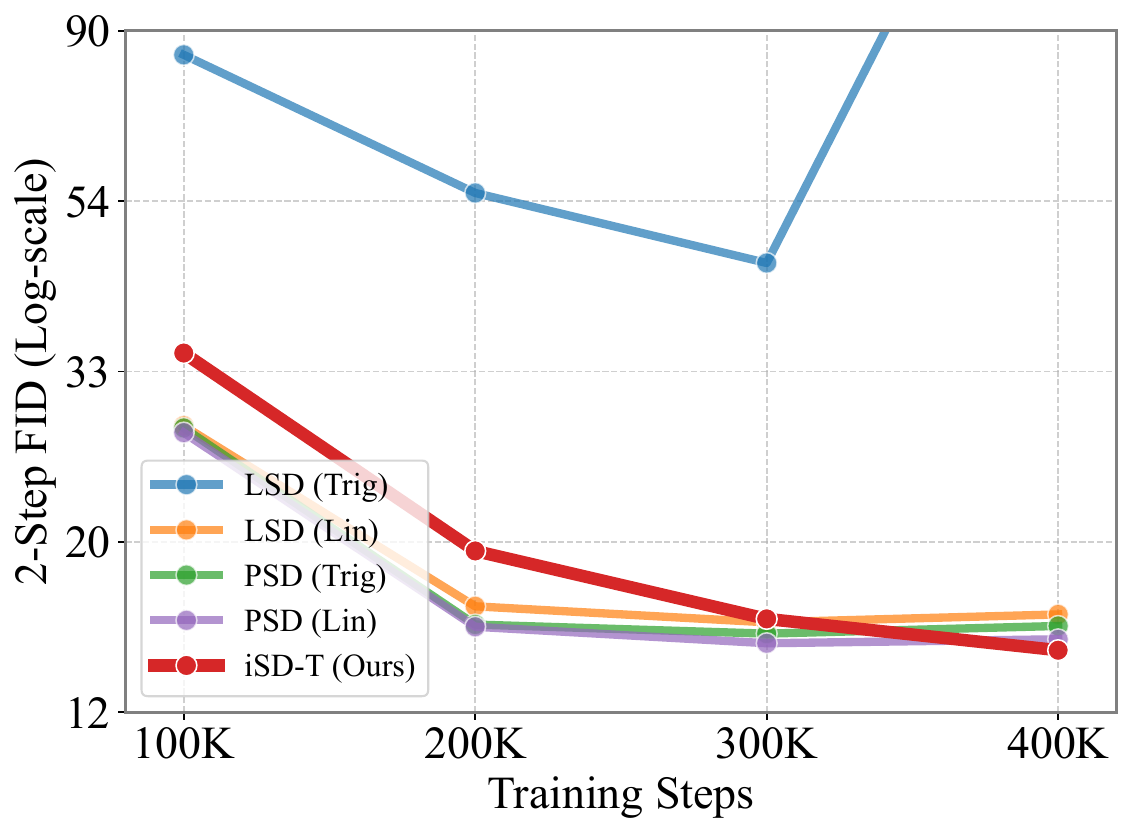}
}
\caption{Training curves of self-distillation methods. The experimental settings follow the ImageNet ablation study. All listed methods adopt Pre-CFG with $\omega=5.0$.}
\label{fig:lsd-psd}
\end{figure}

\textbf{LSD \& PSD.} To compare the stability of other self-distillation methods in \citet{boffi2025}, we conduct experiments with Lagrangian self-distillation (LSD) and progressive self-distillation (PSD). For each method, we use the corresponding objective:
\begin{align*}
    \mathcal L_\mathrm{LSD} &= \lambda_1\mathcal L_\mathrm{CFM}^\omega + \lambda_2\mathbb E_{x,z,t,s}\left[\left\|\partial_s f_\theta(x_t; t, s) - \operatorname{sg}\left[F_\theta(f_\theta(x_t; t, s); s, s)\right] \right\|^2_2\right], \\
    \mathcal L_\mathrm{PSD} &= \lambda_1\mathcal L_\mathrm{CFM}^\omega + \lambda_2\mathbb E_{x,z,t,s,u}\left[\left\|F_\theta(x_t; t, s) - \operatorname{sg}\left[F_\mathrm{PSD}\right]\right\|^2_2\right] \\
    &\mathrm{where}\ F_\mathrm{PSD} = A_{t,s}^{-1}[(A'_{t,s} - A_{u,s}'A'_{t,u})x_t + A_{u,s}'A_{t,u}F_\theta(x_t; t, u)+ A_{u,s}F_\theta(f_\theta(x_t; t, u); u, s)]
\end{align*}
where we assume $\nu=1$, which is satisfied by linear and trigonometric interpolation. This coincides with the linear realization in \citet{boffi2025}, when $u = \gamma t + (1 - \gamma)s$ for $\gamma\sim \mathcal U[0, 1]$. In \cref{fig:lsd-psd}, we follow the settings of our ablation study with a Pre-CFG scale of 5.0, and the only difference is the objective.

As shown in \cref{fig:lsd-psd}, we observe that LSD shows more variable gradient norms than PSD and iSD-T, and its trigonometric variant diverges after a gradient spike around 350K training steps. In this Pre-CFG setting, iSD-T shows a lower FID than the compared variants.

\textbf{FACM.} As a concurrent work, FACM~\citep{facm} suggests \textit{flow-anchoring}, which jointly trains the network with the flow matching objective to provide an anchored training signal. This anchored signal is introduced to prevent degenerate model collapse and enable training from scratch. To compare with our method, we train FACM under our ablation settings. As shown in \cref{tab:suppl-training}, FACM also reduces the variance of the resulting metrics compared to the baseline. It can be viewed as a weak version of the time-condition relaxation, which relaxes the condition to $s\in\{0, t\}$ rather than $s = 0$, aligning with our hypothesis.

\textbf{improved MeanFlow.} In line with recent work, improved MeanFlow~\citep[iMF;][]{imf} reorganizes the MeanFlow identity into a $v$-loss formulation and highlights that conditional velocity leakage can arise when formalizing instantaneous velocities from average velocity ($u$-prediction). They note that this leakage contributes to training instability, and that replacing conditional velocity guidance with the model’s own marginal velocity prediction ($v_\psi$) can improve stability. Under linear interpolation, the gradient of the resulting objective can be aligned with our formulation at the expectation level:
\begin{align*}
\nabla_\theta\mathcal L_\mathrm{iMF} &\propto \mathbb E\left[\nabla_\theta F_\theta(x_t; t, s)\cdot \left(F_\theta(x_t; t, s) - v_t(x_t|x) + (t - s)\cdot\left(\partial_t F_\theta + v_\psi(x_t; t)\cdot\nabla_x F_\theta\right)\right)\right] \\
&\approx\mathbb E\left[\nabla_\theta F_\theta(x_t; t, s)\cdot(F_\theta(x_t; t, s) - v^*_t(x_t) + (t - s)\cdot (\partial_t F_\theta + v^*_t(x_t)\cdot\nabla_xF_\theta))\right] \\
&\approx \mathbb E\left[\nabla_\theta F_\theta(x_t; t, s)\cdot (F_\theta(x_t; t, s) - F_{\theta^-}(x_t; t, t) + (t - s)\cdot (\partial_tF_\theta + F_{\theta^-}(x_t; t, t)\cdot \nabla_x F_\theta))\right] \\
&\propto \nabla_\theta\mathcal L_\mathrm{SD\text{-}R},
\end{align*}
where $v_\psi(x_t; t)\approx v^*_t(x_t)$ is a velocity prediction head of the iMF network. We note that the training target of iMF is based on the conditional velocity, while the training target of our $\mathcal L_\mathrm{SD\text{-}R}$ is based on the marginal velocity. This makes the relation hold only at the expectation level. While the motivation and analytical perspective of iMF differ from ours, both approaches arrive at marginal velocity guidance through independent reasoning.

\subsection{Policy Generation}
\label{appendix:policy-generation}

To assess the applicability of iSD to diffusion-based policy learning, we train MeanFlow and iSD to imitate proficient human controls on two tasks, Push-T and Transport, using transformer- and state-based policies in simulation environments.

Unlike diffusion policy, iSD and MeanFlow require additional conditioning on $s$. Following prior work, which concatenates $t$ to the observation context, we extend it by additionally appending $s$. Other settings follow prior work~\citep{chi2023diffusionpolicy}.

\begin{figure}[th]
\centering
\subfloat[Push-T]{%
\includegraphics[width=0.31\linewidth]{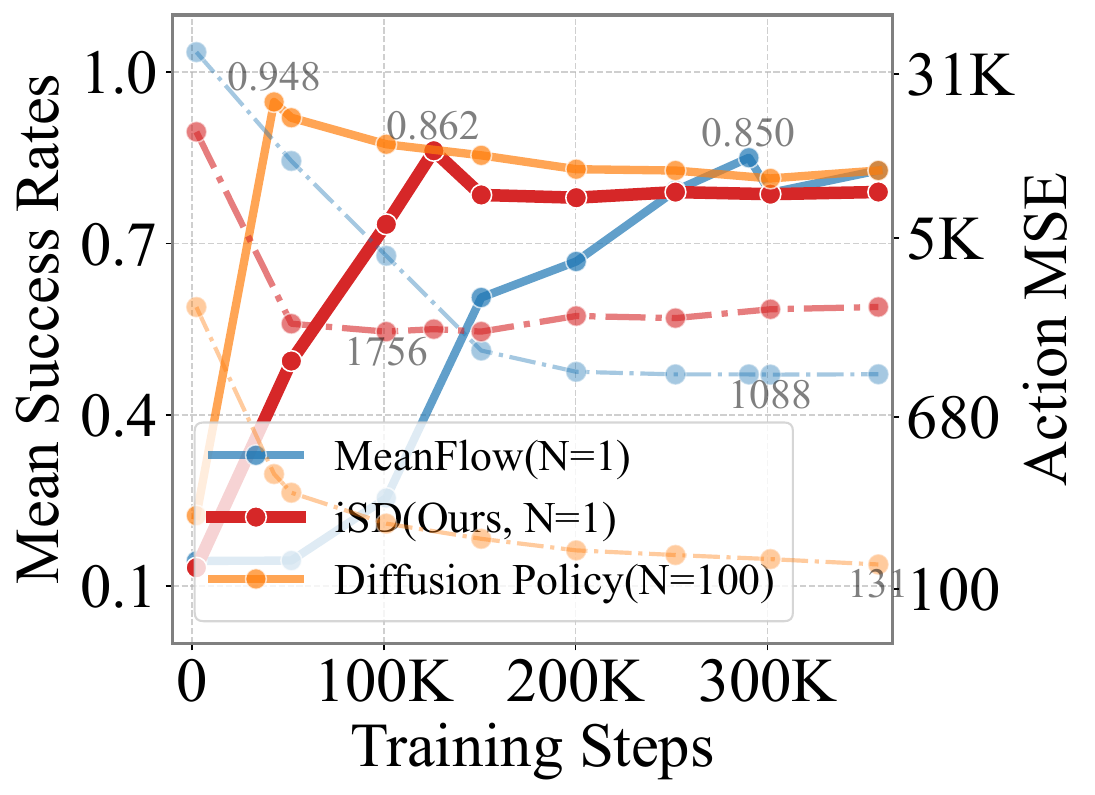}
}
\subfloat[Transport]{%
\includegraphics[width=0.31\linewidth]{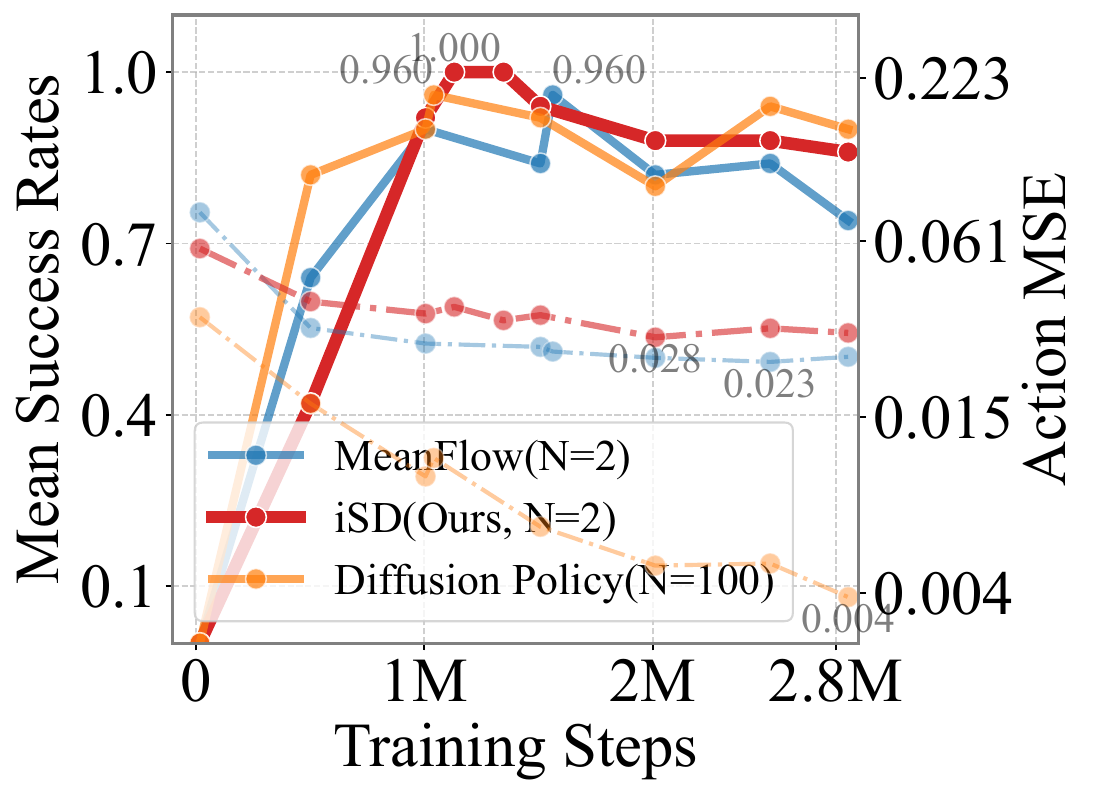}
}
\subfloat[Simulation Results]{%
\includegraphics[width=0.31\linewidth]{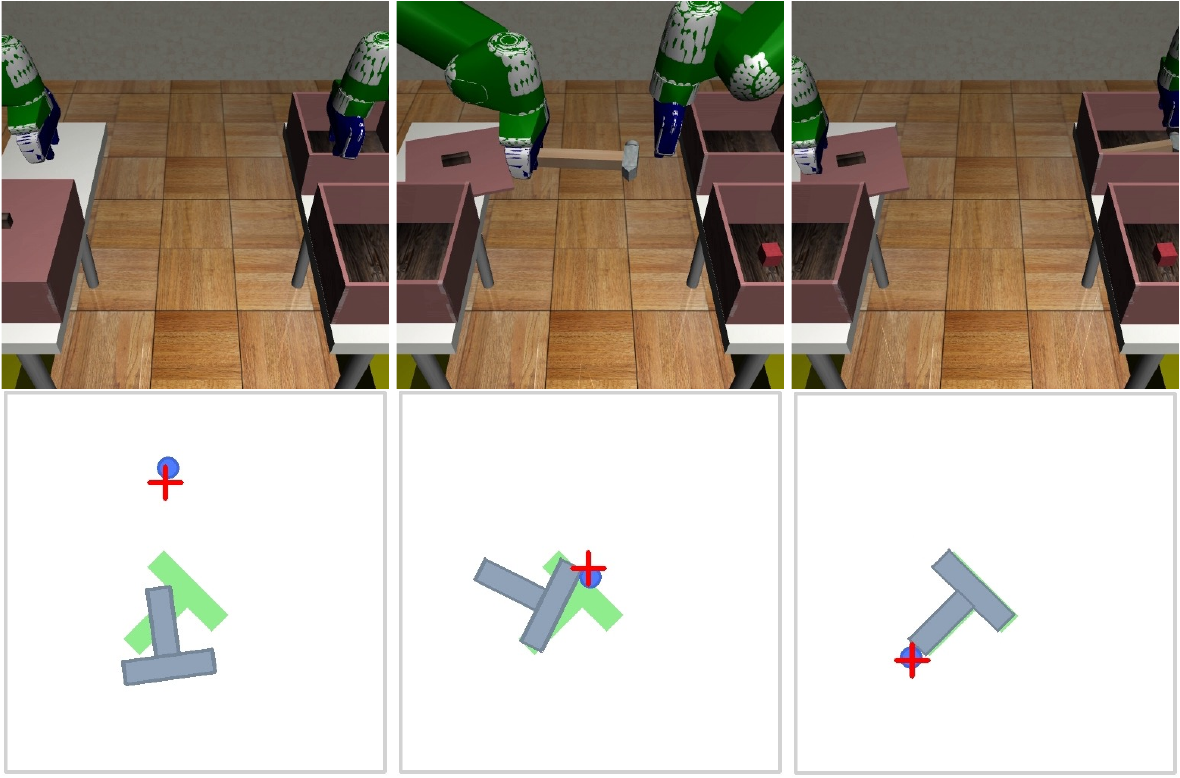}
}
\caption{Results of policy learning. The left two figures show average success rates (solid lines) and action mean-squared errors (dash-dot lines) over training steps. Numeric entries indicate the optimal values across training steps. The rightmost figure shows the simulation results of iSD with NFE=2.}
\label{fig:policy-generation}
\end{figure}

As shown in \cref{tab:diffusion-policy}, iSD with linear interpolation achieves results comparable to those of other few-step methods, and performs particularly well on the Transport task. With 2-step sampling, iSD achieves performance comparable to that of the diffusion policy, which uses 100 sampling steps. Even with 2-step sampling, MeanFlow achieves an average success rate of 0.96, which is lower than iSD's. This suggests that iSD can be a reasonable candidate for policy generation.

In \cref{fig:policy-generation}, we observe that iSD consistently exhibits higher action error than the other methods. However, it achieves success rates comparable to or better than those of other approaches. Even though the diffusion policy achieves a low action error of 0.004 on the Transport task, its peak success rate occurs earlier in training. In this case, the action error does not fully reflect the success rate, and iSD can achieve higher success rates despite yielding higher action errors than other methods.

We note that trigonometric interpolation underperforms linear interpolation under iSD settings, achieving an average success rate of 0.605 on Push-T with NFE=2. This is consistent with the observation that interpolation exhibits varying performance across settings, and that linear interpolation appears to be a more suitable choice for policy generation in this setting. 

\subsection{Loss Landscape}
\label{appendix:loss-landscape}

\begin{figure}[th]
\centering
\subfloat[Consistency Model]{%
\includegraphics[width=0.24\linewidth]{figures/Ba_landscape.pdf}
}
\subfloat[Time Relaxation w/o $\mathcal L_\mathrm{CFM}$]{%
\includegraphics[width=0.24\linewidth]{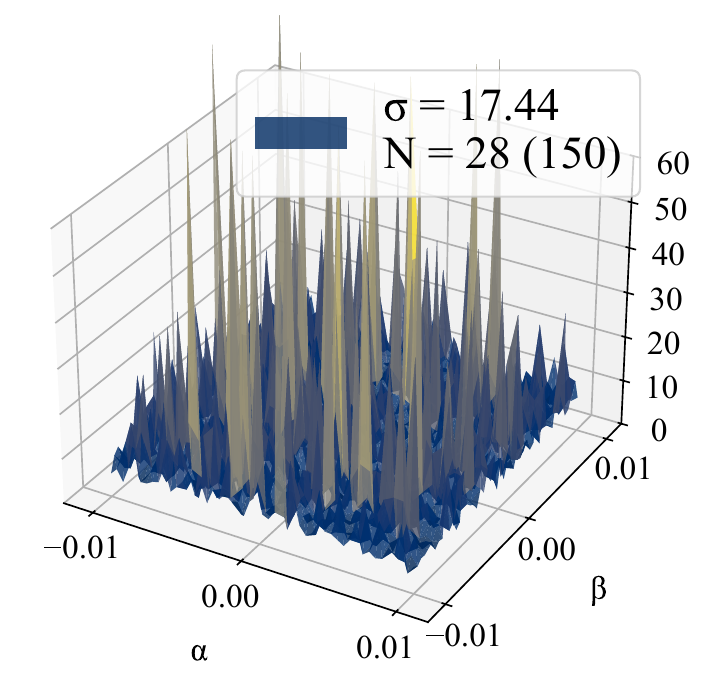}
}
\subfloat[iSD w/o $\mathcal L_\mathrm{CFM}$]{%
\includegraphics[width=0.24\linewidth]{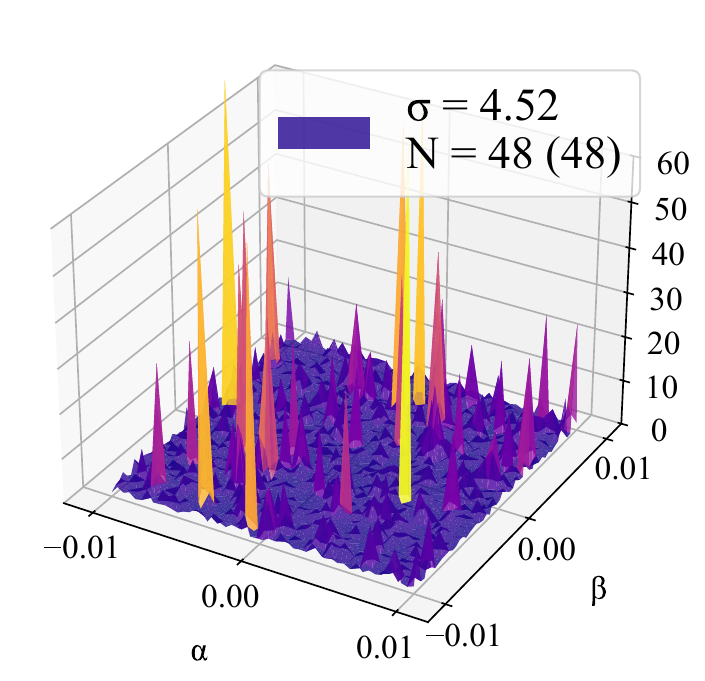}
\label{fig:isd-wo-cfm}
}

\subfloat[Flow Matching]{%
\includegraphics[width=0.24\linewidth]{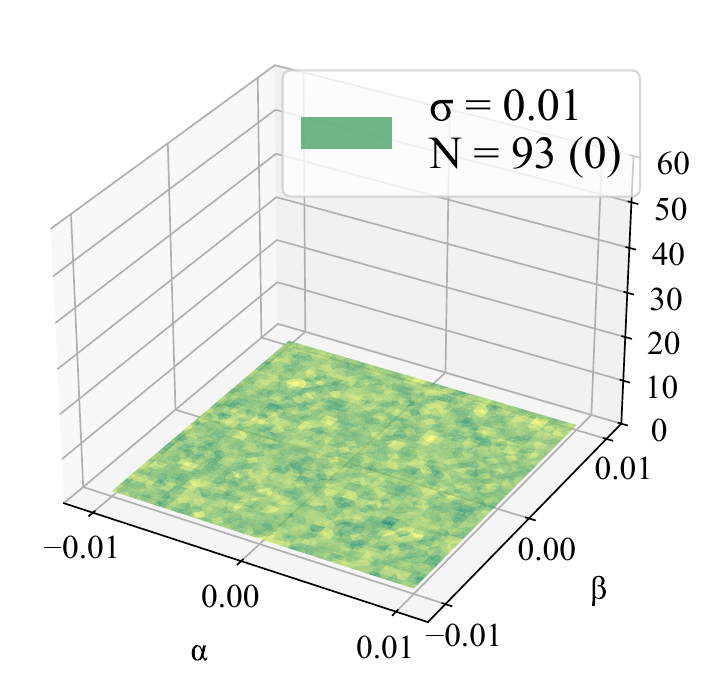}
}
\subfloat[Time Relaxation]{%
\includegraphics[width=0.24\linewidth]{figures/Be_landscape.pdf}
}
\subfloat[Self-Distillation]{%
\includegraphics[width=0.24\linewidth]{figures/Bg_landscape.pdf}
}
\subfloat[iSD (Ours)]{%
\includegraphics[width=0.24\linewidth]{figures/Bf_landscape.pdf}
}
\caption{Loss landscapes of four methods at 0K training steps. $\alpha$ and $\beta$ denote the top-2 eigenvectors of the Hessian on ImageNet-1K with DiT-B/4. $\sigma$ denotes the standard deviations of the landscape fields, and $N$ the number of samples outside each method's own empirical 95\% range. Values in parentheses report the number of samples exceeding the 95\% bound defined by iSD, as a common reference.}
\label{fig:landscape}
\end{figure}

\begin{figure}[th]
\centering
\subfloat[Consistency Model]{%
\includegraphics[width=0.24\linewidth]{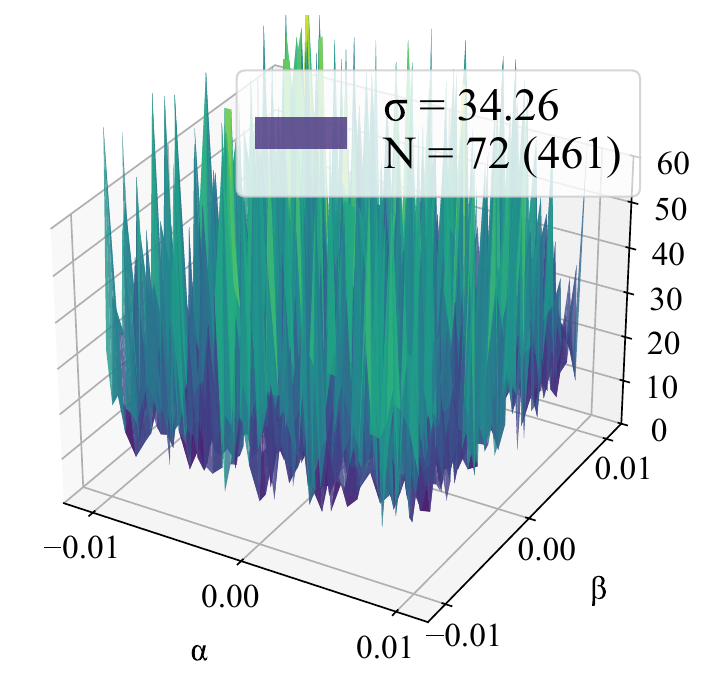}
}
\subfloat[Time Relaxation w/o $\mathcal L_\mathrm{CFM}$]{%
\includegraphics[width=0.24\linewidth]{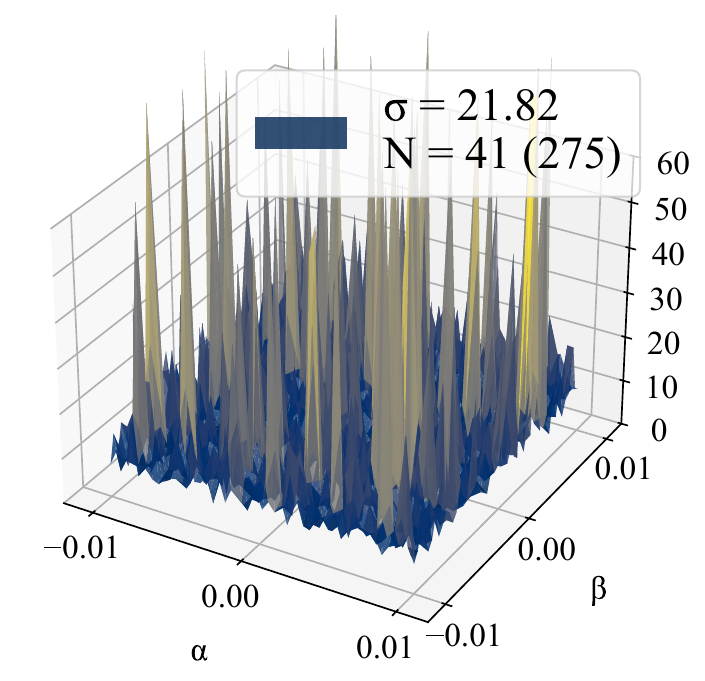}
}
\subfloat[iSD w/o $\mathcal L_\mathrm{CFM}$]{%
\includegraphics[width=0.24\linewidth]{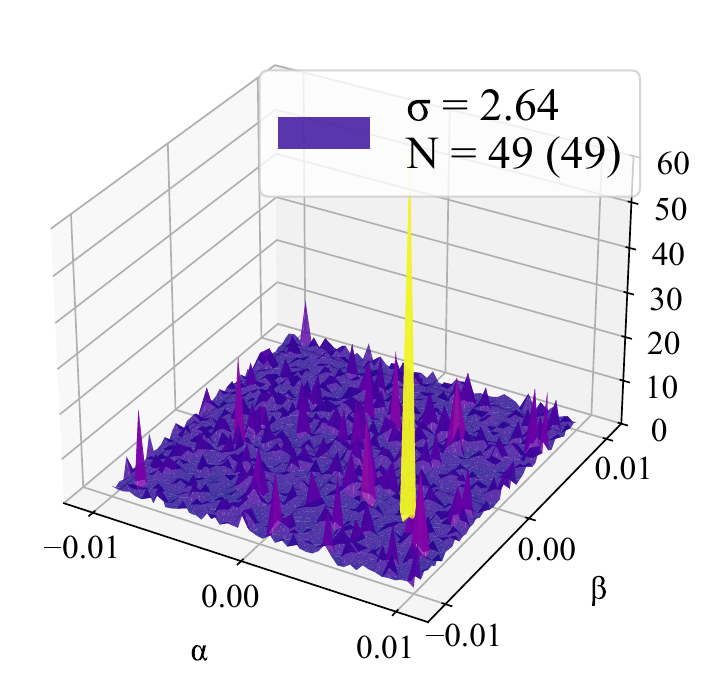}
}

\subfloat[Flow Matching]{%
\includegraphics[width=0.24\linewidth]{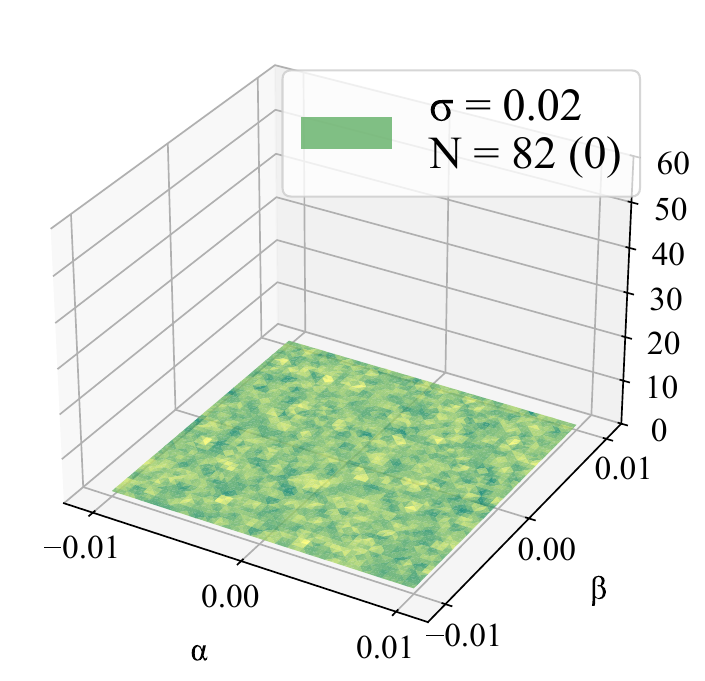}
}
\subfloat[Time Relaxation]{%
\includegraphics[width=0.24\linewidth]{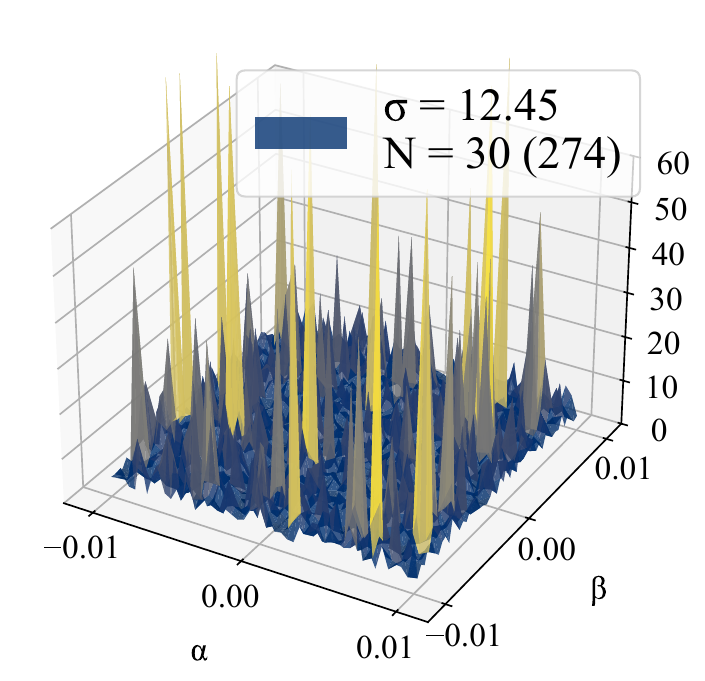}
}
\subfloat[Self-Distillation]{%
\includegraphics[width=0.24\linewidth]{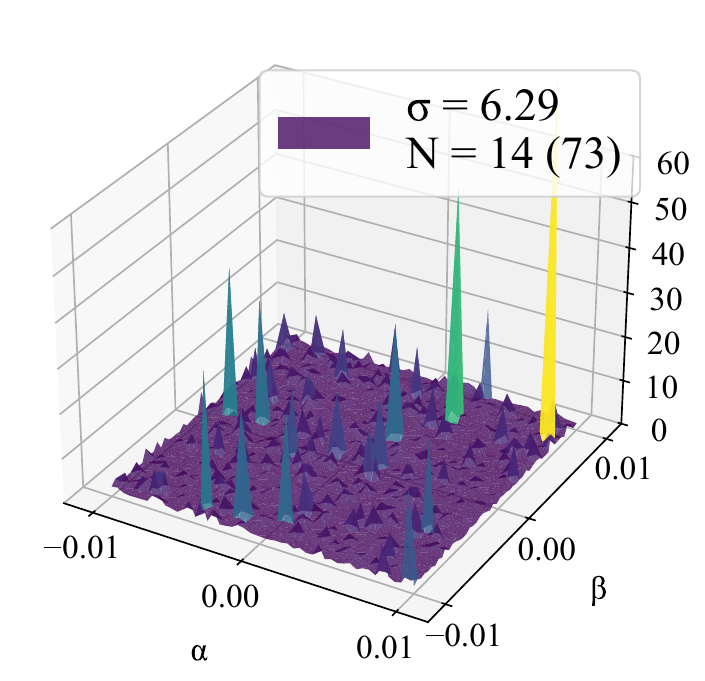}
}
\subfloat[iSD (Ours)]{%
\includegraphics[width=0.24\linewidth]{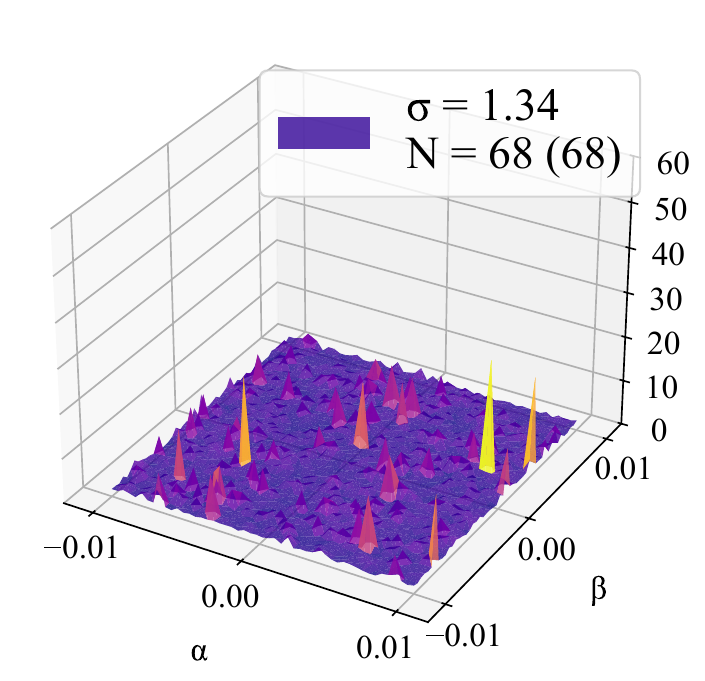}
}
\caption{Loss landscapes at 200K training steps. Other settings are identical to those in \cref{fig:landscape}.}
\label{fig:landscape-200K}
\end{figure}

We conduct loss landscape analysis using PyHessian~\citep{pyhessian} on ImageNet-trained DiT-B/4. For each objective, we compute the top-2 eigenvectors of its Hessian to explore the landscape along the maximum curvature. To probe different training stages, we compute the Hessian at 0K and 200K training steps and visualize the landscapes in \cref{fig:landscape} and \cref{fig:landscape-200K}, respectively. We measure the loss variance and count spikes by identifying outliers beyond an empirical 95\% range, providing empirical support for the linearization cost hypothesis.

We examine each objective without joint training with flow matching in the upper rows. As shown in \cref{fig:landscape} and \cref{fig:landscape-200K}, time-condition relaxation reduces loss spikes and variance, supporting the linearization cost hypothesis. In \cref{fig:isd-wo-cfm}, iSD shows even lower variance. These observations suggest that, independent of joint training with flow matching, both reduced linearization cost and marginal velocity guidance are associated with a more stable optimization landscape. In the lower rows with joint training, time-condition relaxation, self-distillation, and iSD show progressively lower variance and fewer spikes under the common reference. Across both checkpoints, these observations remain consistent with an association between time-condition relaxation, self-distillation, and a more stable optimization landscape.

\vspace*{\fill}
\begin{table}[ht]
\caption{Quantitative results (2-NFE). The numeric entries in the header denote training steps. Reported results are averaged over three random seeds unless otherwise specified; entries marked with an underline are averaged over five seeds. \textit{Std.} denotes the standard deviation at 400K steps. Hyphens in \textit{Std.} indicate cases evaluated with a single run, where standard deviation was not computed. \textit{Halted} denotes cases where training diverged and was stopped earlier than the baseline.}
\label{tab:suppl-training}
\subfloat[Fr\'echet Inception Distance]{%
\resizebox{1.0\linewidth}{!}{%
\begin{tabular}{llllcccccc}
\toprule
Loss & Interp. & JVP & Arch. & Pre-CFG $\omega$ & 100K & 200K & 300K & 400K & Std. \\
\midrule
Baseline & Linear & Exact & DiT-B/4 & - & 137.84 & 127.23 & 122.76 & 120.49 & \underline{3.71} \\ 
Joint training & Linear & Exact & DiT-B/4 & - & 132.84 & 105.90 & 94.94 & 96.28 & \underline{15.86} \\ 
\midrule
$\mathcal L_\text{CT}$ & Linear & Exact & DiT-B/4 & - & 194.95 & 142.92 & 73.32 & 69.18 & \underline{2.84} \\ 
$\mathcal L_\text{CT}$ & Trig & Exact & DiT-B/4 & - & 131.12 & 110.28 & 102.79 & 97.84 & 3.40 \\ 
$\mathcal L_\text{iSD}$ & Linear & Exact & DiT-B/4 & - & 155.30 & 128.15 & 119.98 & 114.56 & \underline{2.80} \\ 
$\mathcal L_\text{iSD}$ & Trig & Exact & DiT-B/4 & - & 147.01 & 125.83 & 117.19 & 110.18 & 4.04 \\ 
\midrule
$\mathcal L_\text{CT}$ & Linear & Approx. & DiT-B/4 & - & 94.53 & 74.23 & 67.64 & 63.02 & \underline{0.42} \\ 
$\mathcal L_\text{CT}$ & Trig & Approx. & DiT-B/4 & - & 223.10 & 227.90 & 228.70 & Halted & - \\ 
$\mathcal L_\text{iSD}$ & Linear & Approx. & DiT-B/4 & - & 146.37 & 122.32 & 113.79 & 108.70 & \underline{2.83} \\ 
$\mathcal L_\text{iSD}$ & Trig & Approx. & DiT-B/4 & - & 139.95 & 123.56 & 113.88 & 108.33 & - \\ 
\midrule
$\mathcal L_\text{iSD-U}$ & Linear & Approx. & DiT-B/4 & 1.5 & 124.94 & 93.20 & 82.15 & 75.53 & 1.48 \\ 
$\mathcal L_\text{iSD-U}$ & Trig & Approx. & DiT-B/4 & 1.5 & 119.86 & 86.21 & 74.48 & 67.66 & 0.83 \\ 
\midrule
$\mathcal L_\text{iSD-C}$ & Linear & Approx. & DiT-B/4 & 1.5 & 166.93 & 128.24 & 110.30 & 100.47 & - \\ 
$\mathcal L_\text{iSD-C}$ & Linear & Approx. & DiT-B/4 & 3.0 & 200.17 & 120.56 & 98.73 & 88.93 & - \\ 
$\mathcal L_\text{iSD-C}$ & Linear & Approx. & DiT-B/4 & 7.0 & 270.50 & Halted & - \\ 
$\mathcal L_\text{iSD-C}$ & Trig & Approx. & DiT-B/4 & 1.5 & 164.34 & 133.90 & 119.94 & 111.56 & - \\ 
$\mathcal L_\text{iSD-C}$ & Trig & Approx. & DiT-B/4 & 3.0 & 147.15 & 105.09 & 88.08 & 79.64 & - \\ 
$\mathcal L_\text{iSD-C}$ & Trig & Approx. & DiT-B/4 & 7.0 & 192.14 & 127.65 & 111.93 & 110.49 & - \\ 
\midrule
$\mathcal L_\text{iSD-T}$ & Linear & Approx. & DiT-B/4 & 1.5 & 115.18 & 83.02 & 70.88 & 63.81 & - \\ 
$\mathcal L_\text{iSD-T}$ & Linear & Approx. & DiT-B/4 & 3.0 & 61.77 & 39.68 & 31.80 & 27.62 & 0.40 \\ 
$\mathcal L_\text{iSD-T}$ & Linear & Approx. & DiT-B/4 & 5.0 & 51.77 & 35.20 & 29.81 & 26.91 & \underline{0.50} \\ 
$\mathcal L_\text{iSD-T}$ & Linear & Approx. & DiT-B/4 & 7.0 & 51.88 & 36.85 & 31.46 & 28.61 & - \\ 
$\mathcal L_\text{iSD-T}$ & Linear & Approx. & DiT-B/4 & 10.0 & 54.90 & 41.77 & 37.07 & 33.86 & - \\ 
$\mathcal L_\text{iSD-T}$ & Trig & Approx. & DiT-B/4 & 1.5 & 106.54 & 75.85 & 65.50 & 61.19 & - \\ 
$\mathcal L_\text{iSD-T}$ & Trig & Approx. & DiT-B/4 & 3.0 & 46.01 & 24.35 & 18.42 & 15.56 & 0.12 \\ 
$\mathcal L_\text{iSD-T}$ & Trig (Ours) & Approx. & DiT-B/4 & 5.0 & 34.97 & 19.56 & 16.02 & 14.62 & \underline{0.06} \\ 
$\mathcal L_\text{iSD-T}$ & Trig & Approx. & DiT-B/4 & 7.0 & 36.17 & 21.83 & 18.72 & 17.17 & - \\ 
$\mathcal L_\text{iSD-T}$ & Trig & Approx. & DiT-B/4 & 10.0 & 40.54 & 25.69 & 21.75 & 20.18 & - \\ 
\midrule
MeanFlow & Linear & Exact & DiT-B/4 & 3.0 & 37.11 & 18.84 & 13.73 & 11.48 & \underline{1.68} \\ 
MeanFlow & Trig & Exact & DiT-B/4 & 3.0 & 48.36 & 24.73 & 18.19 & 15.08 & 3.11 \\ 
Shortcut Model & Linear & Exact & DiT-B/4 & 1.5 & 96.02 & 81.74 & 77.51 & 87.52 & \underline{24.34} \\ 
FACM & Linear & Exact & DiT-B/4 & 1.75 & 59.45 & 39.41 & 30.79 & 25.52 & \underline{1.73} \\ 
\midrule
$\mathcal L_\text{iSD-T}$ & Trig & Approx. & DiT-B/2 & 3.0 & 36.40 & 19.20 & 13.83 & 11.92 & - \\ 
$\mathcal L_\text{iSD-T}$ & Trig & Approx. & DiT-B/2 & 5.0 & 31.00 & 18.96 & 16.75 & 15.93 & - \\ 
$\mathcal L_\text{iSD-T}$ & Trig & Approx. & DiT-L/2 & 3.0 & 26.02 & 14.86 & 12.11 & 11.17 & - \\ 
$\mathcal L_\text{iSD-T}$ & Trig & Approx. & DiT-XL/2 & 3.0 & 24.69 & 14.33 & 11.82 & 11.08 & - \\ 
$\mathcal L_\text{iSD-T}$ & Trig & Approx. & DiT-XL/2 & 5.0 & 28.30 & 18.89 & 17.33 & 17.03 & - \\ 
\midrule
$\mathcal L_\text{iSD-T}$ & Linear & Approx. & DiT-XL/2$^\dagger$ & 6.0 (Eval) & 14.53 & 5.82 & 4.30 & 3.79 & - \\ 
$\mathcal L_\text{iSD-T}$ & Linear & Approx. & DiT-XL/1$^\dagger$ & 6.0 (Eval) & 7.15 & 3.01 & 2.64 & 2.52 & - \\
\bottomrule
\\ \\

\toprule
Loss & Interp. & JVP & Arch. & Pre-CFG $\omega$ & 500K & 600K & 700K & 800K \\
\midrule
$\mathcal L_\text{iSD-T}$ & Trig & Approx. & DiT-XL/2 & 3.0 & 10.66 & 10.48 & 10.29 & 10.30 \\ 
$\mathcal L_\text{iSD-T}$ & Trig & Approx. & DiT-XL/2 & 5.0 & 16.87 & 16.85 & 16.96 \\ 
$\mathcal L_\text{iSD-T}$ & Linear & Approx. & DiT-XL/2$^\dagger$ & 6.0 (Eval) & 3.47 & 3.28 & 3.09 & 2.98 \\ 
$\mathcal L_\text{iSD-T}$ & Linear & Approx. & DiT-XL/1$^\dagger$ & 6.0 (Eval) & 2.40 & 2.30 \\
\bottomrule
\end{tabular}
}}
\end{table}
\vspace*{\fill}

\vspace*{\fill}
\begin{table}[ht]
\ContinuedFloat
\captionsetup{position=top}
\subfloat[Inception Score]{%
\resizebox{1.0\linewidth}{!}{%
\begin{tabular}{llllcccccc}
\toprule
Loss & Interp. & JVP & Arch. & Pre-CFG $\omega$ & 100K & 200K & 300K & 400K & Std. \\
\midrule
Baseline & Linear & Exact & DiT-B/4 & - & 7.90 & 8.88 & 9.40 & 9.69 & \underline{0.47} \\ 
Joint training & Linear & Exact & DiT-B/4 & - & 10.95 & 13.01 & 14.32 & 14.11 & \underline{4.15} \\ 
\midrule
$\mathcal L_\text{CT}$ & Linear & Exact & DiT-B/4 & - & 8.76 & 11.34 & 19.78 & 21.45 & \underline{1.21} \\ 
$\mathcal L_\text{CT}$ & Trig & Exact & DiT-B/4 & - & 11.77 & 15.04 & 16.79 & 18.11 & 0.98 \\ 
$\mathcal L_\text{iSD}$ & Linear & Exact & DiT-B/4 & - & 7.21 & 9.60 & 10.43 & 11.04 & \underline{0.41} \\ 
$\mathcal L_\text{iSD}$ & Trig & Exact & DiT-B/4 & - & 9.37 & 12.02 & 13.53 & 14.75 & 0.70 \\ 
\midrule
$\mathcal L_\text{CT}$ & Linear & Approx. & DiT-B/4 & - & 14.86 & 19.50 & 21.92 & 24.08 & \underline{0.29} \\ 
$\mathcal L_\text{CT}$ & Trig & Approx. & DiT-B/4 & - & 3.76 & 3.85 & 3.89 & Halted & - \\ 
$\mathcal L_\text{iSD}$ & Linear & Approx. & DiT-B/4 & - & 8.18 & 10.48 & 11.21 & 11.80 & \underline{0.36} \\ 
$\mathcal L_\text{iSD}$ & Trig & Approx. & DiT-B/4 & - & 10.02 & 11.80 & 13.68 & 14.93 & - \\ 
\midrule
$\mathcal L_\text{iSD-U}$ & Linear & Approx. & DiT-B/4 & 1.5 & 10.60 & 15.38 & 17.70 & 19.63 & 0.45 \\ 
$\mathcal L_\text{iSD-U}$ & Trig & Approx. & DiT-B/4 & 1.5 & 12.72 & 20.71 & 25.57 & 29.26 & 0.52 \\ 
\midrule
$\mathcal L_\text{iSD-C}$ & Linear & Approx. & DiT-B/4 & 1.5 & 6.45 & 9.33 & 11.18 & 12.54 & - \\ 
$\mathcal L_\text{iSD-C}$ & Linear & Approx. & DiT-B/4 & 3.0 & 4.88 & 10.16 & 12.55 & 13.98 & - \\ 
$\mathcal L_\text{iSD-C}$ & Linear & Approx. & DiT-B/4 & 7.0 & 1.83 & Halted & - \\ 
$\mathcal L_\text{iSD-C}$ & Trig & Approx. & DiT-B/4 & 1.5 & 8.04 & 10.95 & 12.58 & 13.88 & - \\ 
$\mathcal L_\text{iSD-C}$ & Trig & Approx. & DiT-B/4 & 3.0 & 9.14 & 14.30 & 17.17 & 18.84 & - \\ 
$\mathcal L_\text{iSD-C}$ & Trig & Approx. & DiT-B/4 & 7.0 & 4.69 & 10.79 & 12.42 & 13.12 & - \\ 
\midrule
$\mathcal L_\text{iSD-T}$ & Linear & Approx. & DiT-B/4 & 1.5 & 11.87 & 17.14 & 20.84 & 23.85 & - \\ 
$\mathcal L_\text{iSD-T}$ & Linear & Approx. & DiT-B/4 & 3.0 & 24.94 & 41.33 & 53.74 & 63.90 & 0.99 \\ 
$\mathcal L_\text{iSD-T}$ & Linear & Approx. & DiT-B/4 & 5.0 & 30.43 & 50.69 & 63.82 & 74.70 & \underline{2.43} \\ 
$\mathcal L_\text{iSD-T}$ & Linear & Approx. & DiT-B/4 & 7.0 & 31.42 & 50.69 & 63.81 & 75.09 & - \\ 
$\mathcal L_\text{iSD-T}$ & Linear & Approx. & DiT-B/4 & 10.0 & 28.87 & 42.85 & 53.79 & 63.18 & - \\ 
$\mathcal L_\text{iSD-T}$ & Trig & Approx. & DiT-B/4 & 1.5 & 15.68 & 25.63 & 31.76 & 35.41 & - \\ 
$\mathcal L_\text{iSD-T}$ & Trig & Approx. & DiT-B/4 & 3.0 & 39.45 & 84.45 & 114.30 & 135.60 & 0.19 \\ 
$\mathcal L_\text{iSD-T}$ & Trig (Ours) & Approx. & DiT-B/4 & 5.0 & 53.20 & 117.46 & 157.83 & 184.53 & \underline{0.77} \\ 
$\mathcal L_\text{iSD-T}$ & Trig & Approx. & DiT-B/4 & 7.0 & 52.79 & 117.45 & 157.02 & 186.79 & - \\ 
$\mathcal L_\text{iSD-T}$ & Trig & Approx. & DiT-B/4 & 10.0 & 46.28 & 100.12 & 137.87 & 164.59 & - \\ 
\midrule
MeanFlow & Linear & Exact & DiT-B/4 & 3.0 & 45.38 & 100.49 & 140.32 & 167.60 & \underline{20.39} \\ 
MeanFlow & Trig & Exact & DiT-B/4 & 3.0 & 39.22 & 89.40 & 123.32 & 147.18 & 22.16 \\ 
Shortcut Model & Linear & Exact & DiT-B/4 & 1.5 & 13.58 & 15.75 & 16.25 & 14.94 & \underline{3.68} \\ 
FACM & Linear & Exact & DiT-B/4 & 1.75 & 24.05 & 39.32 & 52.87 & 65.42 & \underline{4.84} \\ 
\midrule
$\mathcal L_\text{iSD-T}$ & Trig & Approx. & DiT-B/2 & 3.0 & 44.96 & 93.25 & 132.21 & 156.90 & - \\ 
$\mathcal L_\text{iSD-T}$ & Trig & Approx. & DiT-B/2 & 5.0 & 56.63 & 122.93 & 153.30 & 170.99 & - \\ 
$\mathcal L_\text{iSD-T}$ & Trig & Approx. & DiT-L/2 & 3.0 & 61.29 & 122.22 & 159.68 & 178.42 & - \\ 
$\mathcal L_\text{iSD-T}$ & Trig & Approx. & DiT-XL/2 & 3.0 & 64.95 & 127.97 & 165.02 & 185.29 & - \\ 
$\mathcal L_\text{iSD-T}$ & Trig & Approx. & DiT-XL/2 & 5.0 & 66.44 & 140.28 & 179.87 & 201.29 & - \\ 
\midrule
$\mathcal L_\text{iSD-T}$ & Linear & Approx. & DiT-XL/2$^\dagger$ & 6.0 (Eval) & 97.80 & 172.30 & 204.08 & 219.71 & - \\ 
$\mathcal L_\text{iSD-T}$ & Linear & Approx. & DiT-XL/1$^\dagger$ & 6.0 (Eval) & 200.26 & 278.90 & 295.43 & 304.74 & - \\ 
\bottomrule
\\ \\

\toprule
Loss & Interp. & JVP & Arch. & Pre-CFG $\omega$ & 500K & 600K & 700K & 800K \\
\midrule
$\mathcal L_\text{iSD-T}$ & Trig & Approx. & DiT-XL/2 & 3.0 & 199.70 & 212.66 & 225.43 & 233.06 \\ 
$\mathcal L_\text{iSD-T}$ & Trig & Approx. & DiT-XL/2 & 5.0 & 220.07 & 233.32 & 234.85 \\ 
$\mathcal L_\text{iSD-T}$ & Linear & Approx. & DiT-XL/2$^\dagger$ & 6.0 (Eval) & 235.11 & 245.51 & 251.24 & 257.96 \\ 
$\mathcal L_\text{iSD-T}$ & Linear & Approx. & DiT-XL/1$^\dagger$ & 6.0 (Eval) & 307.30 & 313.50 \\
\bottomrule
\end{tabular}
}}
\end{table}
\vspace*{\fill}

\FloatBarrier

\section{Qualitative Results}

\begin{figure}[H]
\centering
\begin{minipage}{0.05\linewidth}
    \centering
    \rotatebox{90}{\small{Class 7}}
\end{minipage}
\begin{minipage}{0.75\linewidth}
    \centering
    \includegraphics[width=\linewidth]{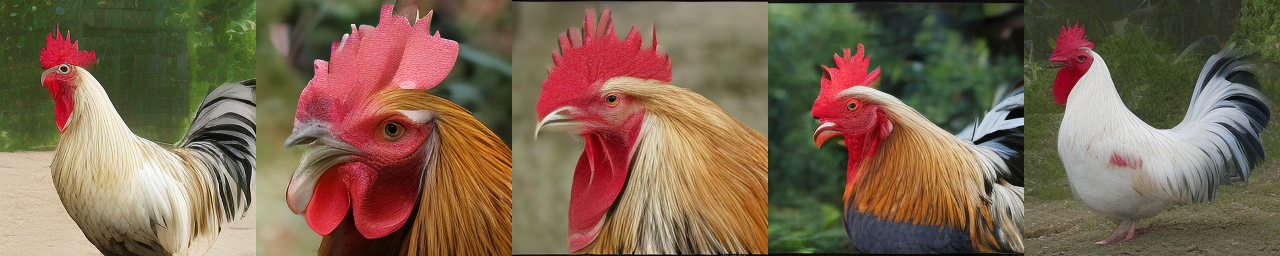}
\end{minipage}

\begin{minipage}{0.05\linewidth}
    \centering
    \rotatebox{90}{\small{Class 139}}
\end{minipage}
\begin{minipage}{0.75\linewidth}
    \centering
    \includegraphics[width=\linewidth]{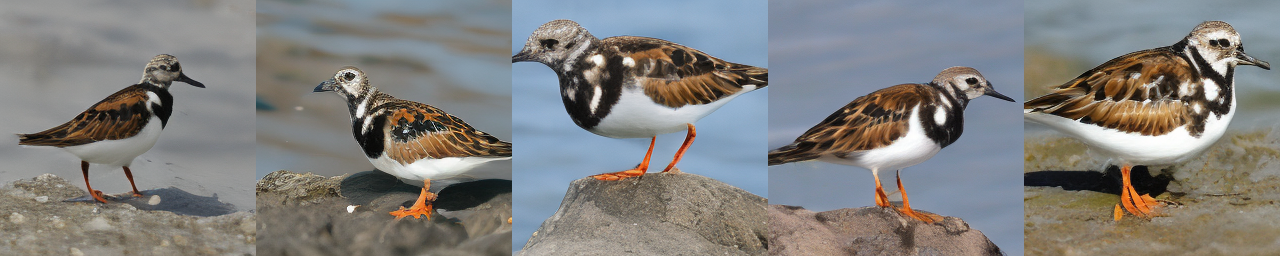}
\end{minipage}

\begin{minipage}{0.05\linewidth}
    \centering
    \rotatebox{90}{\small{Class 193}}
\end{minipage}
\begin{minipage}{0.75\linewidth}
    \centering
    \includegraphics[width=\linewidth]{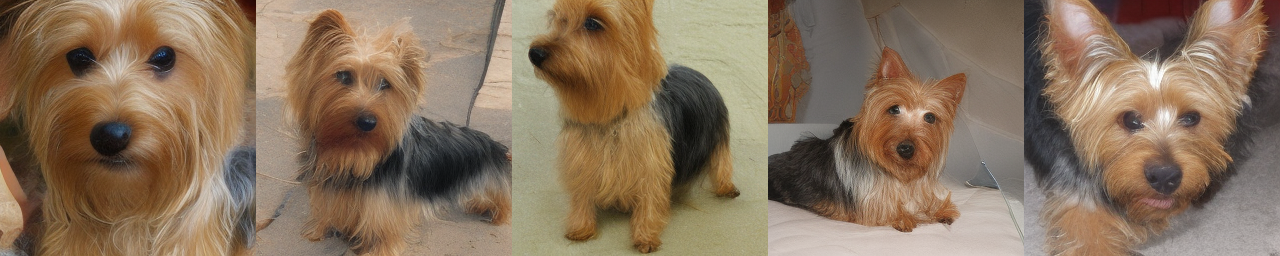}
\end{minipage}

\begin{minipage}{0.05\linewidth}
    \centering
    \rotatebox{90}{\small{Class 607}}
\end{minipage}
\begin{minipage}{0.75\linewidth}
    \centering
    \includegraphics[width=\linewidth]{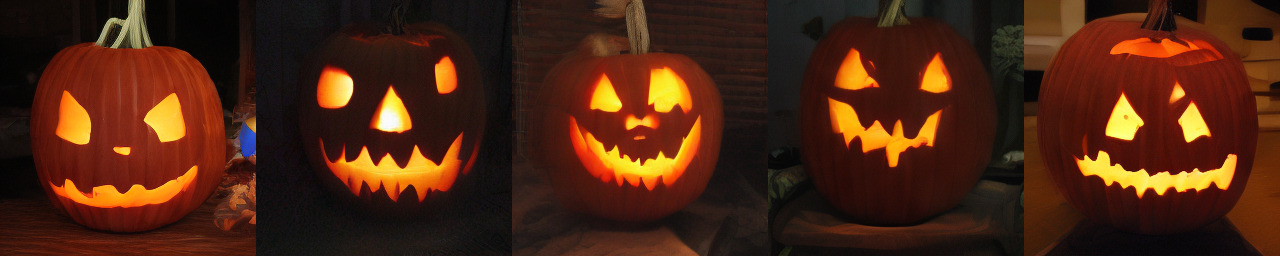}
\end{minipage}

\begin{minipage}{0.05\linewidth}
    \centering
    \rotatebox{90}{\small{Class 959}}
\end{minipage}
\begin{minipage}{0.75\linewidth}
    \centering
    \includegraphics[width=\linewidth]{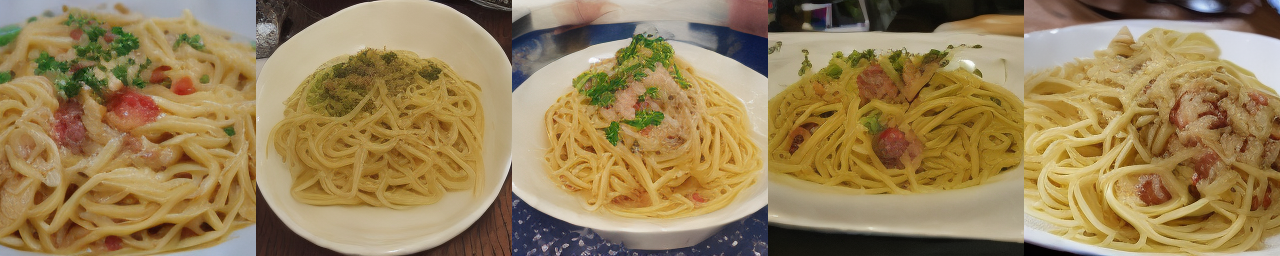}
\end{minipage}

\begin{minipage}{0.05\linewidth}
    \centering
    \rotatebox{90}{\small{Class 970}}
\end{minipage}
\begin{minipage}{0.75\linewidth}
    \centering
    \includegraphics[width=\linewidth]{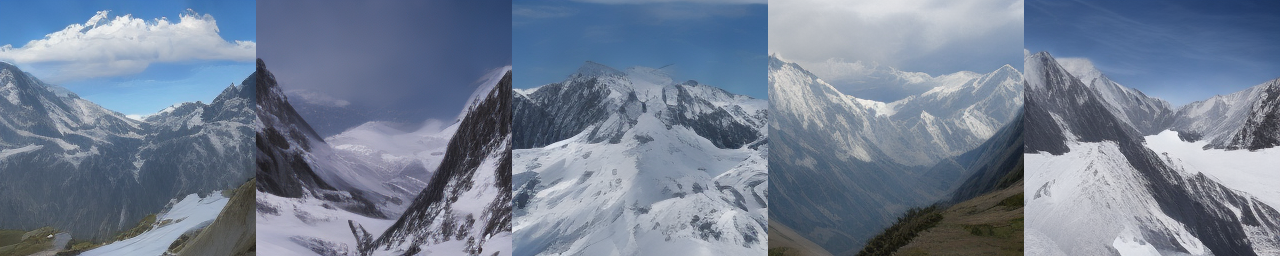}
\end{minipage}

\begin{minipage}{0.05\linewidth}
    \centering
    \rotatebox{90}{\small{Class 973}}
\end{minipage}
\begin{minipage}{0.75\linewidth}
    \centering
    \includegraphics[width=\linewidth]{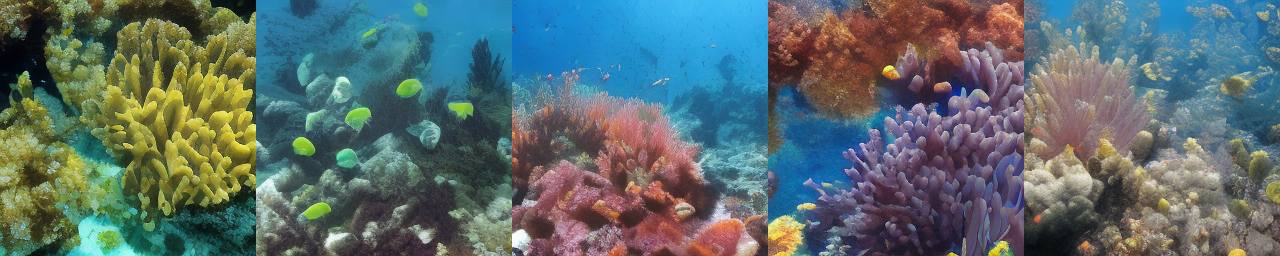}
\end{minipage}

\begin{minipage}{0.05\linewidth}
    \centering
    \rotatebox{90}{\small{Class 985}}
\end{minipage}
\begin{minipage}{0.75\linewidth}
    \centering
    \includegraphics[width=\linewidth]{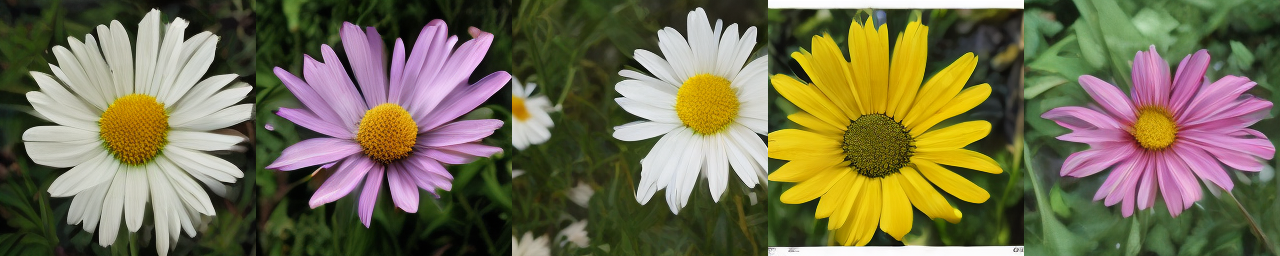}
\end{minipage}
\caption{Class-level samples generated by iSD-T with two-step sampling on ImageNet $256\!\times\!256$.}
\end{figure}

\vspace*{\fill}
\begin{figure}[H]
\centering
\includegraphics[width=0.9\linewidth]{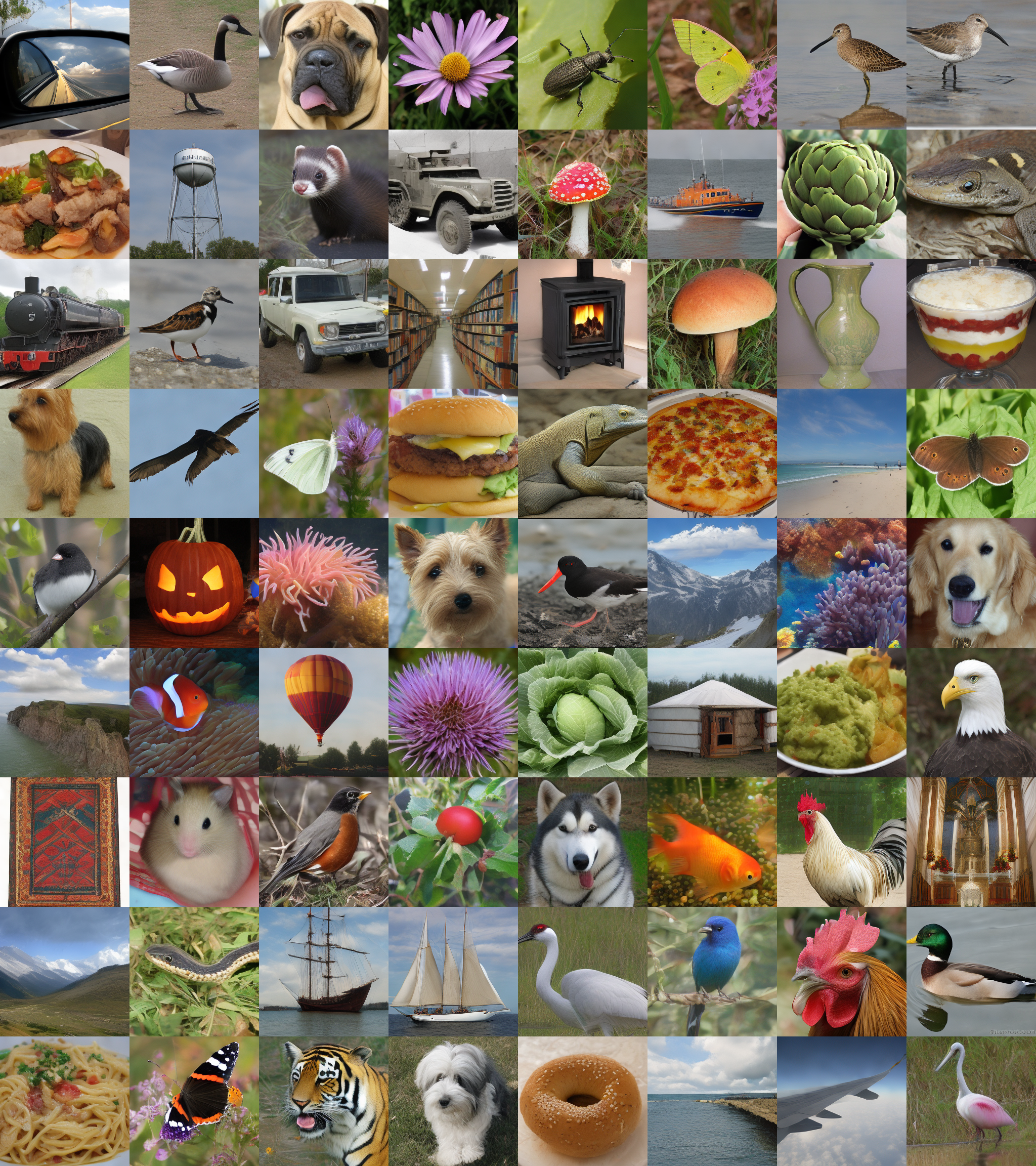}
\caption{Two-step samples from iSD-T on ImageNet $256\!\times\!256$.}
\end{figure}
\vspace*{\fill}

\end{document}